\documentclass[10pt,twocolumn,letterpaper]{article}

\usepackage[pagenumbers]{cvpr} 

\usepackage{epsfig}
\usepackage{graphicx}
\usepackage{amsmath}
\usepackage{amssymb}
\usepackage{float}
\usepackage{booktabs}
\usepackage{multirow} 
\usepackage{amsthm}
\usepackage{mathrsfs}
\usepackage[T1]{fontenc}
\usepackage{amsthm}
\usepackage{bbm}
\usepackage{listings}
\usepackage{algorithm}  
\usepackage{algorithmic}  

\definecolor{cvprblue}{rgb}{0.21,0.49,0.74}
\usepackage[pagebackref,breaklinks,colorlinks,allcolors=cvprblue]{hyperref}

\def\paperID{9111} 
\def\confName{CVPR}
\def\confYear{2025}

\title{MageBench: Bridging Large Multimodal Models to Agents}

\makeatletter
\def\thanks#1{\protected@xdef\@thanks{\@thanks
        \protect\footnotetext{#1}}}
\makeatother

\author{%
    Miaosen Zhang$^{1}$\footnotemark[2] \ \ 
    Qi Dai$^{2}$\footnotemark[3] \ \ 
    Yifan Yang$^{2}$ \ \ 
    Jianmin Bao$^{2}$ \ \ 
    Dongdong Chen$^{2}$ \ \ 
    Kai Qiu$^{2}$
    \\
    Chong Luo$^{2}$ \ \ 
    Xin Geng$^{1}$\footnotemark[3] \ \ 
    Baining Guo$^{2}$\footnotemark[3] 
    \vspace{0.2cm}\\
    {$^1$Southeast University} \quad  
    {$^2$Microsoft} \\
    \texttt{\small\{miazhang,xgeng\}@seu.edu.cn} \quad 
    \texttt{\small\{qid,bainguo\}@microsoft.com}\\
}

\begin{document}
\twocolumn[{%

\vspace{-40pt}

\maketitle

\vspace{-30pt}
\begin{figure}[H]
\hsize=\textwidth 
\centering	
\includegraphics[width=1\textwidth, trim=80 70 80 50, clip]{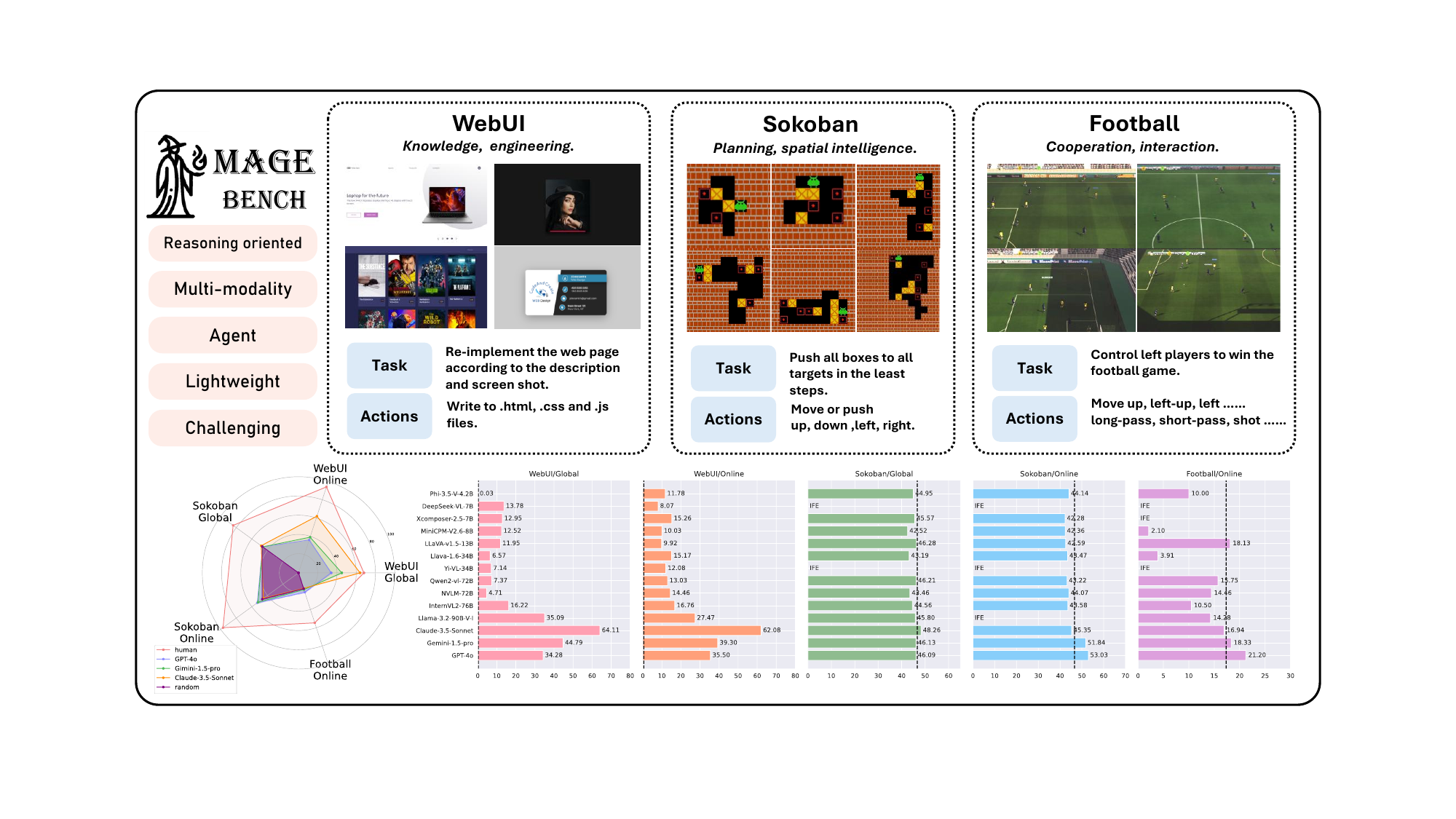}
	\caption{Overview of the MageBench. MageBench is a multi-modality agent benchmark as well as a lightweight and fast platform for reasoning oriented agent researches. It currently contains three enviroments: WebUI, Sokoban and Football. The results indicate that the existing models are still far from reaching human-level performance as an agent. Only a few models outperform the results of random actions, represented by the black dashed line in the bar chart.}
	\label{fig:overview}
\end{figure}
}]

\renewcommand{\thefootnote}{\fnsymbol{footnote}}
\footnotetext[2]{The work is completed during internship at Microsoft Research Asia.}
\footnotetext[3]{Corresponding authors.}

\vspace{-5pt}
\begin{abstract}
LMMs have shown impressive visual understanding capabilities, with the potential to be applied in agents, which demand strong reasoning and planning abilities.
Nevertheless, existing benchmarks mostly assess their reasoning abilities in language part, where the chain-of-thought is entirely composed of text.
We consider the scenario where visual signals are continuously updated and required along the decision making process.
Such vision-in-the-chain reasoning paradigm is more aligned with the needs of multimodal agents, while being rarely evaluated.
In this paper, we introduce \textbf{MageBench}, a reasoning capability oriented multimodal agent benchmark that, while having light-weight environments, poses significant reasoning challenges and holds substantial practical value. This benchmark currently includes three types of environments: WebUI, Sokoban, and Football, comprising a total of 483 different scenarios. 
It thoroughly validates the agent's knowledge and engineering capabilities, visual intelligence, and interaction skills. 
The results show that only a few product-level models are better than random acting, and all of them are far inferior to human-level. 
More specifically, we found current models severely lack the ability to modify their planning based on visual feedback, as well as visual imagination, interleaved image-text long context handling, and other abilities. 
We hope that our work will provide  optimization directions for LMM from the perspective of being an agent. We release our code and data at \url{https://github.com/microsoft/MageBench}.
\end{abstract}

\section{Introduction}
The advent of Large Language Models (LLMs)~\cite{gpt3, palm, llama1, llama2, llama3, OPT}, and Large Multimodal Models (LMMs)~\cite{llava, gpt4v, gemini, flamingo} has revolutionized the fields of natural language processing and computer vision. These models have demonstrated remarkable capabilities across a variety of classical tasks, including translation~\cite{llmtrans, llmtransadaptive, llmtransdocument, llmtransprompt}, summarization~\cite{llmsumbenchmarking, llmsumfew, llmsummarization}, VQA~\cite{lmmvqaicl, lmmvqaq, VQAv2, lmmvqaimage}, captioning~\cite{lmmcapqcaption, mmiclexploring} and etc.
The more recent o1 model~\cite{o1} also stands out due to its exceptional reasoning abilities, particularly on math and coding.
The leap in reasoning capability of LLMs has paved the way for the development of LLM-based agents, which harness the power of these models to autonomously perform a range of sophisticated tasks, from engaging in meaningful dialogue to executing intricate problem-solving strategies. 

Compared to LLM-based agents, the LMM-based agents can be more attractive with the involvement of visual signals, which explicitly expands the boundaries of applications, \emph{e.g.} robotics and autonomous driving.
Consequently, there remains a critical need for comprehensive benchmarks that can evaluate the reasoning and planning capabilities of LMMs from the perspective of agents. 
Unfortunately, rare efforts have been made -- existing benchmarks for LMMs mainly focus on the simple VQA problems~\cite{VQAv2, GQA, seedbench, mmbench, mmvet, MME}; their reasoning assessment generally relies on the language part, which does not require interleaved involvement of visual signals~\cite{scienceqa, CCoT, mmcot, cocot, ddcot, cantor} . 
Such evaluations are not suitable for the evolving demands of visual agents. 
While there exist plenty works of LMM agent~\cite{appmmmobile, wukong, webarena} that can be adapted to evaluate models, they often focus too much on the environment or agent solution and have a complex and specific scenario. This not only makes the evaluations costly, but also leads to limited generalization capability when environment changed.

\begin{figure}
    \centering
    \includegraphics[width=0.95\linewidth, trim=70 120 70 40, clip]{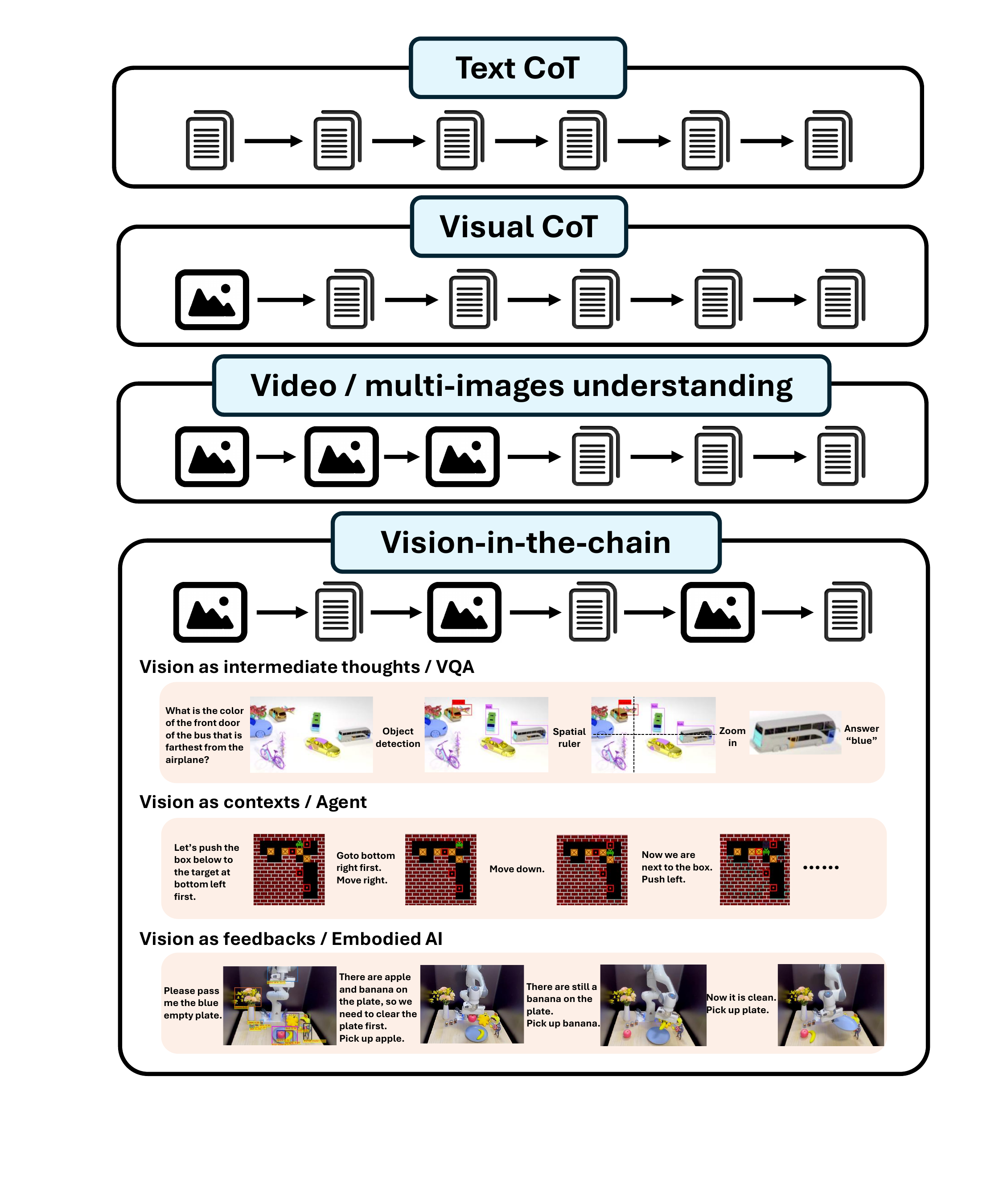}
    \caption{The difference between vision-in-the-chain reasoning and existing reasoning paradigm. Example images are adapted from~\cite{imageofthought, sokoban, vila}.}
    \label{fig:related-work}
    \vspace{-15pt}
\end{figure}

In this work, we target at building a reasoning capability oriented benchmark for evaluating LMM's potential of being an agent regarding complex visual tasks.
When defining `complex visual task,' we expect not only the reasoning of initial visual input, but also the continuous understanding of visual feedback throughout the entire process.
These tasks require models to dynamically interact with visual information, continually updating their understanding and decisions based on new visual cues, much like a human would. 
We refer to this novel reasoning paradigm, which integrates other modality (vision) into the reasoning chain, as Vision-in-the-Chain (ViC), as illustrated in the last block of Figure \ref{fig:related-work}.
Technically, the ViC paradigm is fundamentally different from previous reasoning paradigms, \emph{e.g.} text chain-of-thought (CoT)~\cite{cot, selfconsist, 0scot, star, autocot} and visual CoT~\cite{ CCoT, mmcot, cocot, ddcot, cantor}.
The latter two paradigms only perform text-based reasoning with multiple intermediate steps, without incorporating the visual signals at each step, as shown in Figure \ref{fig:related-work}.
The continuous integration of visual feedback in ViC ensures that models can handle intricate tasks, \emph{e.g.} navigation and driving, which is more aligned with the needs of agents~\cite{agentreview, agentsurvey, multiagentreview} and robotics~\cite{robolarge, roboticlarge}.

With the above concept, we present \textbf{MageBench}, a \textbf{M}ultimodal \textbf{AGE}nt benchmark for evaluating the LMM's capability of being an agent on complex visual tasks.
In addition to language reasoning capabilities, we believe that for LMMs to become truly multimodal agents, they must possess several additional competencies. As human assistants, they require cross-modal knowledge and engineering capabilities. In the field of robotics, spatial intelligence and planning abilities are essential. Moreover, social and interactive capabilities, as fundamental characteristics of intelligent agents~\cite{agentformalizing, agentintelligent, agentreview}, are crucial for the development of more powerful multi-agent systems. To address the gap in evaluating these competencies, we have introduced three elaborately designed testing environments \emph{i.e.} WebUI, sokoban puzzle, and football game, in MageBench.
WebUI requires the agent to build a target webpage given its screenshot and description, demanding rich knowledge and engineering ability of agent.
Sokoban puzzle asks the agent to push boxes to get them to the correct locations, in which the planning ability and spatial intelligence are expected.
Football game requests the agent to operate one player (\emph{i.e.}, always control the ball handler) to win the game, which inspects the cooperation and interaction skills of the agent.
We investigated and rejected a large number of environments and ultimately chose these three because they are simple yet representative from reasoning perspective, with enormous planning space, reflecting the visual reasoning and planning capabilities of the agent.

We propose two baseline agent setting: Global (the model only observes the initial state and gives all actions) and Online (the model interacts with the environment to continuously obtain image observations and output actions), which correspond to Visual CoT and ViC types of reasoning, respectively. 
We tested 14 strongest open-source and close-source LMMs selected from each model family, and the level of human performance in Tab. \ref{tab:model-noHK-eval}, and more models in Appendix. \ref{app:more-result}. %
The results are summarized as follows: Firstly, In the Online setting, only GPT-4o and Gemini-1.5-pro outperformed the random level, and all of them are far inferior to human-level\footnotemark[2]. This shows that they severely lack ViC-type reasoning capabilities, making existing LMMs far from ideal for agent and robotics applications. In addition, the existing models do a good job in the Global setting of WebUI. Claude can even approach human-level. However, they failed to boost the result with browser's rendering feedback, but human can continually adapt their codes to nearly perfection.
This is possibly caused by the shortcomings of interleaved image-text long context handing.
Last but not least,
their best-of-N performances in Sokoban are far worse than human best-of-1, which indicates that the existing models lack mechanism-level shortcomings in visual imagination and thinking-ahead ability. 

\footnotetext[2]{Human level: In this context, "human level" refers to the performance metrics achieved by highly educated individuals with PhDs who were aware that their work would contribute to our reported results in this specific benchmark. More details can be found in Sec. \ref{sec:standardsetting}}

\section{Related Work}

\noindent\textbf{Large Multimodal Models.}
The advent of large language models (LLMs)~\cite{gpt3, gpt4v, llama3, palm} has demonstrated remarkable reasoning capabilities~\cite{cot, selfconsist} and the potential for general intelligence~\cite{agifar}. By employing a single model with different prompts, a multitude of tasks can be accomplished~\cite{promptreview, promptsurvey}. A natural extension of this concept is to apply similar methods to other modalities to achieve general multimodal intelligence. Flamingo~\cite{flamingo} was among the first to explore multimodal in-context learning~\cite{mmiclexploring, mmiclglance, mmiclvisual}, followed by the emergence of numerous large multimodal models~\cite{gpt4v, gpt4o, gemini, claude, grok2, llava}. These models employ various technical approaches~\cite{tmaligning, tmlearning, tmvsr, llava, llava15, llavanext}. As technology has progressed, product-level multimodal large models such as GPT-4V~\cite{gpt4v}, GPT-4O~\cite{gpt4o}, Gemini~\cite{gemini}, Claude~\cite{claude}, and Grok-2~\cite{grok2} have showcased state-of-the-art performance.

\noindent\textbf{Visual Reasoning.}
Chain-of-thought prompting~\cite{cot, selfconsist, 0scot}, flow engineering~\cite{alpgacodium}, self-reflection~\cite{react, reflexion}, and their various variants~\cite{LATS} have demonstrated significant improvements. In visual tasks, the primary evaluation datasets for visual reasoning are those based on VQA tasks, such as ScienceQA~\cite{scienceqa} and MathVista~\cite{mathvista}. Due to the limitations of these evaluation datasets, many existing studies~\cite{CCoT, mmcot, cocot, ddcot, cantor} on visual reasoning using ``CoT" as a keyword mainly focus on extracting information from multimodal problems, and then utilize text-based intermediate processes such as captioning~\cite{mmcot}, rationales~\cite{scienceqa}, relational graphs~\cite{CCoT}, and question tables~\cite{ddcot}. 

Some recent works have attempted to incorporate procedural information from other modalities, not just text, into the reasoning process to accomplish classic visual tasks. For instance, Image-of-thought~\cite{imageofthought} introduces image editing tools to facilitate better information perception by the model. Similarly, DetToolChain~\cite{dettoolchain} incorporates auxiliary metrics and editing tools to enhance the model's capability in performing high-precision object detection tasks. 
However, these ViC type reasoning works are constrained to classical vision tasks.

\noindent\textbf{Large Model Based Agents.}
AI agents are artificial entities that sense their environment, make decisions, and take actions. With the development of large models, there has been an explosive increase in research treating large models as the brains of agents~\cite{agentreview}. Researchers are utilizing LLMs to study single-agent systems~\cite{sinaautogpt, sinamind2web, sinamultimodal, sinavoyager}, multi-agent collaboration~\cite{mulacamel, mulachateval, mulachatllm, mulametagpt}, and human-agent collaboration~\cite{humanartificial, humansapien} in specific environments. The emergence of LMMs has further enhanced the perceptual modalities of agents, enabling LMM agents to perform tasks in multimodal scenarios~\cite{Cerebrum, lmmagentservey, toollmm}.
These efforts focus more on the specific environment itself and the agent solutions. They often come with complex dependencies (both hardware and software)~\cite{wukong, micraft, ravens, Cerebrum}, require database setups~\cite{webarena, webshop}, making them not convenient and general to directly evaluate across models.

\noindent\textbf{Benchmarking LMMs.}
There are numerous evaluation datasets for LMMs that comprehensively assess various capabilities such as general VQA~\cite{VQAv2, GQA, seedbench}, visual expertise~\cite{mmmu}, visual perception~\cite{blink}, and visual reasoning~\cite{scienceqa, mathvista}. From the perspective of application domains, some studies also evaluate models' understanding of tables~\cite{tablevqa}, charts~\cite{chartqa, mchartqa, multichartqa}, geometry~\cite{geoeval, olympiadbench}, and real-world scenarios~\cite{mmro}. Most of these evaluations are presented in the form of multiple-choice questions, which simplify and abstract real-world problems.

Another approach to evaluating models is to deploy them on agents for task-level end-to-end assessments. 
Previous attempts such as AgentBench~\cite{llmagentbench} and VisualAgentBench~\cite{visualagentbench} integrate several environments in one repository. However, existing works are environment or task oriented, but we are reasoning oriented, \emph{i.e.}, not all meaningful environments require meaningful and representative reasoning skills. We focus on what abilities of LMM are agents' need. We will elaborate more on this in Sec.\ref{sec:magebench} and Appendix. \ref{app:env-select}. %

\section{Introducing MageBench}
\label{sec:magebench}


MageBench aims to select most simple, representative environments from the perspective of reasoning and with generalization ability. We investigate dozens of environments and select those meet the criteria below:
\begin{itemize}
    \item \textbf{Representativeness on Reasoning}: 
    To be an ideal agent, LMMs need real-world engineering knowledge to assist humen (WebUI). To manipulate in real-world scene or virtual UI, they need spatial understanding and planning (Sokoban). In the future, we may also expect the extension to multi-agent system (Football). Many robotics simulation environments(\emph{e.g.}, ~\cite{ravens, virtualhome, innermonologue} and OmniGibson in~\cite{visualagentbench}), Virtual reality game (\emph{e.g.}, ~\cite{wukong, micraft, micraftb} and Micraft in~\cite{visualagentbench}) and app manipulation (\emph{e.g.}, ~\cite{appagent, appbench, appflow, appmmmobile, appmobileagentbench}), although more real and practical with complex actions, their high level planning are actually simple and direct.
    For example, to move an object in simulation environments or search an commodity in a webpage, models may make instruction following errors and perceiving errors, but there isn't much operability at the planning level.
    So we exclude them.

    \item \textbf{Visual Feedback}: 
    The visual feedback must be indispensable for the relevant tasks. In other words, we will not select an agent environment if it can be effectively completed without image input. For example, we found using text feedback of the webpage is enough for many of the tasks in webpage operating environments~\cite{webshop, webarena}.
    
    \item \textbf{Simplicity}: The environments are highly streamlined, with minimal database, library, and hardware requirements. They offer fast simulation speeds and good scalability. On one hand, the simplicity of these environments allows researchers to quickly get started and facilitate dissemination. On the other hand, low latency, minimal cost, and reduced energy consumption are essential for potential future RL training and scaling.

\end{itemize}
Based on the above criteria, we have selected and constructed the following three environments: WebUI, Sokoban, and Football. There are a few that also meet the requirements but are of a similar type, and we have chosen these three as representatives. We will introduce each of our environments in detail below.
\subsection{WebUI}


\begin{figure}
    \centering
    \includegraphics[width=\linewidth, trim=250 60 250 30, clip]{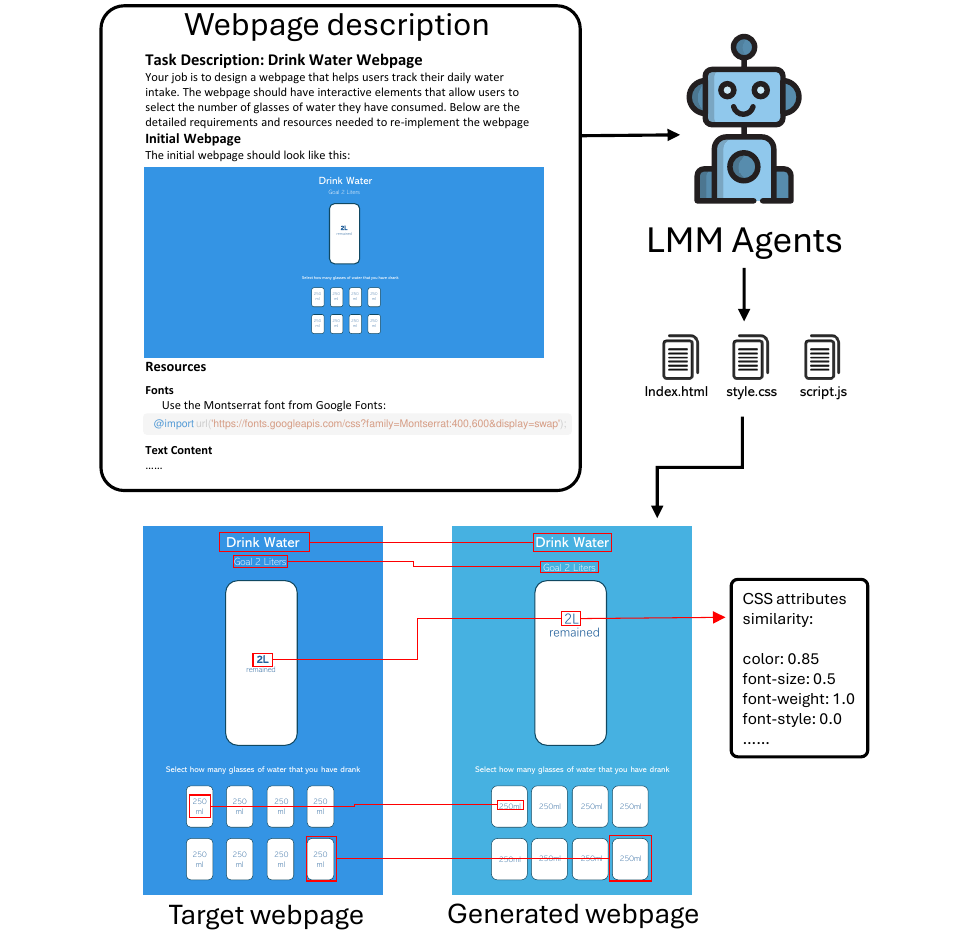}
    \caption{An overview of WebUI and its evaluation. LMM Agents are required to re-generate the webpage according to the description. We match the generated elements with the atomic elements in the ground truth. Then we compare the CSS attributes to obtain a similarity score. A specific example of task description can be found in Appendix. \ref{sec:webtaskdes}. %
    Technique details of evaluation can be found in Appendix. \ref{app:webuieval}. %
    }
    \label{fig:webui-overview}
     \vspace{-15pt}
\end{figure}

We collected minimal web projects from GitHub, strictly adhering to the corresponding licenses, consisting of only a few HTML, JavaScript, and CSS files. For each webpage, we will create a Markdown-formatted webpage description. The webpage description is a image-text interleaved document that provides sufficient information to fully reconstruct the website. It includes detailed webpage descriptions, external resources, and screenshots before and after various interactions and etc. The task of WebUI is to reconstruct the website based on the description and a Google Chrome web driver.

The evaluation of WebUI is based on comparing the CSS properties of atomic elements. We first define \textbf{atomic elements} as follows: Suppose webpage B is a reproduction of webpage A. An HTML tag in webpage A is considered atomic if any attribute it contains is guaranteed to appear in a corresponding tag in webpage B. For example, all headings, texts, and images in webpage A may exist in webpage B as different types of tags (\emph{e.g.}, both h1 and span tags can display text). However, we know that there must be a corresponding tag in webpage B. Such tags that necessarily exist are termed as atomic. In the evaluation process, we first match the atomic elements in the target webpage with the elements in the webpage generated by the agent using a carefully designed matching algorithm. Next, we compare the CSS property similarity of the successfully matched elements, typically using metrics such as relative error or checking if the values are equal. Finally, we provide a weighted similarity score as the evaluation result. We refer to this metric as \textbf{Atomic Element Similarity (AES)}. The technical details involved are extensive and will be presented in the Appendix. \ref{app:webuieval}. %

\subsection{Sokoban}
Sokoban is a well-known two-dimensional logic video game where the task is to maneuver a character to push all boxes onto designated target areas. The game is highly challenging due to the presence of numerous losing states and traps, necessitating strong planning abilities~\cite{sokoban}. The planning and reasoning capabilities of LMMs may effectively mitigate this issue, making this environment ideal for testing an agent's path finding, planning, error correction, and foresight abilities. We utilize the rendering environment provided by ~\cite{sokoban}, simplifying actions to four directions (up, down, left, right) by combining move up and push up. We generated and stored 182 levels of varying difficulty. Further details can be found in Appendix. \ref{app:sokoban}. %

We also adapt the reward value defined in ~\cite{sokoban} to evaluate the LMMs. However, unlike their approach, we use the historically optimal reward throughout the trajectory rather than the final reward. This is because the reward includes a penalty for the number of steps taken. Based on extensive testing, we found that given the current capabilities of LMMs, using the final reward tends to be dominated by factors such as the model's output length, the number of steps we set (for the online setting), the length penalties and etc. This is an outcome we aim to avoid.

\subsection{Football}

\begin{figure}
    \centering
    \includegraphics[width=\linewidth, trim=30 140 20 80, clip]{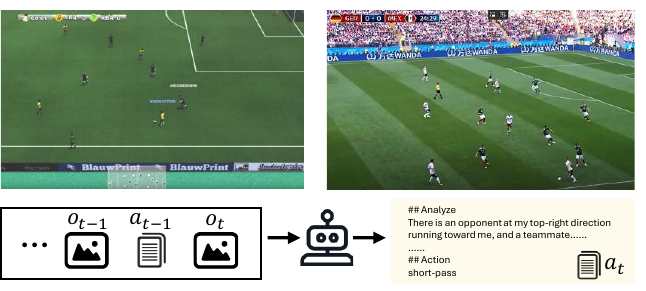}
    \caption{The segment from the Germany vs. Mexico match in the 2018 FIFA World Cup (right), and the initial game scene inspired by it (left). Model analyzes and generates one of the actions (bottom), similar process for the Sokoban-Online.}
    \label{fig:football-overview}
     \vspace{-15pt}
\end{figure}

Football, as one of the most competitive and cooperative sports, can fully demonstrate an agent's spatial intelligence and collective intelligence, potentially providing a research foundation for future LMM multi-agent systems. We chose the rendering platform provided by Google Football Research ~\cite{football} for our study. We generated 108 scenarios as initial states, with each initial scenario serving as a level (analogous to a level in Sokoban). These levels cover different areas of the football field and are categorized into personal (scenarios where good passing routes are unavailable, requiring players to showcase individual skills), teamwork (scenarios suitable for demonstrating team collaboration and passing), and real-world (scenarios from actual World Cup matches). The agent will use text outputs to perform 18 actions  (including moving, long passing, and shot), starting from each scenario and simulating up to 400 frames until a goal is scored or the ball is intercepted. More details can be found in Appendix. \ref{app:football}. %
We also designed an automatic rendering algorithm (see Appendix. \ref{app:auto-render}.%
) that reduces the average number of API calls needed per scenario from 80 to less than 20, without affecting the results.

During the simulation process, the LMM agent will control only one player (always the player in possession of the ball), while the other players are controlled by built-in AI bots. The presence of numerous agents results in a highly stochastic environment simulation. Using metrics such as win-rate leads to high variance and requires extensive repetitions. To address this issue, we carefully designed a more dense reward system to comprehensively evaluate the model's capabilities. The design of this reward system is as follows:
\begin{equation}
\begin{aligned}
    R^{(t)} = &\lambda_1 S_{move}^{(t)} + \lambda_2 S_{oppo}^{(t)} + \lambda_3 \delta_{scored}^{(t)} \\
    &+ \lambda_4 \delta_{stole}^{(t)}\frac{t}{T} + \lambda_5 \delta_{pass}^{(t)}S_{pass}^{(t)}\\
    &+ \lambda_6\delta_{shot}^{(t)}S_{shot}^{(t)},
\end{aligned}
\end{equation}
where $\delta_{event}^{(t)}$ is a indicator function that event happened at time step $t$. $S_{move}^{(t)}$ and $S_{oppo}^{(t)}$ represent the reward values obtained after processing the distance the ball has been moved forward and the number of opponents surpassed, respectively. $S_{pass}^{(t)}$ and $S_{shot}^{(t)}$ are metrics that quantify the quality of passing and shot.
For additional details and specific parameters, please refer to Appendix. \ref{app:footballeval}. %

\begin{table*}[t!]
  \centering
  \small
  \caption{Evaluation on MageBench test-mini subset with unified prompt. IFE stands for ``Instruction Following Error''. It is defined as follows: if more than 90\% of the outputs are not parsed into valid actions, or if 90\% of the actions are the same (indicating that the model is repeating a certain action), it is considered an IFE. $\delta$ represents the significance difference derived from repeated experiments. }
  \scalebox{0.9}{
    \begin{tabular}{l|c|c|c|c|c}
    \toprule
    \multirow{2}[4]{*}{model} & \multicolumn{2}{c|}{WebUI AES (\%)} & \multicolumn{2}{c|}{Sokoban Reward} & Football Reward \\
\cmidrule{2-6}          & Global ($\delta=\pm2.0$) & Online ($\delta=\pm2.6$) & Global ($\delta=\pm1.9$) & Online ($\delta=\pm1.8$) & Online ($\delta=\pm2.2$) \\
    \midrule
    Phi-3.5-V-4.2B~\cite{phi3v} &   0.03 &  11.78
      &   44.95    &   44.14    &  10.00 \\
    DeepSeek-VL-7B~\cite{deepseekvl} &    13.78 &  8.07
             &  IFE        &   IFE     
     &  IFE  \\
     Xcomposer-2.5-7B~\cite{xcomposer} &   12.95       &  15.26
         & 45.57         &   42.28     
     &   IFE  \\
    MiniCPM-V2.6-8B~\cite{minicpm} &    12.52 &  10.03
           &    42.52   &   42.36    &  2.10 \\
    LLaVA-v1.5-13B~\cite{llava15}  &    11.95 &  9.92
        &   46.28    &    42.59   &  18.13   \\
    Llava-1.6-34B~\cite{llavanext}   &    6.57 &  15.17
            &    43.19   &    43.47   & 3.91 \\
    Yi-VL-34B~\cite{Yi}   &   7.14   &  12.08
            &     IFE  &   IFE    &  IFE  \\
    Qwen2-vl-72B~\cite{qwen2vl}   &   7.37 &   13.03
              &    46.21  & 43.22     & 15.75 \\
    NVLM-72B~\cite{nvlm}   &    4.71  &  14.46
           &    43.46   &  44.07     & 14.46 \\
    InternVL2-76B-LLaMA3~\cite{internvl}   &  16.22   &  16.76
           &    44.56   &   43.58    &   10.50   \\
    Llama-3.2-90B-Vision-Instruct~\cite{llama3}   &    35.09  & 27.47
          &    45.80   &   IFE   &  14.28 \\
    \midrule
    Claude-3.5-Sonnet~\cite{claude} &    \textbf{64.11}   &   \textbf{62.08}
       &   \textbf{48.26}
      &  45.35  &  16.94  \\
    Gemini-1.5-pro~\cite{gemini} &   44.79  &    39.30
           &   46.13  
      &  51.84  &   18.33 \\
    GPT-4o~\cite{gpt4o} &   34.28   &  35.50
       &    46.09   &  \textbf{53.03}    & \textbf{21.20} \\
    \midrule
    Idle Baseline & 0.00&  0.00 
        &  41.18 & 41.18  &  2.53  \\
    Random Baseline &  0.00  &  0.00
        &  46.61   &   46.61  &  17.33  \\
    Human &   68.71 &  
      94.32   &   83.63  
      &  96.85  &  54.68  \\
    \bottomrule
    \end{tabular}%
    }
  \label{tab:model-noHK-eval}%
   \vspace{-10pt}
\end{table*}%

\section{LMM-as-Agents}
We represent our Agent-Environment system and its evaluation using a Partially Observable Markov Decision Process $(\mathcal{S}, \mathcal{A}, \mathcal{T}, \mathcal{R}, \mathcal{O})$, indicating that the agent $\pi_\theta$ cannot directly access the complete state from state space $\mathbf{s_t} \in \mathcal{S}$. Instead, the agent has to observe vision feedback (and probably few text feedback for WebUI) in observation space $\mathbf{o_t} \in \mathcal{O}$ , and leverage its planning ability to generate actions within a discrete action space $\mathbf{a_t} \in \mathcal{A}$. The environment will serve as the transition function to update the state based on the agent's actions: $\mathcal{T}: \mathcal{S} \times \mathcal{A} \rightarrow \mathcal{S}$. Finally, our designed reward assigning function $\mathcal{R}$ will provide the evaluation results. 

There will also be some pre-defined prompts involved in the agent. System prompt $p_{sys}$ introduce the rule, target, available actions of the system. 
In WebUI, task prompt $p_{task}$ is a specific description of a Webpage. For other environment, $p_{task} = \mathbf{o_1}$.
CoT prompt $p_{cot}$ and IO prompt $p_{io}$ designate the model's inner thought flow and output format so that we can parse the text-based output to actions. 

In addition to the basic formulation mentioned above, we will also formalize the following two fundamental agent designs, which will serve as baseline agents for the dataset proposed in this paper:
\begin{itemize}
    \item
    \textbf{Global Planner Agent.} We consider the following basic Agent design: it only observes the initial environment once (in the task prompt). Then, it continuously makes all subsequent decisions, formulated as 
    \vspace{-7pt}
    \begin{equation*}
    \pi_\theta(p_{sys}, p_{task}, p_{cot}, p_{io})\rightarrow \mathbf{a_1}, \mathbf{a_2}, ..., \mathbf{a_T}.
    \end{equation*}
    \vspace{-15pt}
    \item
    \textbf{Online Planner Agent.} This baseline agent analyzes each step and takes actions online while possessing two attributes: action-memory ($AM$) and observation-memory ($OM$), where $OM \leq AM$, which are hyperparameters of the agent. For example, when $AM = 3$ and $OM = 2$, the decision-making process of the agent can be formalized as follows: 
    \vspace{-7pt}
    \begin{equation*}
    \pi_\theta(p_{sys}, \mathbf{a_{t-2}}, \mathbf{a_{t-1}}, \mathbf{o_{t-1}}, \mathbf{a_{t}}, \mathbf{o_{t}}, p_{cot}, p_{io})\rightarrow \mathbf{a_{t+1}}.
    \end{equation*}
    We separate these two kinds of memory because the number of input images is limited for most of the LMMs. For WebUI, we will also add $p_{task}$ after the $p_{sys}$.
\end{itemize}

Global planner can better reflect the planning capabilities of the model itself. It is similar to the process used to generate zero-shot CoT results for existing QA-based benchmarks. The online planner represents another batch of agents that receive feedback from the system and adapt themselves, such as Reflexion~\cite{reflexion}, ReACT~\cite{react}, ViLA~\cite{vila}, etc.

\section{MageBench Results and Analysis}

\subsection{Standard setting}
\label{sec:standardsetting}

We use the MageBench and the unified agent setting as the carrier to study what kind of LMMs have the potential to become an agent. It is worth stating that these models may perform better in MageBench with specially designed agents, prompts, and settings, but for the sake of a fair comparison, we will use the same standard settings below.
\begin{itemize}
    \item Except for WebUI, the $p_{cot}$ will  prompt the model to analyze before actions, but will not specify how to. This is similar to zero-shot CoT~\cite{0scot} and the prompt is shared across all environment. This is more effective in validating the model's ability to self-plan rather than being compensated for by the instruction-following ability.
    \item Online planner will use $AM=5$ and $OM=1$, so that we can fairly evaluate models without multi-images capability. According to our tests, these two types of memory do not have a significant impact on the performance for current models, See Appendix. \ref{app:agentmem}. %
    \item In order to minimize the influence of non-reasoning factors, we will give the model two retries only if it caused a parsing error.
    \item Agent-based evaluation always has a large stochastic variance, and in addition to the careful design we did for the metric, we also require all experiments to be repeated and averaged. For MageBench mini set, Sokoban and WebUI should repeat 3 times and Football repeat 10 times. For MageBench complete set (in Appendix. \ref{app:more-result}
    ), all experiments reported are averaged over 3 repetition.
\end{itemize}

\begin{figure}
    \centering
    \includegraphics[width=\linewidth]{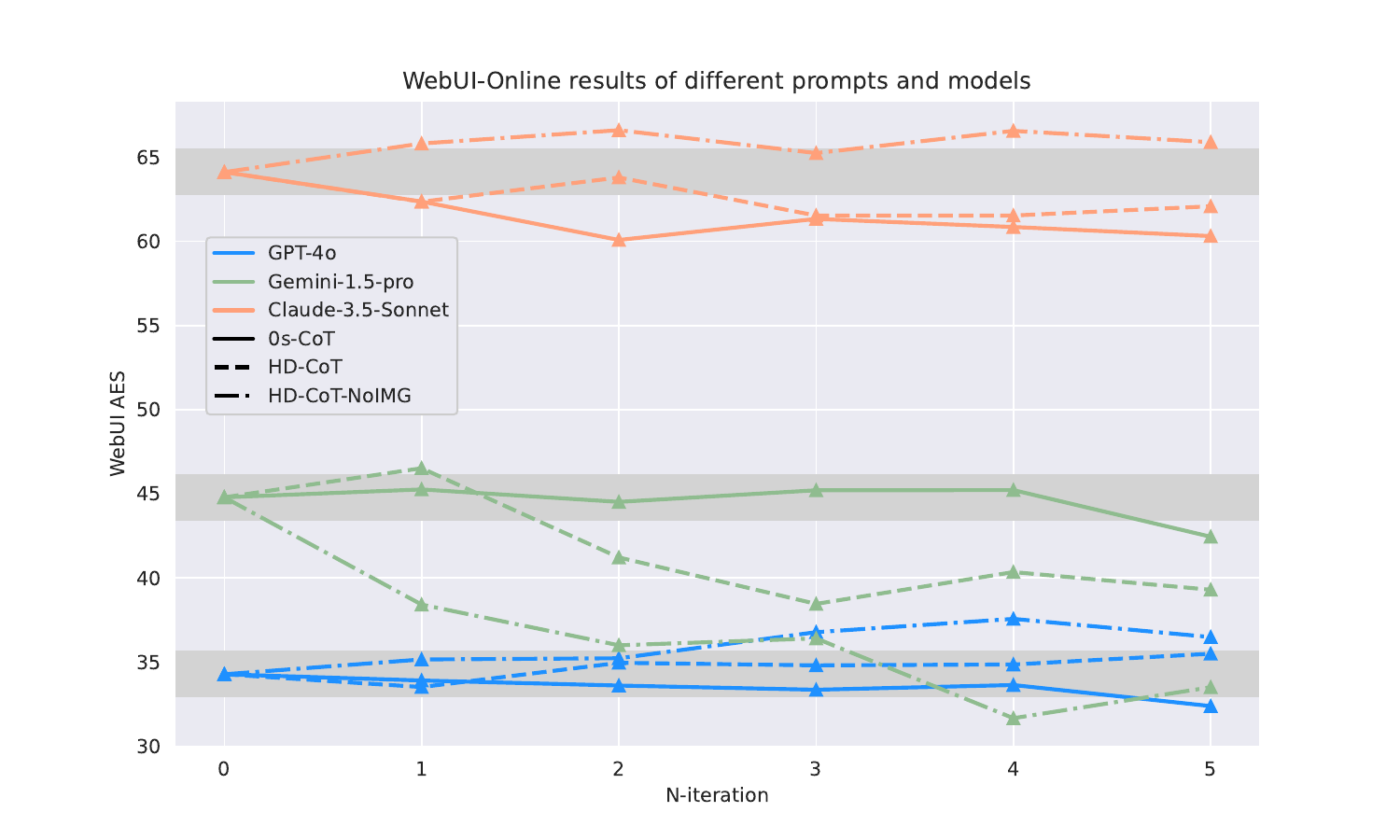}
    \vspace{-20pt}
    \caption{WebUI-Online results. Different line styles represent different prompt types, while different colors denote different models. The horizontal axis shows the number of iterations the model takes to modify the webpage code based on feedback. The gray-shaded areas indicate regions of variance.}
    \label{fig:webui-online}
    \vspace{-12pt}
\end{figure}

We select LMMs that are trained for general usages and support flexible interleaved image-text inputs, and have at least 4096 context length to evaluate. 
Table \ref{tab:model-noHK-eval} presents the test results under the standard settings on the MageBench mini subset. 
We evaluated the strongest models from each open-source LMM series (first block) and the results of three closed-source product-level models (second block). 
We also provided idle and random baselines, as well as human-level results in the third block. For Sokoban and Football in the idle baseline, no actions were taken (an idle action is available in the football environment). The random baseline refers to randomly selecting a possible action. During human-level testing, annotators were selected from several PhD candidates with strong reasoning abilities. These annotators might practice in scenarios outside of the mini set, but the testing conditions were completely fair when compared to the models. For example, in the results for Sokoban-Global, humans could not control the player and could only observe the initial screen and record all actions using their imagination. In WebUI-Global, humans were not allowed to view the browser's rendered output while writing code, whereas in the WebUI-Online setting, humans were permitted to observe the rendered screen.

Overall, we found that although open-source models have achieved performance levels comparable to closed-source models on many VQA tasks, they still fall significantly short of the requirements for AI agents. In the Sokoban and Football, only GPT-4o and Gemini performed better than the random baseline under the online setting. This may be attributed to the optimization of product-level models for multi-turn dialogue and multi-image scenarios. Claude's performance under the Global setting was very impressive, being the only model that could work in the Global setting for Sokoban, but it still lagged far behind human-level performance. This demonstrates that humans possess strong imaginative and think-ahead abilities, which are substantially lacking in current LMMs.

We were pleasantly surprised to observe that Claude demonstrated performance close to that of computer science PhD candidates in the WebUI-Global results. However, while humans can modify webpage code based on rendered screen to make it almost identical to the target webpage, current models fail to achieve this. In Figure \ref{fig:webui-online}, we evaluated the online agent capabilities of three product-level models in modifying code based on rendered images across different prompts. Specifically, ``0s-CoT'' refers to the model being asked to analyze and adapt the code without specifying how to analyze it; ``HD-CoT'' stands for human-designed analysis processes, such as instructing the model to first observe the differences in the rendered images and then describe how to modify the code; ``HD-CoT-NoIMG'' does not provide the rendered images, only prompts the models to observe errors in the code.
In Table \ref{tab:model-noHK-eval}, We used ``HD-CoT''.

Unfortunately, these models almost uniformly failed to perform well across all settings. Although they may show slight and unstable gains under self-reflection due to correcting some syntax errors.

The prompts and details of all environment and agent settings can be found in Appendix. \ref{app:datasetdetails} and \ref{app:examples}. %

\begin{figure}
    \centering
    \includegraphics[width=0.9\linewidth]{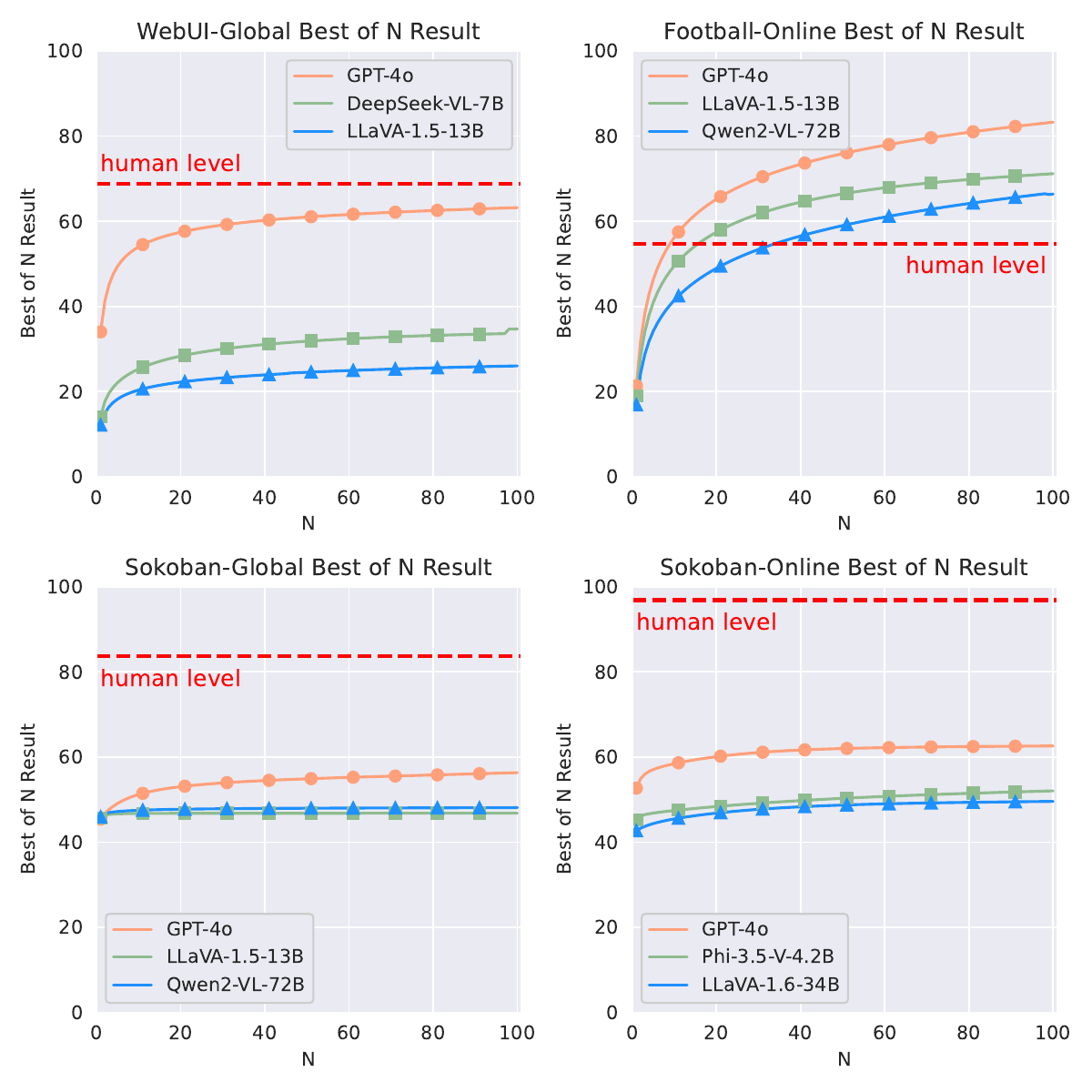}
    \vspace{-12pt}
    \caption{Best-of-N results of selected models.}
    \label{fig:BoN-result}
    \vspace{-18pt}
\end{figure}

\subsection{Best-of-N Result}
Previous subsection has demonstrated the feasibility of existing LMMs as agents. In the following, we use best-of-N scaling curves to investigate the potential of the models in relevant tasks in Fig. \ref{fig:BoN-result}. Firstly, we observe that in the Football environment, many models can surpass human performance by computing the best-of-N, with a steep upward slope. This indicates that there is a significant opportunity to achieve substantial improvements in this task using RL algorithms. However, in the Sokoban task, the best-of-N curve exhibits slow growth, suggesting that directly using RL reinforcement models may not be as effective due to the difficulty of generating valuable trajectories through trial and error and experience accumulation. This implies that current large models may lack the essential mechanisms and training required to achieve human-like imagination or spatial planning capabilities for such problems. The best-of-N curve on WebUI is very similar to the pass@N metric in code generation. We can observe that both model enhancement and increasing N can lead to substantial gains.

\subsection{Error Statistics and Analysis}
We have categorized the types of errors for different models on MageBench. For Sokoban and Football, since most errors stem from model decision failures, we additionally categorize parsing errors during the parsing process, as shown in Table \ref{tab:error-type}. Among these, ``Invalid Actions'' refer to errors arising from outputs that do not conform to the required format, leading to parsing errors or incorrect actions. ``Repeating Actions'' is a common comprehensive error in LLMs, which refers to the phenomenon where the same token is repeated multiple times. If such output is successfully parsed, it results in a large number of repetitive actions.

As shown in Table \ref{tab:error-type}, we observed that open-source models exhibit a significant number of IA and RA type errors in the online setting, and Claude's performance is similar. This phenomenon is highly consistent with the online results presented in Table \ref{tab:model-noHK-eval}. Regarding RA type errors, some studies have pointed out that they are commonly found in scenarios where the training and testing domains do not match. This indicates that both the open-source models and Claude have deficiencies in training with the Vision-in-the-Chain type of data studied in this paper.

\begin{table}[htbp]
  \vspace{-5pt}
  \centering
  \small
  \caption{Error type summarization. ``IA'' stand for ``Invalid Actions'' and ``RA'' for ``Repeating Actions''. Both types of error are the reflection of instruction following failure. The values are presented in percentages (\%). Football results with mark $\dag$ mean these models sometimes present a trajectory only consists of long-pass, short-pass and high-pass, we assign this situation as ``RA'' too.}
    \begin{tabular}{c|c|c|c|c|c|c}
    \toprule
    \multirow{2}[4]{*}{Model} & \multicolumn{2}{c|}{Sokoban-G} & \multicolumn{2}{c|}{Sokoban-O} & \multicolumn{2}{c}{Football-O} \\
\cmidrule{2-7}          & \multicolumn{1}{c|}{IA} & \multicolumn{1}{c|}{RA} & \multicolumn{1}{c|}{IA} & \multicolumn{1}{c|}{RA} & \multicolumn{1}{c|}{IA} & \multicolumn{1}{c}{RA} \\
    \midrule
    GPT-4o &   0.0    &   5.0    &   0.0    &  6.7     &   0.0    &  7.5 \\
    Gemini &   15.0    &    3.3   &   25.0    &  0.0     &    0.0   & 3.0 \\
    Claude &    0.0   &   6.7    &   0.0    &     1.7  &   0.0    &  $43.0^\dag$ \\
    InternVL2 &   0.0    &   5.0    &   5.0    &   80.0    &    0.0   &    70.0 \\
    LLaVA-1.5 &    0.0   &   3.3    &    5.0   &  35.0     &    0.0   &  $25.0^\dag$ \\
    MiniCPM &   0.0    &    6.5   &   0.0    &    33.3   &   6.7    &  $50.0^\dag$  \\
    Phi-3-V &    0.0   &   0.0    &    0.0   &  23.3     &    0.0   &  24.5 \\
    \bottomrule
    \end{tabular}%
  \label{tab:error-type}%
  \vspace{-5pt}
\end{table}%

\begin{figure}[htbp]
    \centering
    \includegraphics[width=0.8\linewidth, trim=90 0 0 0, clip]{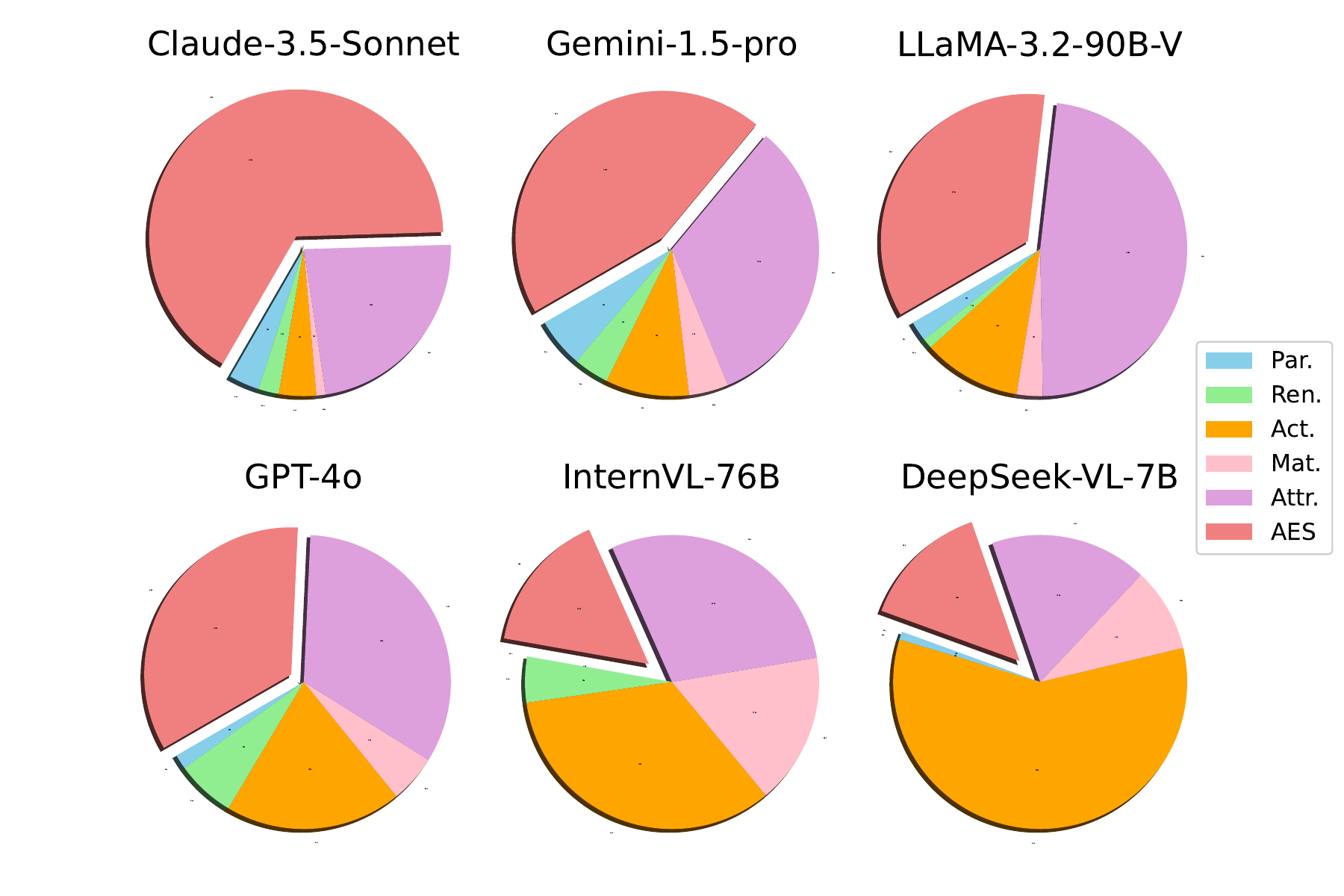}
    \caption{WebUI error construction. Each part of the pie graph is the corresponding score that model lost (or earned for AES part). ``Par.''$=$Parsing Error (or invalid actions); ``Ren.'' $=$ Rendering Error; ``Act.'' $=$ Webpage Interaction Error; ``Mat.'' $=$ HTML Tag Matching Error; ``Attr.'' $=$ Attribute Similarity Lost.}
    \label{fig:web-err}
    \vspace{-10pt}
\end{figure}

Figure \ref{fig:web-err} illustrates the composition of lost scores for different models on the WebUI. The AES section represents the scores obtained by the models. If a parsing error (unable to recognize code) or a render error (unable to render the initial web page due to some compilation issues) occurs, the model loses all the scores. If interaction error happened, the model loses the scores of the sub-webpage. The attribute similarity is only calculated when a model does not encounter ``Par.'', ``Ren.'', or ``Act.'' errors and successfully matches the corresponding HTML tags.

From Figure \ref{fig:web-err}, it can be observed that stronger models tend to have fewer ``Par.," ``Ren.," and ``Act." type errors (functional errors) caused by syntax and formatting issues, with a higher proportion of attribute setting errors instead. These models require stronger visual grounding capabilities to achieve further improvements. Conversely, weaker models have a higher proportion of functional errors, indicating their insufficient knowledge regarding web pages.

\section{Summarization}
In this paper, we introduce a new benchmark called MageBench, which is designed with a capability-oriented approach and includes three lightweight yet highly challenging environments. We conducted tests on a wide range of both open-source and close-source LMMs, with two baseline agent settings. The results indicate that current models lack ViC type reasoning abilities, as well as capabilities in cross-modal long context comprehension, visual imagination, and spatial planning. Our findings and the provided environments may inspire future research directions in the development of LMMs, which we discuss further in the Appendix. \ref{app:openquestion}. %
We hope to offer LMM developers valuable insights and optimization directions, and we will release the code and data as open-source in the near future.

\textbf{Limitations and future work.}  Currently, we have a limited number of environments. Additionally, to validate the model's capabilities, we use a standardized and simple agent setup. If this work garners attention within the community, we plan to substantially increase the number of environments in future work. This will allow us to explore technical details at the agent level, such as long-short term memory, retrieval-augmented generation, code involvement, and more. Furthermore, we could conduct studies on integrating LMMs with RL training within our environments.


\newpage
\appendix

\section{Dataset details}
\label{app:datasetdetails}
\subsection{WebUI}
\subsubsection{Web page collection and prepossessing}
\label{sec:intro-webui}

We searched and collected a large number of open-source, minimalist web design examples from GitHub. These small projects are often educational examples, and we filtered them to include only web projects composed of a small number of HTML, JavaScript, and CSS files.We ensure that the entire process of data collection, processing, application to our benchmark, and re-release as open source fully complies with the original open-source licenses of the GitHub repositories.

We then pre-process the projects with the following steps:
\begin{itemize}
    \item \textbf{Resource Download:} First, we download images and other resources from the webpage to a local folder. We then rename these files and correspondingly update the image address names in the webpage.
    \item \textbf{Content Adjustment:} We made several content adjustments to the web pages, including layout and attributes. Overly long web pages were split into multiple pages. Additionally, some web page content was modified to prevent existing models from having encountered these examples during data scraping on GitHub for training purposes.
    \item \textbf{Atomic Element Annotation:} In the website's HTML files (including parts of JavaScript that contain HTML code), we identify all atomic elements (as defined in the main part). For each corresponding HTML tag, we add two attributes: \emph{data-filter-by} and \emph{data-evalby}. The \emph{data-filter-by} attribute contains a single HTML or CSS property (such as the text property of a paragraph or the URL property of an image tag). This attribute, present in only a few elements, highlights the characteristics of the atomic element and is used to optimize the precision of atomic element matching. The \emph{data-evalby} attribute consists of multiple CSS properties (such as font, color, background, etc.). It displays the scoring criteria of the atomic element, and the evaluation of the element is based on the similarity of the CSS properties listed in \emph{data-evalby}. You can find an example of the annotated origin webpage in \ref{sec:webpage}.
    \item \textbf{Web Interaction Identification:} We identify and design meaningful web interactions (such as scrolling, clicking buttons, entering content, etc.) for each webpage. These interactions are implemented as Python functions using Google Chrome and the Selenium library. Additionally, we capture and store screenshots of the webpage after each interaction. An example interaction python code can be found here \ref{sec:actcode}.
    \item \textbf{Task Description:} We write a website reconstruction task description in markdown format of interleaved image-text based on the information of the website, which will include: 1. The function of the website. 2. Screenshots of the web page after the initial and each interaction. 3. The resources used in the website construction and the places where they are used (including image url, long texts, external CSS links, etc.). 4. Specify the class/id/name of the elements that can be interacted with, This will be used for automated interaction testing. An example of task description can be found in \ref{sec:webtaskdes}
\end{itemize}

\subsubsection{System prompt $p_{sys}$}
\label{sec:webuisys}
The system prompt for this environment is shown as follow:
\begin{verbatim}
Your job is to re-implement a Web page 
UI given target description and screen 
shot, by coding with `index.html`, and 
probably `style.css` and `script.js` 
if needed.
You will be provided a task description 
and available actions. Actions including 
writing to files and interact with 
your generated webpage.
Here are the action ID and explaination:

'write_html' - Input: str, Write the input 
str into 'index.html', if called second 
times, it will override the file.

'write_javascript' - Input: str, Write the 
input str into 'script.js', if called 
second times, it will override the file.

'write_css' - Input: str, Write the input 
str into 'style.css', if called second 
times, it will override the file.

\end{verbatim}

\subsubsection{Evaluation}
\label{app:webuieval}
\begin{figure*}
    \centering
    \includegraphics[width=\linewidth, trim=30 0 30 0, clip]{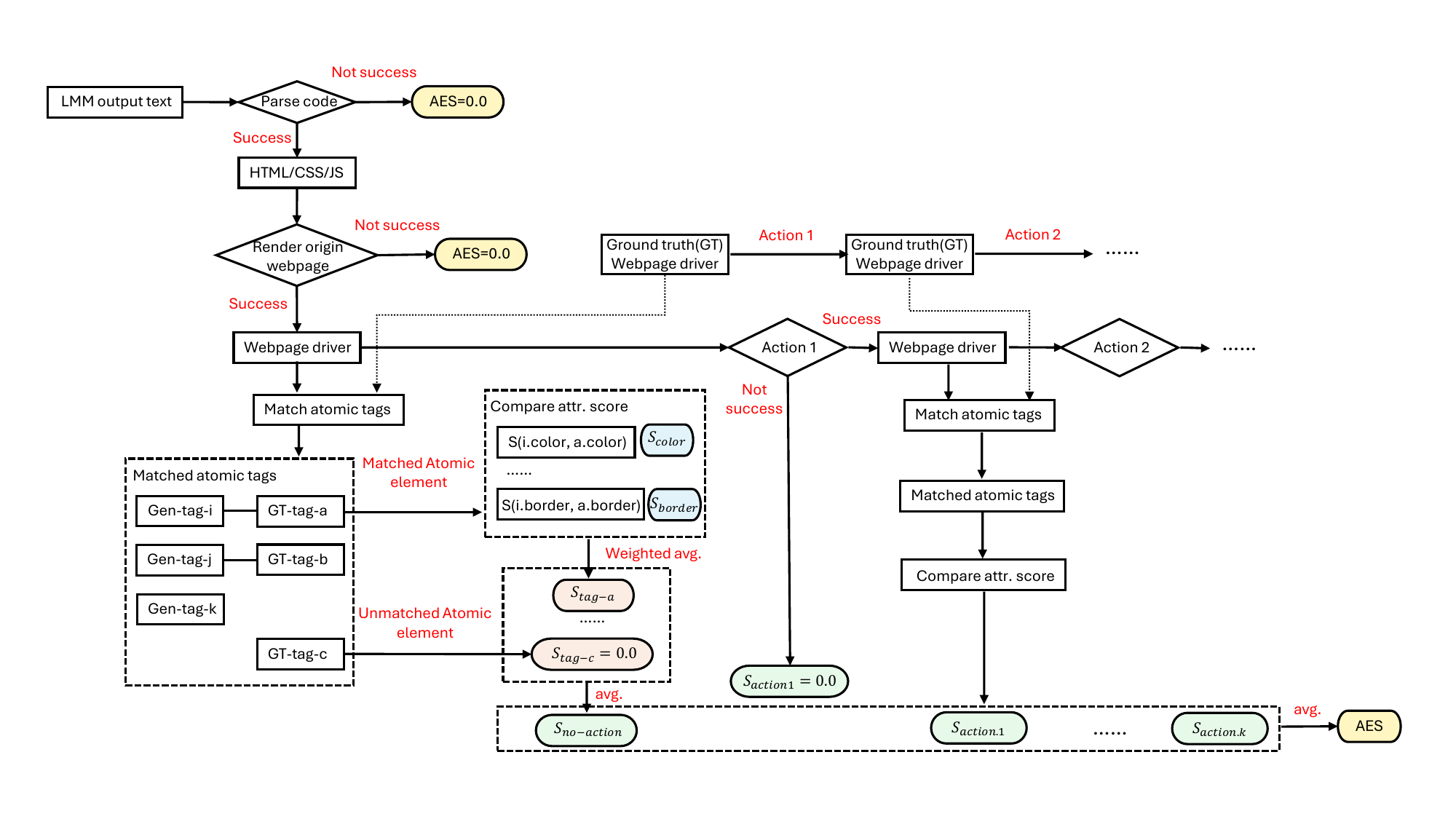}
    \caption{The complete evaluation pipeline of WebUI.}
    \label{fig:webui-eval-details}
\end{figure*}

The evaluation process of WebUI is illustrated in Figure \ref{fig:webui-eval-details}. In general, if there is an error in webpage parsing or initial page rendering, we directly assign a score of 0 to that webpage. When the webpage renders successfully, we treat the original page and each subsequent page after an interaction as independent static webpages for evaluation (excluding duplicated parts). For each static webpage, we perform two processes: atomic element matching and similarity calculation of the CSS properties of matched elements. Below, we will detail each part step by step.

\textbf{Atomic Tags Matching.} From a top-down perspective, atomic element matching is implemented using the Hungarian algorithm to establish a matching between the set 
$$
    \mathbb{E}_{gen}=\{e: \text{all elements in the generated webpage} \}
$$
and the set 
$$\mathbb{E}_{gt}=\{e: \text{all atomic elements in the GT webpage}\}.$$
To do this, we need to assign a score $S(e_i, e_j)$ where $e_i \in \mathbb{E}_{gen}, e_j \in \mathbb{E}_{gt}$ for the Hungarian algorithm to maximize. In our implementation, the definition of $S:\mathbb{E}_{gen}\times\mathbb{E}_{gt}\rightarrow \mathbb{R}$ is as follows:

\begin{equation}
\begin{aligned}
    S(e_i \in \mathbb{E}_{gen}, e_j \in \mathbb{E}_{gt}) :=& GIoU(e_i\text{.bbox}, e_j\text{.bbox}) \\
    &+ e_j\text{.filter}(e_i) \\
    &- \epsilon \left|e_i\text{.children} - e_j\text{.children} \right|. 
\end{aligned}
\label{eq:matching-score}
\end{equation}
GIoU is the standard generalized Intersection over Union that well-known in object detection. The GIoU value ranges from $[-1, 1]$ and is used to describe the similarity of the positions of the bounding boxes of two elements within a fixed-size frame. There is also a filter function to help better matching accuracy:
\begin{equation*}
e_j\text{.filter}(e_i)=\left\{ 
\begin{aligned}
-1&\qquad\text{if }L(e_i.\text{attr}, e_j.\text{attr})<\lambda_{attr}\\ 0& \qquad\qquad\text{else} 
\end{aligned}
\right. ,
\end{equation*}
where $attr$ is the pre-defined \emph{data-filter-by} attribute of the atomic element of the ground truth webpage, and $L$ is a general attribute similarity scoring function which will be introduced later. $\lambda_{attr}$ is a pre-defined fixed value for a kind of attribute. The attribute may be a CSS attrbute, or a html attribute. 

For example, let $e_j$ be a text tag and its \emph{data-filter-by} attribute is ``text'', and $e_j$ is:
\begin{lstlisting}
<!--this is ej-->
<span>example text tags</span>
\end{lstlisting}
Now suppose we have a generated webpage:
\begin{lstlisting}
<!--this is ek-->
<span>example not match</span>

<div> <!--div is et-->
    <!--this is ei-->
    <h1>example text tags</h1>
</div>
<!--this is ew-->
<span>example not match 1</span>
\end{lstlisting}
In this example, because:
$$L(e_k.\text{text}, e_j.\text{text}) = TIoU(e_k.\text{text}, e_j.\text{text})<\lambda_{attr},$$
where $TIoU$ is term IoU metric for text. So $e_j\text{.filter}(e_k)=-1$, which means wherever the element $e_k$ is incorrectly set, even its bounding box position coincide, it will not match with $e_j$. This filtered out a large amount of irrelevant element during the matching process.

The last term of equation \ref{eq:matching-score} is used to avoid this bug: using the same example above, you will find tag $e_t$ (the div container) has the same bounding box and text attribute with $e_i$ (the inner h1 tag). This is a property of html attribute. By adding the last term with a small punish coefficient $\epsilon=10^{-3}$, $e_j$ will match $e_i$ instead of $e_t$.

\textbf{Attribute Similarity Score. } We need to implement a $L: \mathcal{A}\times\mathcal{A}\rightarrow[0, 1]$, where $\mathcal{A}$ is all html and css attributes. We acturally implement this function according to the attribute types. 
For text attribute, we just use the term IoU.
For continuous variables, including \emph{border-width}, \emph{border-radius}, \emph{font-size}, \emph{height}, \emph{width}, \emph{letter-spacing} and etc, we just use the relative error:
$$L(e_j\text{.attr}, e_i\text{.attr})=\frac{|e_j\text{.attr}-e_i\text{.attr}|}{e_j\text{.attr}}.$$
For discrete variables with few choices, including \emph{background-image}, \emph{background-repeat}, \emph{src}, \emph{type}, \emph{border-style}, \emph{outline-style}, \emph{font-family}, \emph{font-style}, \emph{tab-size}, \emph{display} and etc, we use:
$$L(e_j\text{.attr}, e_i\text{.attr})=\mathbbm{1}(e_j\text{.attr}==e_i\text{.attr}).$$
For color attribute, including \emph{color}, \emph{background-color}, \emph{border-color}, \emph{outline-color} and etc, we use:
$$L(e_j\text{.attr}, e_i\text{.attr})=\frac{1}{3}\sum_{c\in \{r,g,b\}}\frac{1}{256}|e_j\text{.attr.c}- e_i\text{.attr.c}|.$$

\textbf{Hyper-parameters Searching.} As mentioned before, the final AES score can be writen as the average of each static webpage score (after certain actions, including scrolling, click):
$$S_{AES} = \sum_{act} S_{act}, $$
and $S_{act}$ is weighted averaged with pre-defined attribute weight and element space:
$$S_{act} = \sum_{e_i, e_j \in \text{matched}} \left( \frac{\sum_{at} L(e_j\text{.at}, e_i\text{.at}) \alpha_{at}}{\sum_{at}\alpha_{at}}\right) (e_j\text{.space})^\beta.$$
This means we still need a series of pre-defined hyper-parameters $\{\alpha\}$ and $\beta$, which are difficult to set manually, because it need to consider not only the importance of the attribute, but also the range of variation of the property. Instead, we first generate a large amount of webpages using varies models, to re-implement a certain webpage of our benchmark. Then human annotators rank those model implementations. We then used a CUDA-accelerated GPU version of Particle Swam Algorithm(PSO) to search the parameters to align with the human preference. As a result, the evaluation results with searched parameters agree 93\% of the human preferences, from 68\% with random parameters.

\subsection{Sokoban}
\label{app:sokoban}

\begin{figure}
    \centering
    \includegraphics[width=\linewidth, trim=300 60 300 60, clip]{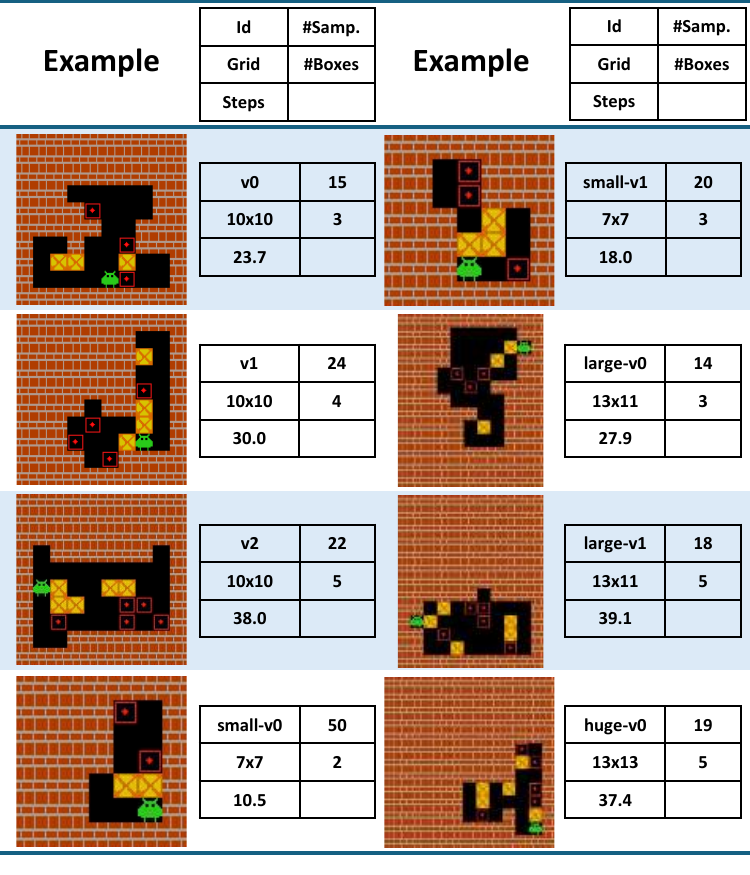}
    \caption{Details of different difficulties of the environments. `\#Samp.' refer to the number of the environments and `Steps' for the averaged minimum number of steps to fulfill the task.}
    \label{fig:sokoban-level-details}
\end{figure}

\subsubsection{Level generation}
\label{sec:sokobansys}
Based on the environment configurations provided by deepmind, we generated environments of varying difficulty levels. By considering the grid size of the environment and the number of boxes, we categorized all environments into eight difficulty levels, resulting in a total of 182 environments. Examples and detailed information for each difficulty level are shown in Fig. \ref{fig:sokoban-level-details}. We employed a pruned breadth-first search (BFS) algorithm to determine the minimum number of steps required to complete each environment, ensuring that all generated environments can be completed within fifty steps. This constraint is aimed at reducing the computational budget required for model invocation.
\subsubsection{System prompt $p_{sys}$}
The system prompt for this environment is shown as follow:
\begin{verbatim}
You're going to play a game of Sokoban, 
where the goal is to manipulate the green 
character to push the yellow box into 
the target area (an area with a red dot 
in the center). 
You'll use text commands to manipulate 
your character and restart the game, and 
here are the possible commands and 
corresponding meanings:

- 'Left': The action character moves to 
           the left, and if there is a
           box on the left, pushes the 
           box to the left.
- 'Right': The action character moves to 
           the right, and if there is 
           a box on the right, pushes 
           the box to the right.
- 'Up': Character moves up, and if 
        there is a box above, pushes 
        the box up.
- 'Down': The action character moves 
          down, and if there is a box 
          underneath, pushes the box down.


Note: 
1. It is not possible to push two boxes 
side by side at the same time from the 
side-by-side direction.
2. The red brick pattern is the wall. 
The game can go into a lose-lose state, 
for example if you push the box into a 
corner where the corner is not the target 
area. In this case, you should restart 
the game.

\end{verbatim}

\subsubsection{Evaluation}
Our evaluation metrics are adapted from those proposed in deepmind's paper. Specifically, for a given time step $t$ (assuming the task is not yet fully completed at this time), we allow the agent to take an action and define the reward at this time step as follows:
\begin{equation*}
R^{(t)}=\left\{
\begin{array}{rcl}
+4.5 & & \text{pushed box to target}\\
-5.5 & & \text{pushed box out of target}\\
+54.5 & & \text{task done}\\
-0.5 & & \text{otherwise}
\end{array} \right.,
\end{equation*}
which is exactly the same as the reward defined in their work except that we scale the value by $5\times$. At the same time, we do not want the model's output length (under the Global setting) or the number of steps set (under the Online setting) to affect the evaluation results. Therefore, we choose the metric of the historically best cumulative reward value, rather than using the cumulative reward value directly. Secondly, due to the inherent differences in the difficulty levels of various environments, the variances across different environments also differ. Directly calculating the mean of the rewards may cause environments with larger variances to dominate the final results. To address this issue, we subtract the reward value corresponding to the optimal trajectory (as provided by BFS) from the final evaluation of each environment. The final evaluation metric is shown as follow:
\begin{equation*}
    R = \max_{t=1}^{T} \left( \sum_{\tau=1}^{t} R^{(\tau)} \right) - R_{best} + 100,
\end{equation*}
where $R_{best}$ is the reward corresponding to the trajectory that achieves the task in the shortest possible steps, and $T$ is the lesser of the number of steps the model takes to successfully complete the task and 50. It is worth noting that, as mentioned earlier, all environments can be completed within 50 steps.

\subsection{Football}
\label{app:football}
\begin{figure}
    \centering
    \includegraphics[width=\linewidth, trim=0 40 0 0, clip]{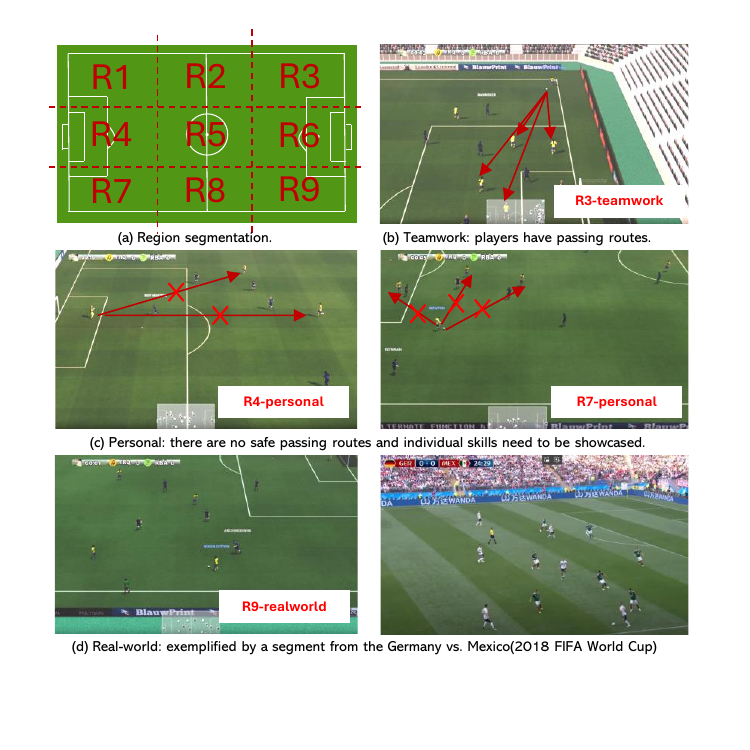}
    \caption{An overview of initial scene setting in the football environment.}
    \label{fig:football-level-details}
\end{figure}
\subsubsection{Initial scene generation.}
To enhance the diversity of tasks, we have designed different initial scenarios as levels. Each level operates under the same game rules but starts with different initial conditions (positions of players and the ball). To comprehensively demonstrate the model's capability to handle various situations, our initial scenarios are divided into three major categories: Personal, Teamwork, and Real-world. The Personal category represents scenarios where the ball handler does not have a clear passing route and needs to showcase individual skills to break through defenders, as illustrated in Figure \ref{fig:football-level-details}(c). In contrast, the Teamwork category represents scenarios where the positioning of players allows for effective collaboration and coordination, as illustrated in Figure \ref{fig:football-level-details}(a). To ensure the tasks are realistic, we have also extracted player positions from actual World Cup matches. This constitutes our Real-world category, as shown in Figure \ref{fig:football-level-details}(d). 

In addition to the aforementioned three major categories, we also evaluate the model's ability to handle the ball in different areas of the soccer field. As shown in Figure \ref{fig:football-level-details}(a), we have divided the soccer field into nine zones, and we will initialize various scenarios in each of these zones. In summary, we have three major categories and nine zones. For each category and zone, we will initialize four different scenarios, resulting in a total of $3 \times 9 \times 4 = 108$ unique scenarios. Table \ref{tab:real-world-cat} presents the specific World Cup matches and times for the Real-world category across all zones. These moments, which occurred in actual games, will be featured in our scenarios.

\begin{table}[htbp]
  \centering
  \scriptsize
  \caption{Details of 36 scenes in Real-world category. }
    \begin{tabular}{r|p{24em}}
    \toprule
    \multicolumn{1}{p{3em}|}{\textbf{Region }} & \textbf{Game; Time} \\
    \midrule
    \multicolumn{1}{r|}{\multirow{4}[2]{*}{R1}} & 1. Brazil vs. Belgium | 2018 FIFA World Cup; 69:32 \\
          & 2. Germany vs. Mexico | 2018 FIFA World Cup; 82:11 \\
          & 3. Brazil vs. Germany | 2014 FIFA World Cup; 12:19 \\
          & 4. Brazil vs. Germany | 2014 FIFA World Cup; 26:55 \\
    \midrule
    \multicolumn{1}{r|}{\multirow{4}[2]{*}{R2}} & 1. France vs. Croatia | 2018 FIFA World Cup Final;  46:53 \\
          & 2. Belgium vs. Japan | 2018 FIFA World Cup; 4:01 \\
          & 3. Germany vs. Mexico | 2018 FIFA World Cup; 53:08 \\
          & 4. France vs. Argentina | 2018 FIFA World Cup; 29:28 \\
    \midrule
    \multicolumn{1}{r|}{\multirow{4}[2]{*}{R3}} & 1. Brazil vs. Belgium | 2018 FIFA World Cup; 14:13 \\
          & 2. Brazil vs. Belgium | 2018 FIFA World Cup; 24:36 \\
          & 3. Germany vs. Mexico | 2018 FIFA World Cup; 13:11 \\
          & 4. France vs. Argentina | 2018 FIFA World Cup; 47:25 \\
    \midrule
    \multicolumn{1}{r|}{\multirow{4}[2]{*}{R4}} & 1. Germany vs. Mexico | 2018 FIFA World Cup; 62:58 \\
          & 2. France vs. Argentina | 2018 FIFA World Cup; 18:06 \\
          & 3. Portugal vs. Spain | 2018 FIFA World Cup; 82:02 \\
          & 4. Brazil vs. Germany | 2014 FIFA World Cup; 26:23 \\
    \midrule
    \multicolumn{1}{r|}{\multirow{4}[2]{*}{R5}} & 1. Brazil vs. Belgium | 2018 FIFA World Cup; 6:22 \\
          & 2. France vs. Argentina | 2018 FIFA World Cup; 42:08 \\
          & 3. France vs. Argentina | 2018 FIFA World Cup; 64:47 \\
          & 4. Portugal vs. Spain | 2018 FIFA World Cup; 31:49 \\
    \midrule
    \multicolumn{1}{r|}{\multirow{4}[2]{*}{R6}} & 1. Belgium vs. Japan | 2018 FIFA World Cup; 8:02 \\
          & 2. Portugal vs. Spain | 2018 FIFA World Cup; 62:08 \\
          & 3. Brazil vs. Germany | 2014 FIFA World Cup; 40:46 \\
          & 4. Netherlands vs. Brazil | 2010 FIFA World Cup; 16:55 \\
    \midrule
    \multicolumn{1}{r|}{\multirow{4}[2]{*}{R7}} & 1. Germany vs. Mexico | 2018 FIFA World Cup; 39:49 \\
          & 2. France vs. Argentina | 2018 FIFA World Cup; 00:52 \\
          & 3. Portugal vs. Spain | 2018 FIFA World Cup; 21:25 \\
          & 4. Portugal vs. Spain | 2018 FIFA World Cup; 64:22 \\
    \midrule
    \multicolumn{1}{r|}{\multirow{4}[2]{*}{R8}} & 1. France vs. Croatia | 2018 FIFA World Cup Final;  0:18 \\
          & 2. Brazil vs. Belgium | 2018 FIFA World Cup; 85:34 \\
          & 3. France vs. Argentina | 2018 FIFA World Cup; 61:53 \\
          & 4. Portugal vs. Spain | 2018 FIFA World Cup; 18:43 \\
    \midrule
    \multicolumn{1}{r|}{\multirow{4}[2]{*}{R9}} & 1. Belgium vs. Japan | 2018 FIFA World Cup; 20:53 \\
          & 2. Germany vs. Mexico | 2018 FIFA World Cup; 24:29 \\
          & 3. Portugal vs. Spain | 2018 FIFA World Cup; 26:31 \\
          & 4. Brazil vs. Germany | 2014 FIFA World Cup; 01:51 \\
    \bottomrule
    \end{tabular}%
  \label{tab:real-world-cat}%
\end{table}%

\subsubsection{System prompt $p_{sys}$}
\label{sec:footballsys}
The system prompt of football environment is shown as follow:
\begin{verbatim}
You're going to play a football game.
The character you control wears a 
yellow jersey with a blue name above 
his head.
Your goal is to attack the goal on the 
**right side** according to standard 
soccer rules. It's your teammates who
wear the yellow jersey, and the 
opponents in the other colors.
You'll use text commands to manipulate 
your character. Here are the action ID
and explaination:
- 'action_left', 'action_top_left', 
  'action_top', 'action_top_right', 
  'action_right', 
  'action_bottom_right', 
  'action_bottom', 
  'action_bottom_left': Run in the 
  direction corresponding to the 
  action ID's name.

- 'action_long_pass': perform a long 
  pass to the player on your team. 
  Player to pass the ball to is auto-
  determined based on the movement 
  direction.
  
- 'action_high_pass': perform a high
  pass, similar to action_long_pass.
  
- 'action_short_pass': perform a short
  pass, similar to action_long_pass.
  
- 'action_shot': perform a shot, 
  always in the direction of the 
  opponent's goal.
  
- 'action_sprint', 
  'action_release_sprint': start and
  stop sprinting, sticky action. If 
  sprinting, player moves faster, but 
  has worse ball handling.
  
- 'action_dribble', 
  'action_release_dribble': start and
  stop dribbling, sticky action. 
  Player moves slower, but it is 
  harder to take over the ball from 
  him.

\end{verbatim}

\subsubsection{Evaluation}
\label{app:footballeval}

The reward system is designed as follow:
\begin{equation}
\begin{aligned}
    R^{(t)} = &\lambda_1 S_{move}^{(t)} + \lambda_2 S_{oppo}^{(t)} + \lambda_3 \delta_{scored}^{(t)} \\
    &+ \lambda_4 \delta_{stole}^{(t)}\frac{t}{T} + \lambda_5 \delta_{pass}^{(t)}S_{pass}^{(t)}\\
    &+ \lambda_6\delta_{shot}^{(t)}S_{shot}^{(t)},
\end{aligned}
\end{equation}
where $\lambda_1=16, \lambda_2=20, \lambda_3=40, \lambda_4=20, \lambda_5=400, \lambda_6=100, T=400$ are fixed numbers. $\delta$ is an event indicator:
\begin{equation*}
\delta_{event}^{(t)} = \left\{ 
\begin{aligned}
    1&\qquad \text{if }event\text{ happened at time step }t \\
    0&\qquad \text{else}
\end{aligned}
\right..
\end{equation*}
Now, lets introduce each term separately.

\textbf{Advancement Reward $S_{move}^{(t)}$.} This metric is used to reward the agent for advancing the ball towards the opponent's goal, which includes actions such as individual dribbling breakthroughs and passing. We only consider this metric when the ball is under the control of our team's players. Therefore, it can be expressed as:
$$S_{move}^{(t)} = \delta_{own-ball}^{(t)}(ball_x^{(t)}-ball_x^{(t')}),$$
where $ball_x$ is the x coordinate of ball and $t'$ is the last time our team owned the ball. If one of our player is always controling the ball, $t'=t-1$. But if a player receive a ball at $t$ from his teammate, them $t'$ will be the frame that his teammate pass the ball.

\textbf{Passing Opponents Reward $S_{oppo}^{(t)}$.} This metric rewards the player for surpassing the opponent, incentivizing the ball to be positioned closer to the opponent's goal than the opponent's players:
$$
S_{oppo}^{(t)} = \frac{1}{11}(Passed^{(t)}-Passed^{(t')}),
$$
where $t'$ is the same above and
$$
Passed^{(\tau)} = \sum_{p \in opponents} \mathbbm{1}[p_x^{(\tau)} < ball_x^{(\tau)}].
$$

\textbf{Scoring Reward and Ball Interception Penalty.} If a goal is scored at this moment, the game will immediately end and a positive reward $\lambda_3$ will be received. If the ball is intercepted by the opponent (or goes out of bounds, or an offensive foul occurs), a reward linearly related to the moment of interception will be provided as $\lambda_4 \delta_{stole}^{(t)}\frac{t}{T}\in (0, \lambda_4]$. This means the later the player is intercepted, the larger the reward received. If the ball is not intercepted by the maximum time limit of $T=400$ frames, the player will receive the maximum reward $\lambda_4$.

\begin{figure}
    \centering
    \includegraphics[width=\linewidth, trim=150 190 150 120, clip]{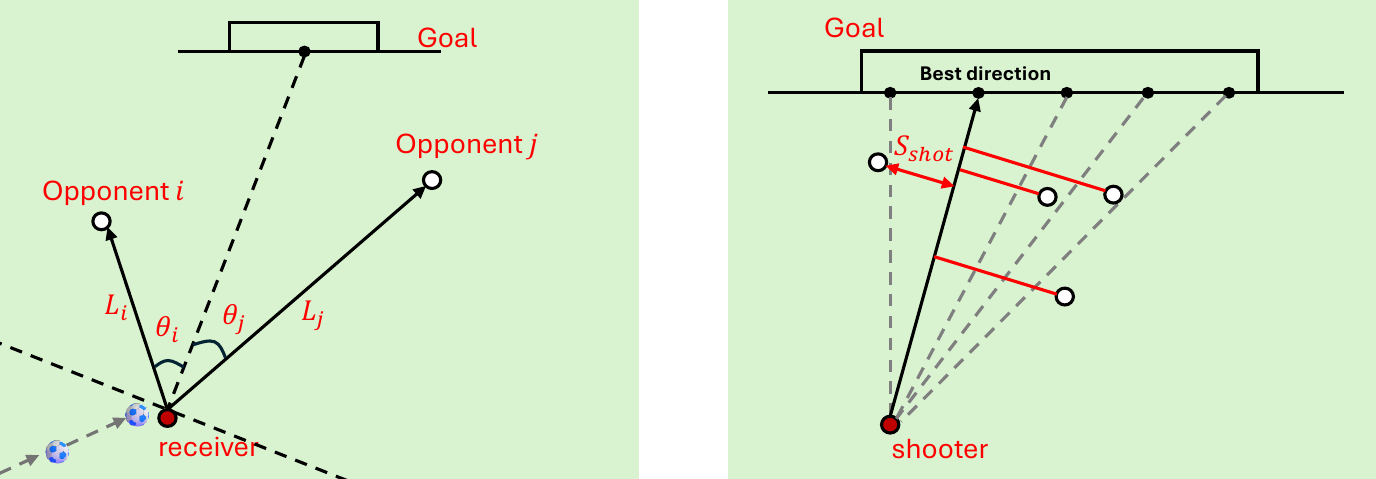}
    \caption{\textbf{Left:} Diagram illustrating the calculation of Passing Ball Reward. \textbf{Right: }Diagram illustrating the optimal shooting angle and opponent distance.}
    \label{fig:football-reward-details}
\end{figure}

\textbf{Passing Ball Reward $S_{pass}^{(t)}$.} We first note that $\delta_{pass}^{(t)}$ is the moment that the ball passing event ends (our team player receives a ball), instead of start passing. We only consider successful passes, as unsuccessful passes have already been penalized in the fourth term. At the moment when a teammate successfully receives the pass, we will provide a reward based on the distance of the nearest opponent to the receiver and the level of threat posed by the opponent. This incentivizes the model to pass the ball to players who are in more open and advantageous positions for breakthroughs. It is defined as follow:
$$
S_{pass}^{(t)} = \min_{i=1}^{11} \left[\frac{L_i}{\beta\max(\cos \theta_i, 0)+\epsilon}\right],
$$
where $\beta=10$ and $\epsilon=1$. As shown in Figure \ref{fig:football-reward-details} left, suppose the vector from the receiver to the opponent's goal is $\Vec{a}$. We define $L_i$ as the straight-line distance from the $i$-th opponent player to the receiver, and $\theta_i$ as the angle between the vector from the receiver to this opponent and vector $\Vec{a}$. We further interpret this metric as follows: First, we can easily observe the expression:
$$\frac{L}{\beta\max(\cos \theta, 0)+\epsilon} \in [\frac{L}{\beta +\epsilon}, \frac{L}{\epsilon}].$$
As $L$ increases, meaning the opponent is farther away from the receiver, the reward increases, indicating that the threat from the opponent is reduced. Conversely, as $\max(\cos \theta, 0)$ increases, it means the opponent's position is more obstructive to the player's path towards the goal, representing a higher threat. In this case, the reward decreases.

\textbf{Shot Rewards $S_{shot}^{(t)}$. } Given that scoring a goal is extremely challenging and carries a significant degree of randomness in the current model, we instead assign a score to a shot based on the quality of the opportunity at the moment the shot is taken. As shown in Figure \ref{fig:football-reward-details} right, we divide the goal into five target points for shooting. Assuming there are five shooting paths from the current player to these points, we first determine the optimal shooting path by calculating the distances of opponent players to these shooting paths. The optimal path is selected based on these distances, as indicated by the solid line in the Figure \ref{fig:football-reward-details} right. We then define $S_{shot}^{(t)}$ as the shortest distance from these opponent players to the chosen shooting path.

\begin{figure*}
    \centering
    \includegraphics[width=\linewidth, trim=75 150 90 120, clip]{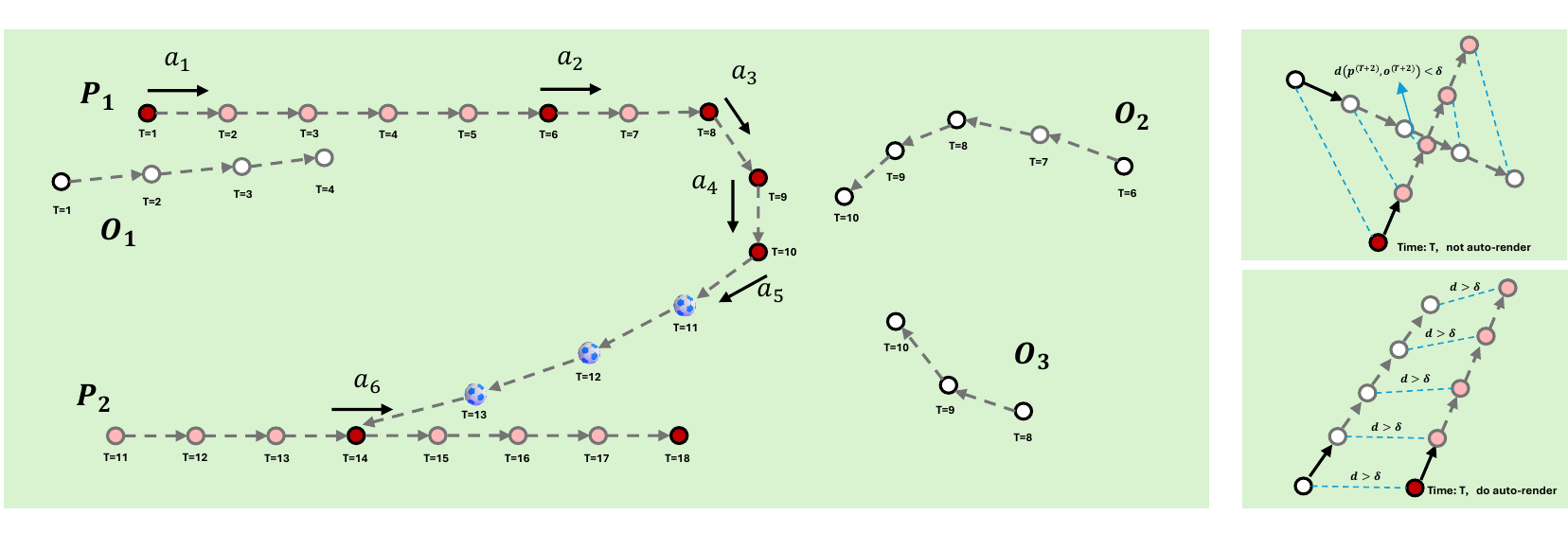}
    \caption{\textbf{Left: }Illustration of our auto-rendering algorithm. $P_1$ and $P_2$ represent the trajectories of two players on our team, while $O_1$, $O_2$, and $O_3$ represent the trajectories of three opponent players. All moments shown in light color indicate that they are automatically rendered, whereas the moments shown in dark color indicate the times when players need to make decisions. \textbf{Right:} Pre-rendering technology ensures that control is promptly returned to the player when the opponent's position and orientation pose a threat (\textbf{upper figure}), while reducing the amount of player control when there is no threat (\textbf{lower figure}).}
    \label{fig:football-auto-render}
\end{figure*}

\subsubsection{Auto-rendering algorithm.}
\label{app:auto-render}
Since we set the rendering frame rate to 400 frames in the simulated football environment, completing a single scenario requires 400 API calls. This results in significant time and cost expenditures for evaluation. In fact, each time an action is input and a frame is rendered, the characters in the game only move a very small distance, which implies that there will be a large number of repetitive actions in between. Our idea is to employ certain algorithms to automatically repeat these actions, allowing players to make decisions at critical moments. We have identified two scenarios where we can automate rendering to reduce redundant calls: 1. Passing sequences. 2. Straight-line running.

\textbf{Passing sequences. } According to the rules of the rendering environment, when a character initiates a pass or a shot, any control during the ball's time in the air is inherently ineffective, and the controlled player will switch during this process. Therefore, any control input during the entire duration when the ball is in the air is meaningless, and we will directly skip this period (corresponding to the process from $T=10$ to $T=13$ in the left panel of Figure \ref{fig:football-auto-render}). During this period, the player's action is defaulted to using the built-in AI.

\textbf{Straight-line running. } When a player inputs the same directional movement for two consecutive frames, we enable auto-rendering for straight-line running. The maximum number of frames for automatic rendering is set to 10 frames. For each of these 10 frames, we execute the following algorithm: assuming the current moment is $T$, we pre-render 5 frames based on the current positions and directions of all players. If, during these pre-rendered 5 frames, the distance between an opponent and the ball carrier is less than a preset threshold $\delta$, we will terminate the automatic rendering at moment $T$ and return control to the player. Otherwise, we will continue rendering in the current direction and speed until moment $T+1$.

\begin{algorithm}  
\small
\caption{Automatic Linear Running}  
\begin{algorithmic}[1]  
\REQUIRE player.pos, player.v, opponents
\ENSURE Automatic rendering of linear running  
  
\STATE max\_frames $\leftarrow 10$  
\STATE current\_frame $\leftarrow 0$  
\STATE threshold $\leftarrow \delta$  
  
\WHILE{current\_frame $<$ max\_frames}  
    \STATE collision\_detected $\leftarrow$ \textbf{false}  
    \FOR{$i \leftarrow 1$ to $5$}  
        \STATE player\_future $\leftarrow$ player.x $ + i \times $ player.v 
        \FORALL{oppo in opponents}
            \STATE oppo\_future $\leftarrow$ oppo.x $ + i \times $ oppo.v 
            \IF{d(player\_future, oppo\_future) $<$ threshold}
                \STATE collision\_detected $\leftarrow$ \textbf{true} 
                \STATE \textbf{break}
            \ENDIF  
        \ENDFOR  
        \IF{collision\_detected}  
            \STATE \textbf{break} 
        \ENDIF  
    \ENDFOR  
    \IF{collision\_detected}  
        \STATE \textbf{break}  
    \ENDIF  
    \STATE $\text{renderFrame}(T)$
    \STATE $T \leftarrow T + 1$  
    \STATE current\_frame $\leftarrow$  current\_frame $+ 1 $
\ENDWHILE  

\STATE \text{return control to player}  
  
\end{algorithmic}  
\label{alg:auto-rendering}
\end{algorithm} 

The details of the algorithm are presented in Pseudocode \ref{alg:auto-rendering}. In Figure \ref{fig:football-auto-render}, the automatic rendering corresponds to the periods from $T=1$ to $T=6$ and from $T=14$ to $T=18$. It is worth noting that within the maximum 10 frames of automatic rendering, the player is defaulted to move in a certain direction, which might prevent the player from passing the ball or performing other actions during this interval. However, in the rendered game visuals, the duration of 10 frames may only correspond to one or two steps for the player, covering a real-world distance of about two to three meters, which is still very short. Additionally, the algorithm ensures that there are no nearby opponents causing interference when the automatic rendering ends. Therefore, the impact can be considered negligible.

\begin{figure}
    \centering
    \includegraphics[width=\linewidth, trim=0 0 0 0, clip]{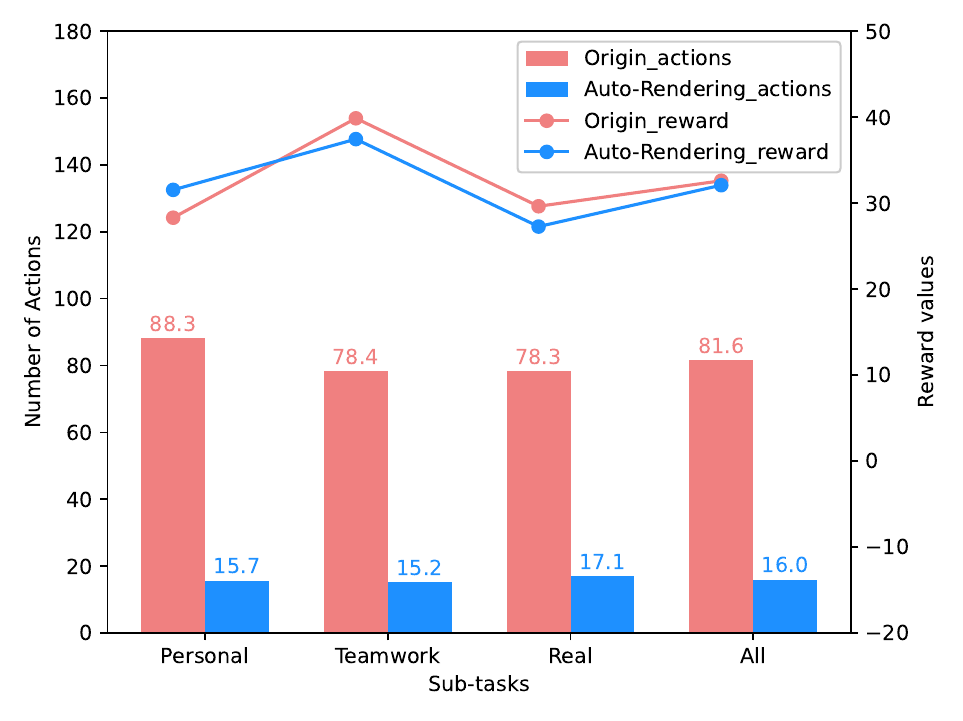}
    \caption{Effectiveness of auto-rendering technique. It reduce nearly 80\% of the actions but maintaining the similar evaluation results.}
    \label{fig:auto-render}
\end{figure}

We also conducted numerical experiments to demonstrate the effectiveness and correctness of our algorithm. As shown in Figure \ref{fig:auto-render}, we used an AI bot as the player. The line graph represents the rewards obtained by the player in different subsets, which serves as the evaluation result. The bar chart indicates the average number of operations per scenario required to complete all tasks (if the player is a model, this corresponds to the number of API calls). The red bars represent the results without using the aforementioned auto-rendering technique, while the blue bars represent the results with the technique applied. It is evident that, without significantly affecting the evaluation results, the use of the auto-rendering technique reduced the number of operations by nearly 80\%. This reduction lowered the average number of operations per scenario to 16, substantially saving testing costs for users of our benchmark.

\newpage

\subsection{Environment selection}
\label{app:env-select}
In this section, we will elaborate on the factors considered during our selection process for the environment and discuss the environments that were excluded from our study.

\textbf{Simplicity.} One of the critical requirements in evaluating environments is simplicity and ease of use. In the assessment of large language models (LLMs), tools like eval-harness enable the evaluation of dozens of datasets within a unified framework with minimal modifications, and their installation can be accomplished with a single command. Most LMM evaluations based on VQA datasets follow a similar pattern. However, this is not the case with existing agent environments, which are often highly complex or have significant hardware dependencies.

Many physical simulation platforms, such as Habitat, VirtualHome, and OmniGibson in the VisualAgentBench suite, have stringent requirements for GPU types and CUDA versions, often conflicting with the typically high-version torch and CUDA dependencies of LLM code. Some agent frameworks propose controlling UI operating systems such as Windows and Ubuntu, and software applications on these systems like office suites. These environments impose high demands on the platform, presenting prohibitive costs for researchers seeking to quickly validate model capabilities.
Additionally, certain environments are inherently large and costly to install. For instance, WebArena, VisualWebArena, and WebShop require complex installation processes and extensive web databases. Some environments aim for realism by running on large games via platforms like Steam, such as "Black Myth: Wukong," which can be tens or even hundreds of gigabytes in size, making them impractical.

Each of these unmet requirements can lead to slow environment simulation speeds, inability to perform parallel validations, and hinder rapid dissemination of findings.

\textbf{Visual Feedback.} We require that our environment must include visual feedback, and that such visual feedback is irreplaceable. In certain web-based environments, aside from images, the web pages themselves contain substantial information such as HTML and captions. Our observations indicate that tasks can be effectively accomplished using this information. In fact, in real-world scenarios, web automation testing and agent embedding tend to interact directly with HTML elements rather than relying on visual information as an intermediary. Consequently, we exclude such environments, along with those that do not necessitate visual participation, such as LLMAgentBench, from the scope of this paper.

\textbf{Representativeness on Reasoning.} For each environment we select, we must be able to clearly identify the required capabilities. This implies that the environment must have substantial planning space and the possibility for errors, with a high degree of freedom at both the high-level and the action-level. Most existing environments offer significant freedom at the action-level but lack substantial freedom at the high-level. We will illustrate this point with the following examples.

For instance, in our Sokoban environment, high-level planning primarily involves deciding which box to push to which area and the sequence in which to do so. In this game, pushing a box to the wrong area or in the incorrect sequence can result in a deadlock. In the context of Football, high-level planning refers to the various tactical strategies that can be employed.
In WebUI, the HTML part mainly constructs the layout of the webpage, while CSS determines attributes such as the color and font size of elements. At the high-level, HTML can have multiple layout configurations that achieve the same visual outcome, and the corresponding CSS properties may differ. This illustrates that the generation of HTML reflects the model's understanding of spatial layout during the structured generation process.

Contrary to our approach, some works impose limitations at the reasoning level. For example, in the CSS environment of VisualAgentBench, the HTML is fixed, and the CSS of the original webpage is modified, requiring the model to debug the corresponding CSS properties. Aside from a small amount of CSS knowledge, this task essentially becomes one of comparing images and identifying differences. Such tasks significantly undermine the generalization capability of structured visual generation, as we have demonstrated through qualitative experiments in Section \ref{sec:svg}.
Additionally, various physical simulation platforms like Ravens and VirtualHome also exhibit limitations in high-level planning space. For instance, tasks such as placing a red ball into a red bowl on a cluttered desk or retrieving a cloth from a drawer can fail due to the model's lack of common sense or perception errors. However, high-level planning tasks like "pick up the red ball, place it in the red bowl, ..." are quite naive and straightforward. They do not adequately reflect the model's planning and reasoning capabilities, primarily highlighting issues in perception and common sense.
Similarly, in web search tasks, the high-level planning typically follows a fixed process: "search for keywords, click the search button, observe the results, click the relevant item, ...". Errors in such tasks are often due to perception and decision-making mistakes, leaving little room for adjustment at the high-level planning stage.

In fact, beyond the three environments we propose, there are several others that meet our criteria. For instance, DetToolChain, which uses given tools to annotate data, as well as games like Sudoku, puzzle games, and various board games (such as Gomoku and Go), could all serve as robust evaluation environments. However, they may have similarities to those we have already introduced. If we can garner interest from the community, we intend to expand our project to include more such environments.
\begin{table*}[t!]
  \centering
  \small
  \caption{Evaluation on MageBench with unified prompt. }
    \begin{tabular}{l|c|c|c|c}
    \toprule
    \multirow{2}[4]{*}{model} & WebUI AES (\%) & \multicolumn{2}{c|}{Sokoban Reward} & Football Reward \\
\cmidrule{2-5}          & Global ($\delta=\pm0.8$)  & Global ($\delta=\pm0.5$) & Online ($\delta=\pm0.4$) & Online ($\delta=\pm2.0$) \\
    \midrule
    InternVL2-1B   & 10.61 &  46.01   &  46.06    &  -2.47 \\
     Xcomposer-2.5-1.8B &  16.14    &       45.76   &   45.60  
     &  IFE  \\
     InternVL2-2B   &  17.33     &   45.94  &   46.11   &  2.00 \\
     Qwen2-vl-2B   &   9.26   &  46.43   &  46.52    &  0.64    \\
     InternVL2-4B   & 2.92      &  46.32   &   46.28   &  -0.03 \\
    Phi-3.5-V-4.2B &    1.36         &     46.62  &   46.74    &   11.04 \\
    Yi-VL-6B   &   8.08   &  IFE  &    IFE   &  IFE  \\
    DeepSeek-VL-7B &    18.22    &     45.60     &   46.64   &  IFE  \\
     Xcomposer-2.5-7B &    15.32    &   46.84       &      46.39  
     &  IFE  \\
     LLaVA-v1.5-7B  &   11.42   &   45.60    &  46.19  &  -0.80 \\
     Llava-1.6-mistral-7B   &  9.07    &   46.37    &  46.23  &  -3.64 \\
     Qwen2-vl-7B   &    11.98   & 46.38   &   46.41   &    2.68  \\
    MiniCPM-V2.6-8B &   16.06       &   46.53  &   46.86   &  3.61 \\
    InternVL2-8B   &   27.43    &  46.56   &   46.35   &  7.90   \\
    Llama-3.2-11B-Vision-Instruct   &  39.22   &  46.54    &   47.33   &  11.58  \\
    LLaVA-v1.5-13B  &   13.30     &    46.65   &   46.53    &  16.59  \\
    InternVL2-26B   &  31.68    &  46.77   &   46.50   &  15.62 \\
    Llava-1.6-34B   &     1.65    &     46.06  &  46.48   &  3.04 \\
    Yi-VL-34B   &    7.21    &  IFE    &   IFE    &  IFE  \\
    InternVL2-40B   &    7.20   &   45.65  &  46.82   &  12.99 \\
    Qwen2-vl-72B   &      11.70   &  47.14    &   47.84   &   15.02   \\
    NVLM-72B   &   10.36      &   46.39   &    46.68   &   18.43 \\
    InternVL2-76B-LLaMA3   & 28.45       &   46.61   &   47.09   &  8.04   \\
    Llama-3.2-90B-Vision-Instruct   & 42.21     &   47.14    &    IFE  &  8.32  \\
    \midrule
    GPT-4o &   40.02   &   47.46    &  \textbf{ 48.63}    &  17.78 \\
    Gemini-1.5-pro &   \textbf{52.10 } &    \textbf{47.65}  &   48.33    &  \textbf{19.61} \\
    \midrule
    Idle Baseline & 0.00  &  45.60  &  45.60  &   2.36 \\
    Random Baseline &  0.00   &  47.40  &  47.40   &   16.64 \\
    \bottomrule
    \end{tabular}%
  \label{tab:more-model}%
  \vspace{-10pt}
\end{table*}%

\section{More Results}

\subsection{MageBench results}
\label{app:more-result}
In Table \ref{tab:more-model}, we present the test results of additional models on the larger MageBench full set. We selected models of varying sizes from 14 different model families, totaling 26 models. Given the large number of scenarios, each model and scenario only needed to be tested three times to compute the mean, resulting in a very small variance. The related conclusions are similar to those discussed in the main part, so they will not be reiterated here. Notably, we did not measure the results for Claude-3.5-Sonnet and human participants, as we were unable to bear the associated economic costs for the full dataset. Additionally, as mentioned in the main text, the Online setting in the current WebUI environment is essentially non-functional. Consequently, we did not allocate resources to testing in this setting.

\subsection{Agent memory}
\label{app:agentmem}
We selected the most robust model 
in the Sokoban environment under the online setting, 
specifically GPT-4o, to conduct memory ablation experiments.
In this context, ``AM" refers to action memory, which denotes the number of historical actions, while ``OM" signifies observation memory, representing the number of observations inputted for generating the next action. Here, observations refer to the game's visual frames. We have chosen AM values of 1, 5, and 10, and OM values of 1, 2, and 3 for our combinations. For instance, when AM=10 and OM=2, the model's input and output can be described as follows (For the sake of brevity, we have omitted $p_{sys}, p_{cot}, p_{io}$.):
\begin{equation*}
    \pi_\theta(\mathbf{a_{t-9}}, ..., \mathbf{a_{t-2}}, \mathbf{a_{t-1}}, \mathbf{o_{t-1}}, \mathbf{a_{t}}, \mathbf{o_{t}})\rightarrow \mathbf{a_{t+1}}.
\end{equation*}

\begin{table}[htbp]
  \centering
  \vspace{-10pt}
  \caption{Ablation on memory length.}
  \vspace{-5pt}
    \begin{tabular}{|c|c|c|c|}
    \toprule
    reward & AM=1  & AM=5  & AM=10 \\
    \midrule
    OM=1  & 53.508 & 53.417 & 52.458 \\
    \midrule
    OM=2  &   -    & 52.242 & 54.350 \\
    \midrule
    OM=3  &    -   & 53.392 & 53.325 \\
    \bottomrule
    \end{tabular}%
  \label{tab:memabl}%
  \vspace{-5pt}
\end{table}%

In our expectations, as the context length increases, the model, having access to more preceding information, should be able to perform the task better. However, in reality, we have not observed a stable correlation between context length and performance, as shown in Table \ref{tab:memabl}. The values presented in the table are all within the range of variance, thus it cannot be concluded that the impact is stable. This result indicates that the model's performance, based solely on the previous frame and action, does not differ from its performance when more memory is added. This is inconsistent with human behavior. It suggests that the LMM still has deficiencies in handling long text-visual context tasks.

\section{Benchmark generalization capability}

Given that our benchmark is ability-oriented, we aim for the knowledge it provides to have a certain degree of generalizability and to offer insights for future engineering developments. To illustrate this point, we conducted some qualitative experiments to demonstrate that our metrics continue to generalize well when the model is transferred to other similar tasks with engineering significance. Our explanation is divided into two parts: the first part addresses how ViC-type reasoning, as indicated by Sokoban-Online and Football-Online metrics, generalizes to potential applications in robotics and embodied AI. The second part discusses how WebUI generalizes across various structured visual generation tasks. We will elaborate on each part below.

\subsection{Robotics and embodied AI}

We hope to assert that our ViC-type reasoning could be significant in fields such as robotics. However, real-world mechanical arm environments are extremely costly, and the generalizability across different mechanical systems is often poor. In virtual physical environments, the input for mechanical arms typically consists of numerical values like torque and angles, and it is evidently unrealistic for LMMs to generate these values directly. One potential alternative is for LMMs to generate only high-level planning, with the execution of actions relying on predefined action models such as PDDL. However, this approach does not effectively evaluate the planning capabilities of LMMs. This is because the predefined actions are very limited, and if the actor model makes an error, LMM has no means to correct it.

\begin{figure}
    \centering
    \includegraphics[width=\linewidth,]{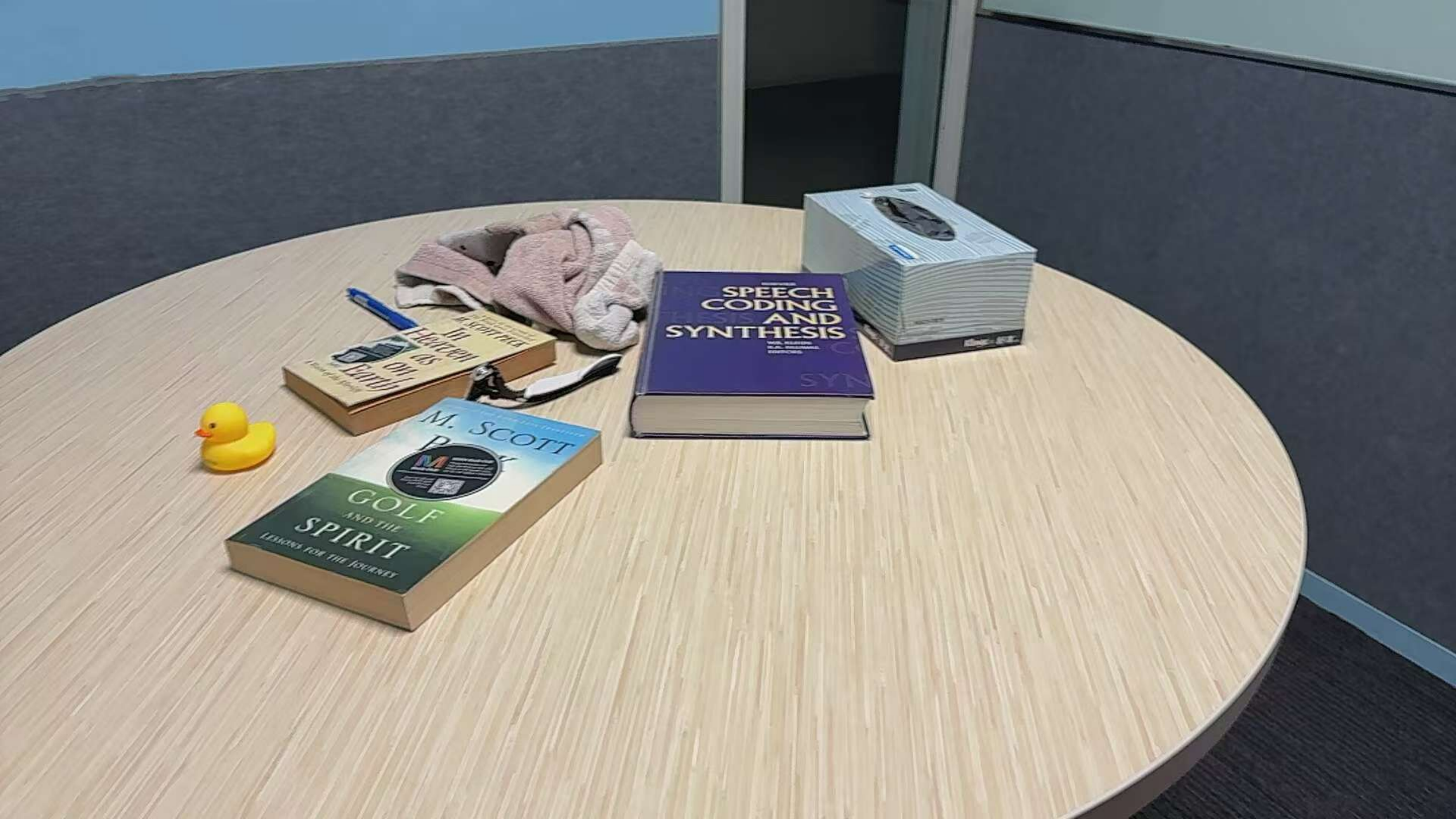}
    \caption{The initial scene for Clean-up table task, which is carefully replicated for each model before testing. This replication is done manually to ensure the scene is as consistent as possible across different tests.}
    \label{fig:clean-init}
\end{figure}

To perfectly decouple planning and control while obtaining a rich planning space, we have designed a human-in-the-loop testing method. Simply put, this involves using humans as robots, with the LMM providing detailed natural language instructions for actions. This way, humans can accurately execute the actions as directed by the LMM. Consequently, the quality of the final outcome is solely determined by the effectiveness of the LMM's planning. We have designed two tasks for LMM in real-world scenarios:

\textbf{Clean-up table.} As shown in Figure \ref{fig:clean-init}, we first replicate the scene depicted in the image each time. This scene features a cluttered desk with various objects scattered about. The LMM is required to provide instructions to the robot to organize the desk so that it appears clean and tidy. The specific details of the task are described as follows.

\begin{verbatim}
Task: Your job is to manipulate the
items on the table, put all the items
orderly on the table and make the 
table looks good.

You cannot put anything out of the 
table.
\end{verbatim}

\begin{figure}
    \centering
    \includegraphics[width=\linewidth, trim=60 320 80 50, clip]{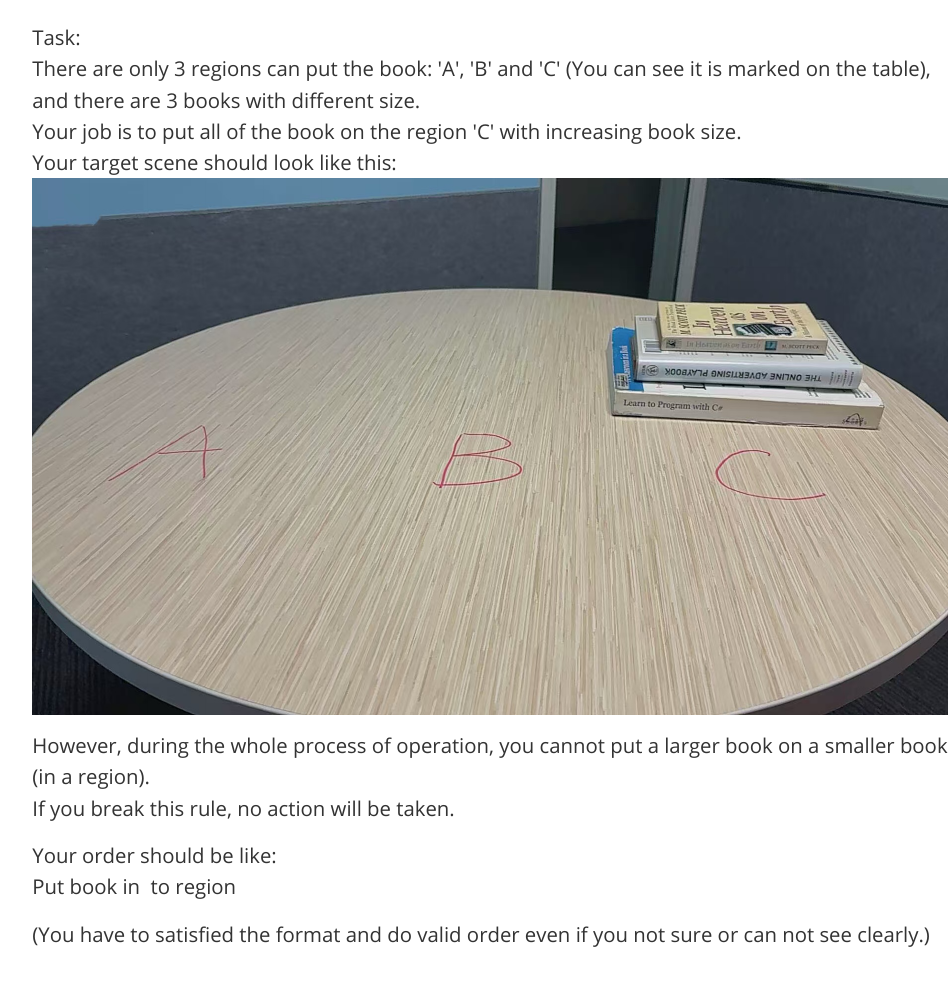}
    \caption{Task description for Book Hanoi tower task.}
    \label{fig:hanoitower-task}
\end{figure}

\begin{figure*}
    \centering
    \includegraphics[width=\linewidth, trim=150 110 130 100, clip]{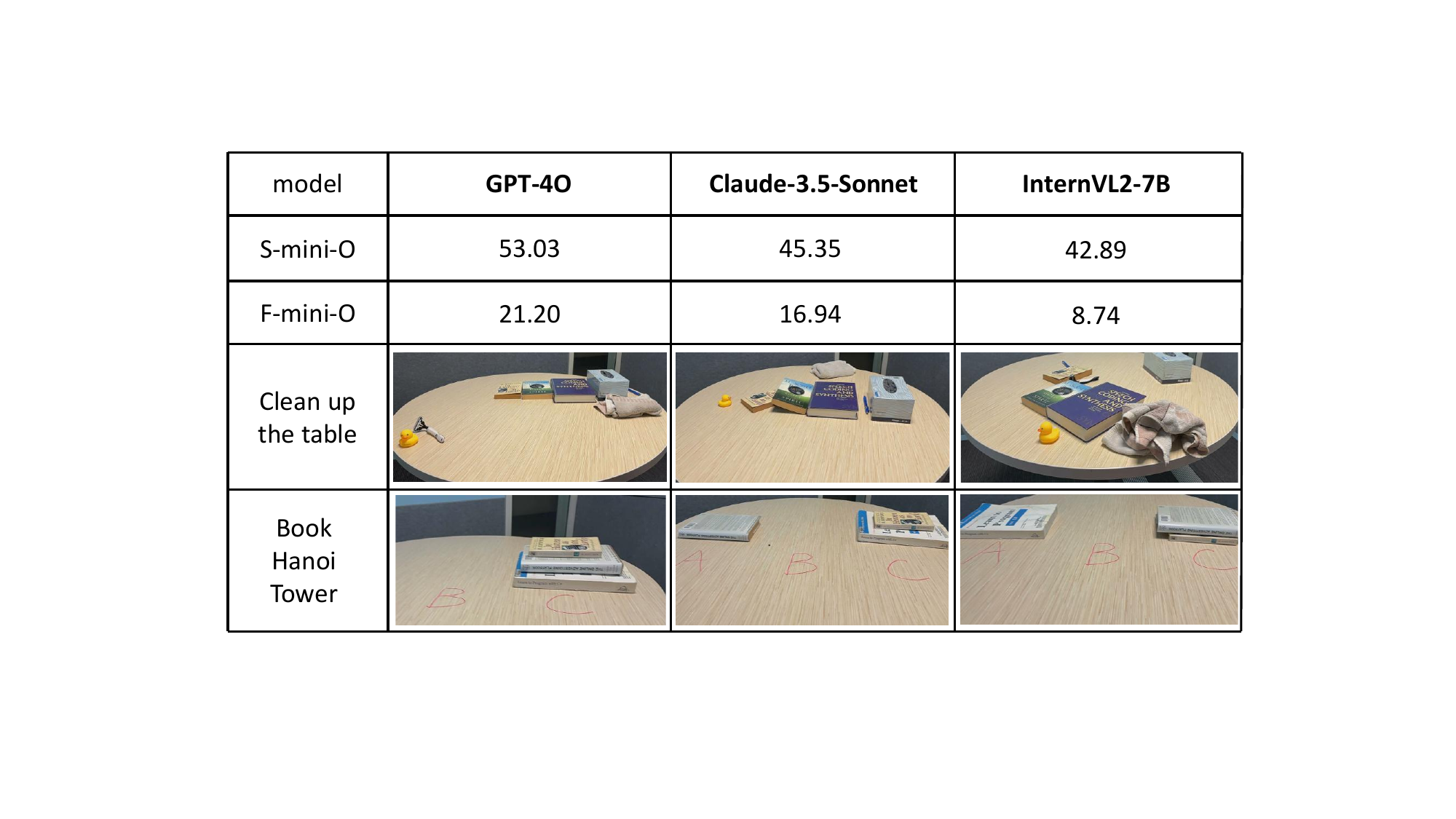}
    \caption{The Sokoban-Online, Football-Online result on MageBench-mini of the 3 models, and the last frame of their performance on our real-world planning evaluation.}
    \label{fig:robot-overall}
    \vspace{-10pt}
\end{figure*}

\begin{figure}
    \centering
    \includegraphics[width=\linewidth,]{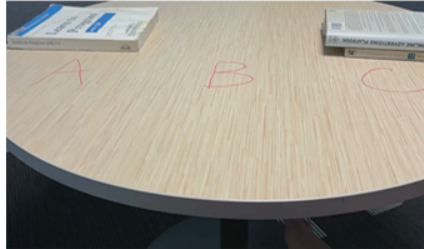}
    \caption{Initial scene for Book Hanoi Tower task.}
    \label{fig:hanoitower-init}
\end{figure}

\textbf{Book Hanoi Tower.} Similar to the Tower of Hanoi game, we divided the desk into three areas. There are a total of three books of different sizes. The model needs to ultimately stack the three books in area ``C'' in order from smallest to largest. Apart from the initial setup, during the moving process, larger books cannot be stacked on top of smaller books. Figure \ref{fig:hanoitower-task} illustrates the task description provided to the model, while Figure \ref{fig:hanoitower-init} shows the initial setup. This initial scene is specifically designed to prevent the model from being overly familiar with the Tower of Hanoi algorithm.

\subsubsection{Overall results}
In Figure \ref{fig:robot-overall}, we directly present the performance of three models, \emph{i.e.,} GPT-4o, Claude-3.5-Sonnet, and InternVL2-7B, under the two newly constructed tasks. The figure illustrates the final environmental scenarios for these tasks. Overall, it can be observed that the planning capabilities in real-world scenarios are highly correlated with the test scores from our MageBench.

GPT-4o successfully achieved the preset objectives in both scenarios, albeit with a number of seemingly meaningless actions in between. In contrast, Claude demonstrated some effective actions but often got stuck at a certain point within a scenario. As discussed in our paper, open-source models still have a considerable gap to close.

The aforementioned results indicate that ViC-type reasoning is indeed representative. In the next section's visualizations, we will also observe that common errors in Agent environments are similarly prevalent in robotic planning. This suggests that our understanding of their error types is generalizable.
\subsubsection{Visualization}
\begin{figure*}
    \centering
    \vspace{-20pt}
    \includegraphics[width=0.95\linewidth, trim=50 160 50 0, clip]{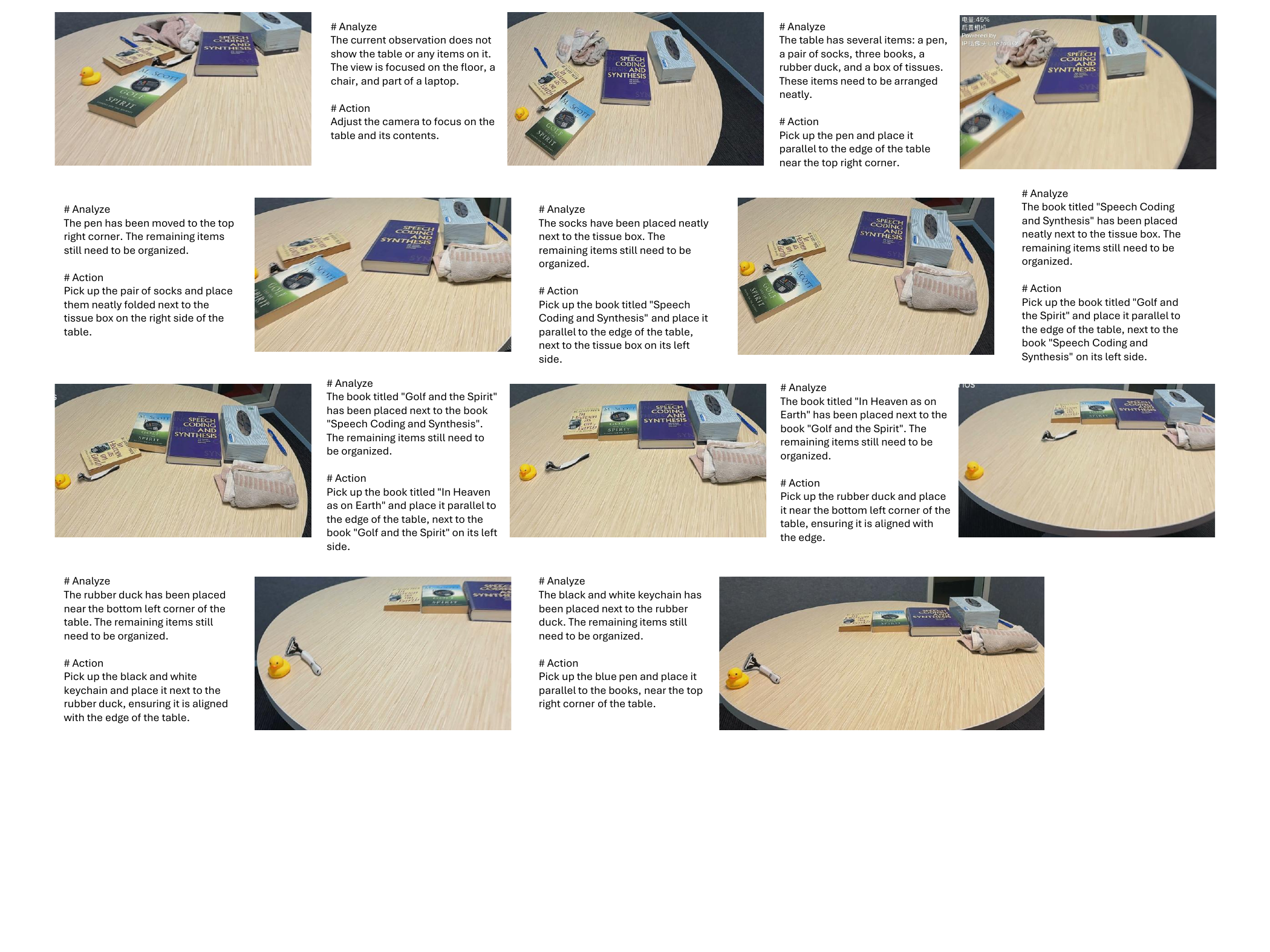}
    \vspace{-10pt}
    \caption{GPT-4o on Clean-up table.}
    \label{fig:g4oclean}
    \vspace{-10pt}
\end{figure*}

\begin{figure*}
    \centering
    \includegraphics[width=0.95\linewidth, trim=50 160 50 0, clip]{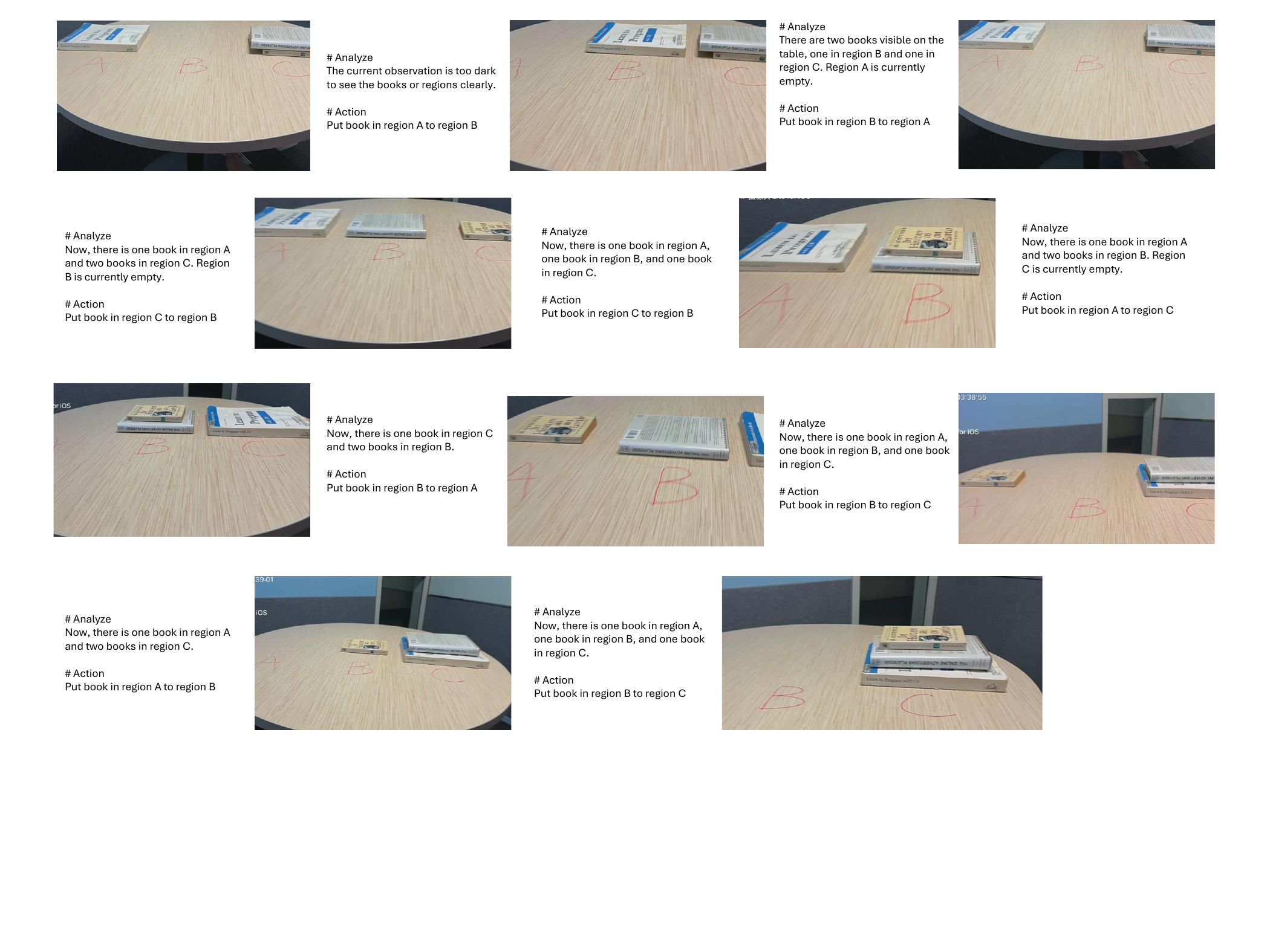}
    \vspace{-10pt}
    \caption{GPT-4o on Book Hanoi Tower.}
    \label{fig:g4ohanoi}
    \vspace{-10pt}
\end{figure*}

Figures \ref{fig:g4oclean} and \ref{fig:g4ohanoi} illustrate the outputs of GPT-4o and the environment feedback across two tasks, encapsulating our proposed Vision-in-the-Chain (ViC) framework. It is evident that GPT-4o exhibits a significant number of repetitive and meaningless actions in the Book Hanoi Tower task. Furthermore, other models, such as InternVL2-7B, predominantly demonstrate errors characterized by the initiation of repetitive and identical analyses and actions after performing two or three initial actions correctly. This pattern aligns with the error types analyzed in MageBench, suggesting that due to the lack of ViC-type and multi-turn chat data, these models encounter instruction-following failures when faced with tasks in an online setting.

\begin{figure*}[htbp]
	\centering
       \vspace{-20pt}
	\begin{minipage}{0.33\linewidth}
		\centering
		\includegraphics[width=\linewidth, trim=0 0 0 0, clip]{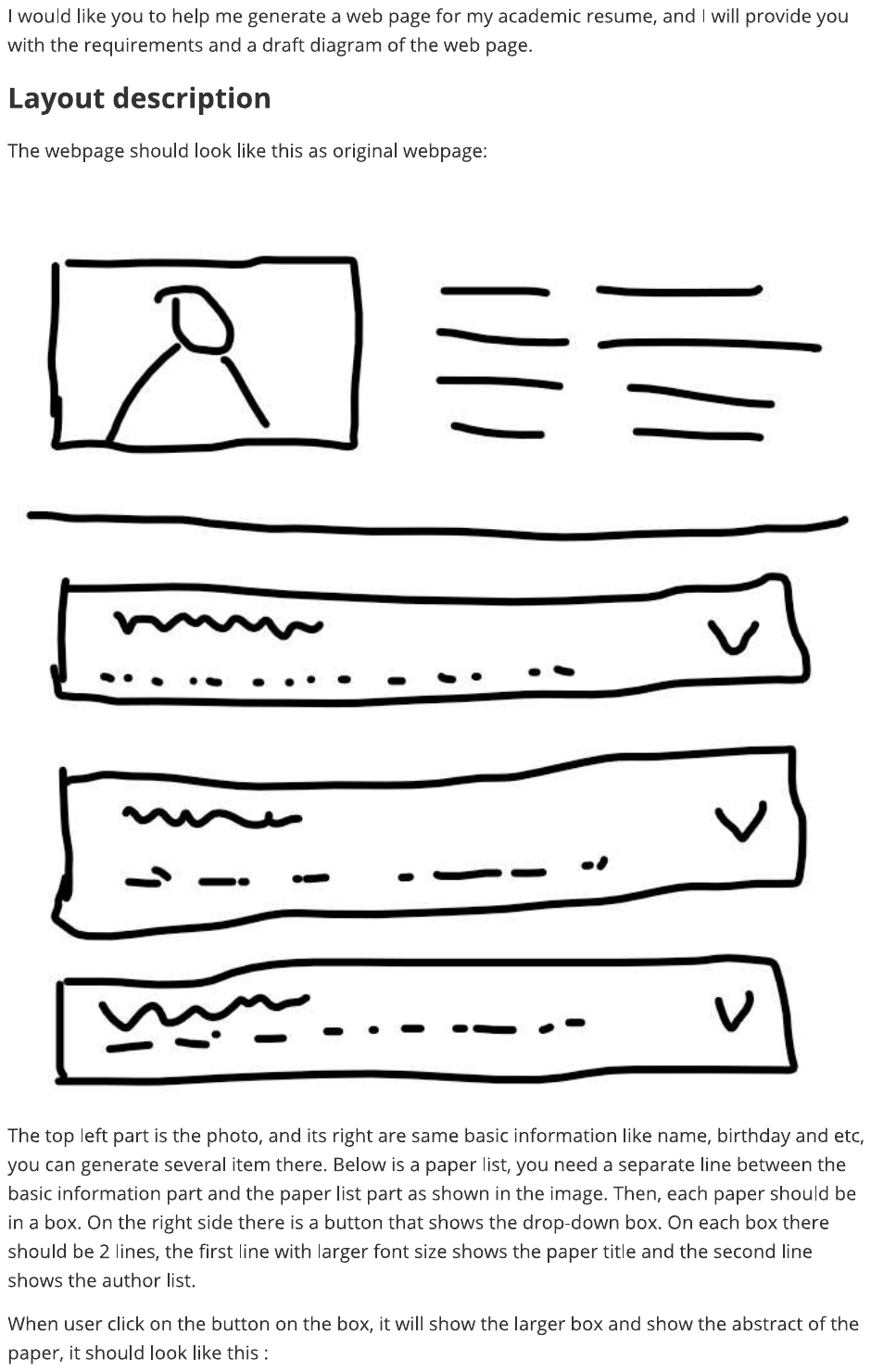}
	\end{minipage}
	\begin{minipage}{0.33\linewidth}
		\centering
		\includegraphics[width=\linewidth, trim=0 0 0 0, clip]{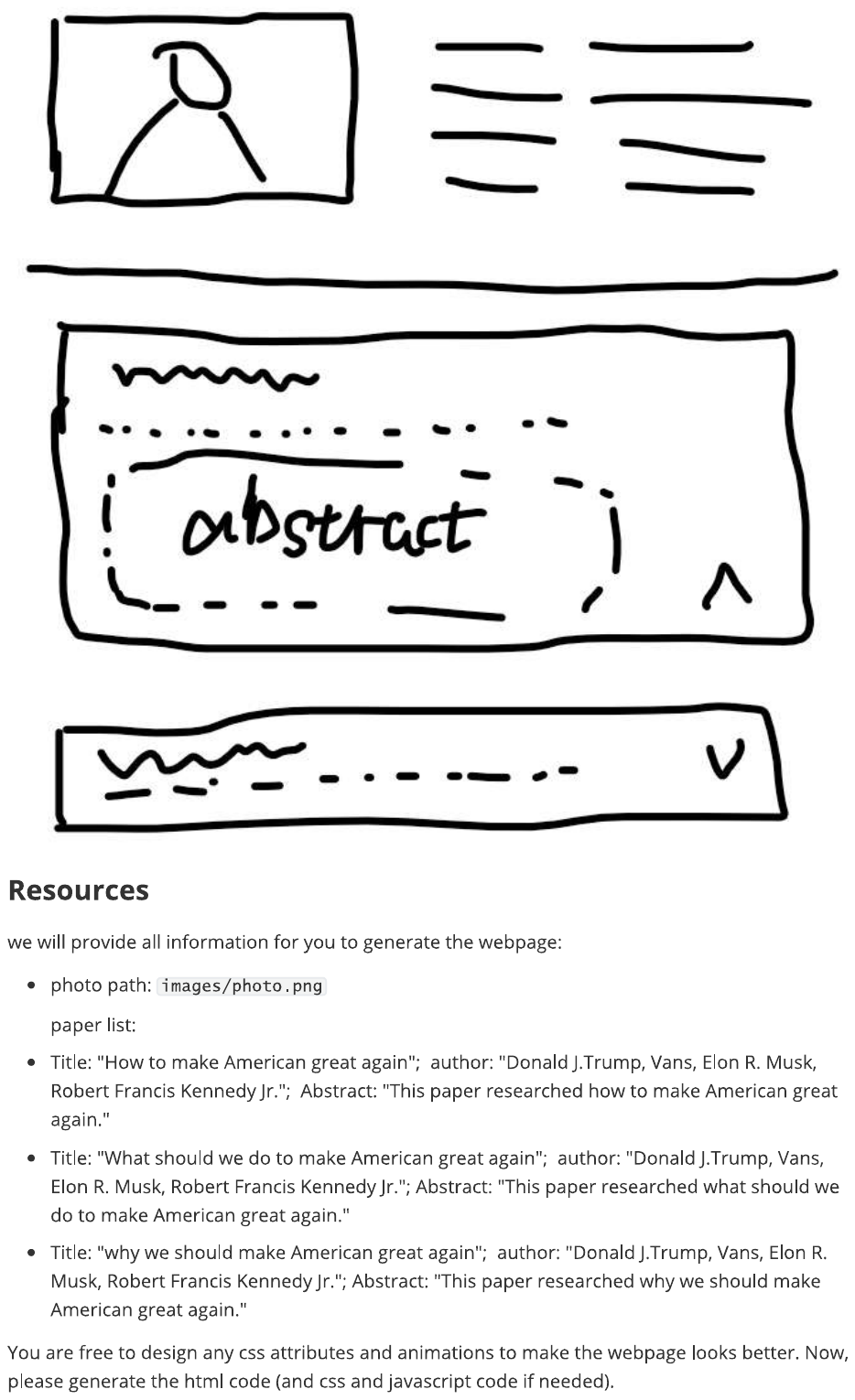}
	\end{minipage}
    \begin{minipage}{0.33\linewidth}
		\centering
		\includegraphics[width=\linewidth, trim=0 0 0 0, clip]{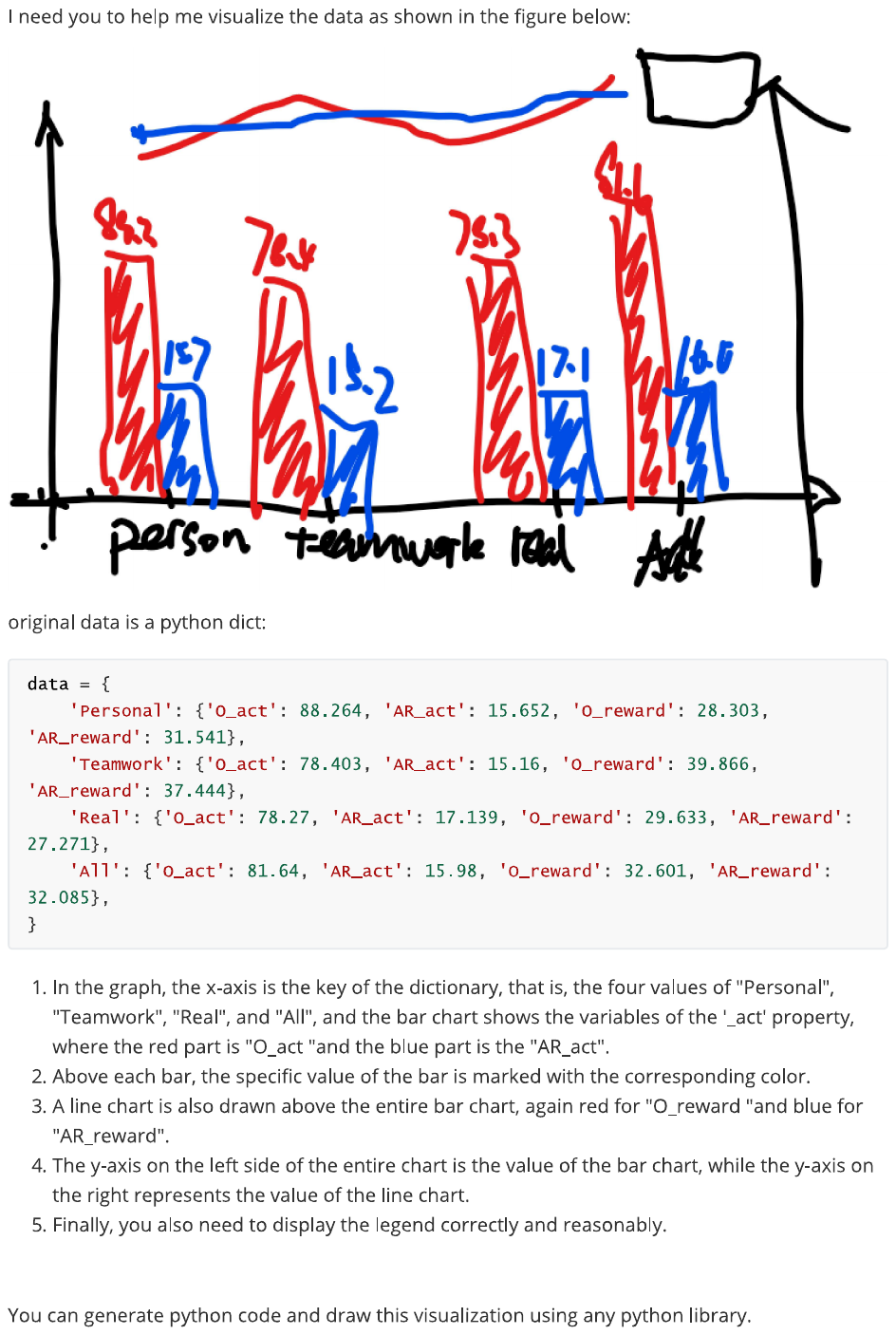}
	\end{minipage}
    \vspace{-10pt}
    \caption{Task description for webpage generation (left and middle) and data visualization (right).}
    \vspace{-10pt}
    \label{fig:oodtasks}
\end{figure*}

\begin{figure}
    \vspace{-10pt}
    \centering
    \includegraphics[width=\linewidth,trim=0 450 0 0, clip]{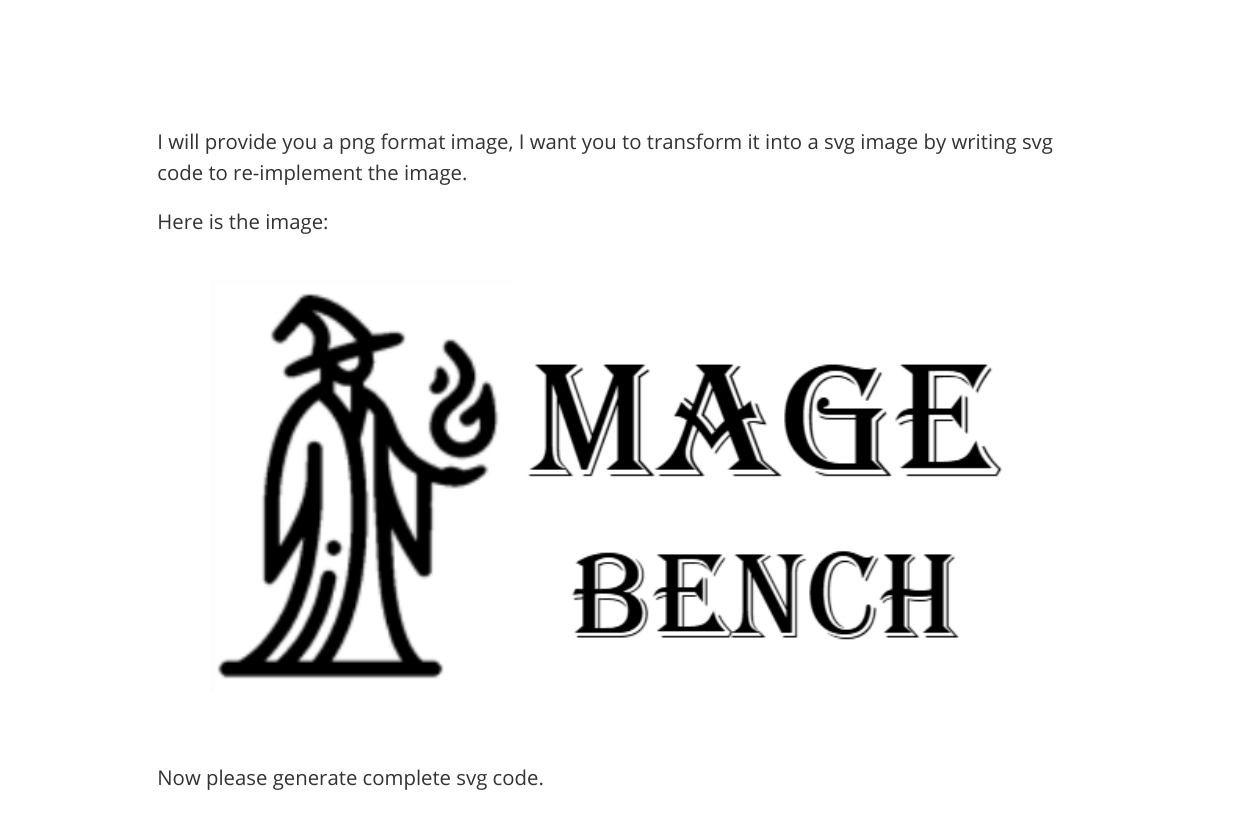}
    \caption{Task description for image to SVG task.}
    \label{fig:svg-td}
    \vspace{-20pt}
\end{figure}

\subsection{Structured Visual Generation}
\label{sec:svg}
In MageBench, our WebUI environment is designed to reconstruct a target website. Our goal is to assess whether this testing environment can accurately reflect the model's capability in structured visual generation, with an eye towards future applications in engineering practice. To this end, we have devised three structured visual generation tasks:
\begin{itemize}
    \item Generate a webpage based on hand-drawn sketches and descriptions.
    \item Generate SVG images (code) from observed pictures.
    \item Create Python code for data visualization.
\end{itemize}
The specific task descriptions for these three tasks, which serve as the model inputs, are presented in markdown format in Figures \ref{fig:oodtasks} and \ref{fig:svg-td}.

\begin{figure*}
    \centering
    \includegraphics[width=\linewidth, trim=160 0 130 0, clip]{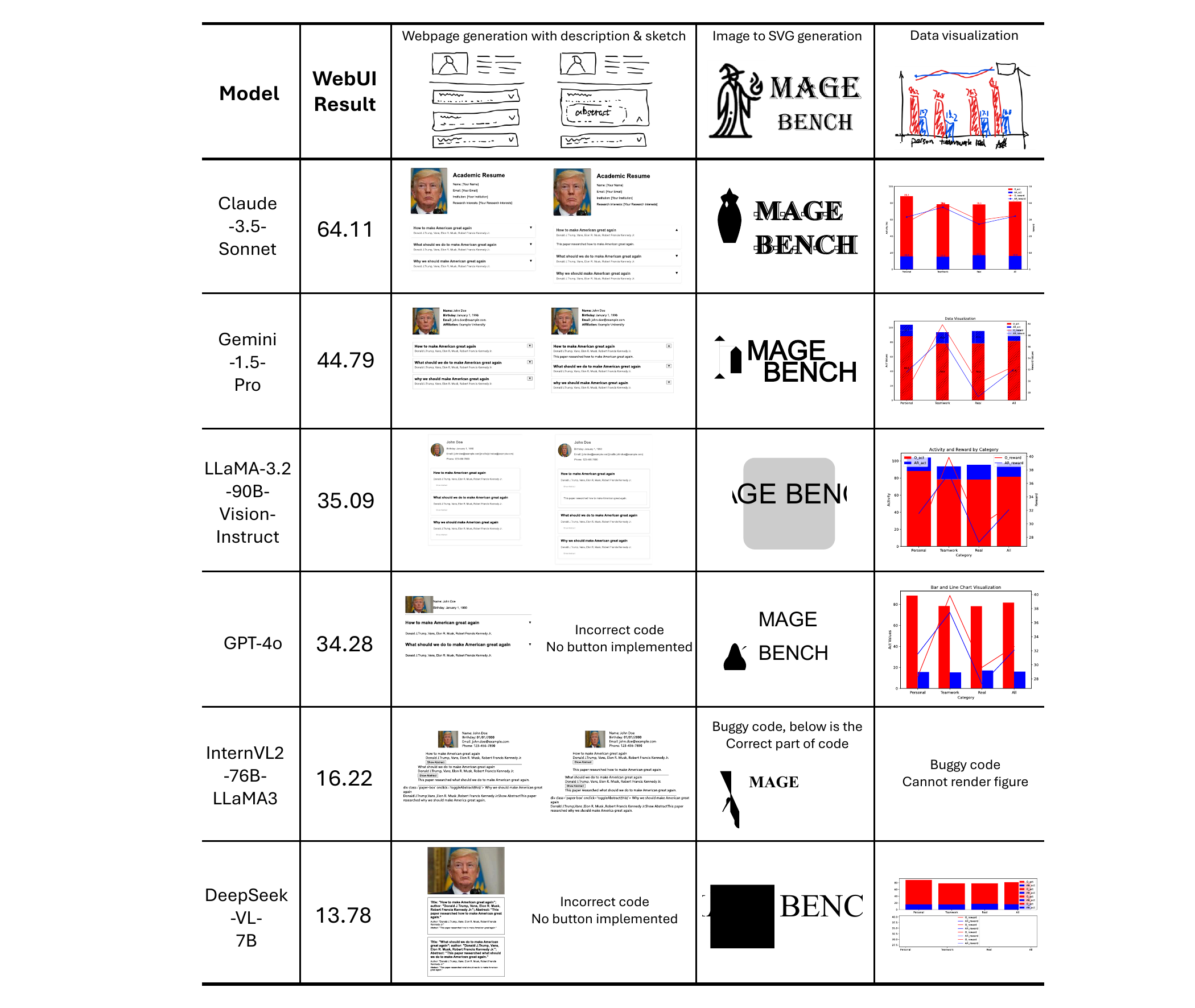}
    \caption{The cross-domain results demonstrate that our WebUI environment can robustly generalize to other engineering-level structured visual generation tasks.}
    \label{fig:web-general}
\end{figure*}

In Figure \ref{fig:web-general}, we evaluate six models: Claude-3.5-Sonnet, Gemini-1.5-pro, LLaMA-3.2-90B, GPT-4o, InternVL2-72B, and DeepSeek-VL. These models represent the top six based on WebUI scores. The figure includes their WebUI scores alongside the visualized results (Best of 5) from the three aforementioned generation tasks. Overall, the qualitative analysis results are highly correlated with the WebUI scores. We elaborate on this correlation in the following.

Firstly, regarding web page generation, the Claude model not only perfectly meets the requirements but also includes additional details such as dropdown animations, hover animations, and container shadow enhancements (although these may not be very apparent in the figure). Gemini correctly implements all functionalities, though the visual appearance is relatively less dynamic. The similarly scored LLaMA-3.2-90B and GPT-4o exhibit certain deficiencies; for instance, LLaMA-3.2-90B has discrepancies in the shape and position of the dropdown button compared to the task description, while GPT-4o has errors in implementing the dropdown functionality. The two lower-scoring models clearly lag behind the others.

In the SVG generation task, it is evident that Claude stands out from the other models, producing results that most closely resemble the specified fonts and logo shapes. The middle three models form a distinct group, while the last two models exhibit noticeable differences that set them apart from the others.

For the data visualization task, we have four core requirements (each worth one point): correct display of the bar chart, correct display of the legend, the line chart should be scaled to the top, and numerical labels in the same color as the bar chart. Claude did not meet the bar chart requirement but addressed all other aspects, earning a score of 3 out of 4. Gemini failed to scale the line chart correctly and had errors in font color and placement, resulting in a score of 1.5. Similarly, LLaMA-3.2-90B and GPT-4o scored between 0.5 and 1 point. The last two models continued to perform poorly in this task.


The aforementioned analysis indicates that our results can be robustly generalized to other structured visual generation tasks, including engineering applications. Notably, in the VisualAgentBench, they introduced an environment called VAB-CSS, where models debug CSS properties. In their results, the performance ranking was as follows: GPT-4o $>$ InternVL2-8B $>$  LLaVA-Next-8B $>$ Claude-3.5-Sonnet $>$ Gemini-1.5-pro $>$ Qwen-VL-9B. This differs from our results and qualitative analysis. The discrepancy arises because debugging CSS properties and end-to-end webpage generation are fundamentally different tasks and may not be strongly correlated. As previously mentioned, GPT-4o demonstrated the best performance in the Online setting, which might give it an edge in the CSS debugging task. However, for selecting a model for end-to-end structured visual generation tasks, our environment provides more relevant insights.

\section{Open questions and future researches}
\label{app:openquestion}
In the process of proposing MageBench and conducting our evaluations, we have identified some capabilities that existing models lack, feasible research directions, and open questions. We hope that these summaries can assist and inspire future work:

\begin{itemize}
    \item Existing open-source models exhibit a significant deficiency in ViC-type reasoning capabilities. How to obtain large amounts of ViC-type training data is an open question worthy of research. 
    \item The models exhibit a lack of imagination and long-term planning abilities in visual spatial reasoning tasks (such as Sokoban). The Best-of-N results indicate that the current models do not possess such potential, necessitating fundamental, mechanism-level research.
    \item For most tasks that exhibit potential, the models show considerable improvement in the Best-of-N results. How can we design a reinforcement learning approach to enhance the models, enabling them to accumulate explainable, text-expressible experiences, similar to how humans accumulate tips and tricks for playing games?
    \item Theoretically, an Online Agent should benefit from having a longer memory when completing tasks such as Sokoban and Football, or when making multiple rounds of modifications to a generated webpage. However, our observations show counterintuitive results where not only is there no benefit, but there is even a slight decline in performance. From an engineering perspective, this could be due to the model's lack of ability to handle long contexts, multi-turn dialogues, and tasks involving multiple images.
\end{itemize}

\newpage
\section{Examples}
\label{app:examples}

\subsection{WebUI}
\subsubsection{Labeled target webpage}
\label{sec:webpage}
\begin{figure*}[htbp]
	\centering
        \vspace{20pt}
	\begin{minipage}{0.33\linewidth}
		\centering
		\includegraphics[width=0.9\linewidth, trim=0 0 0 0, clip]{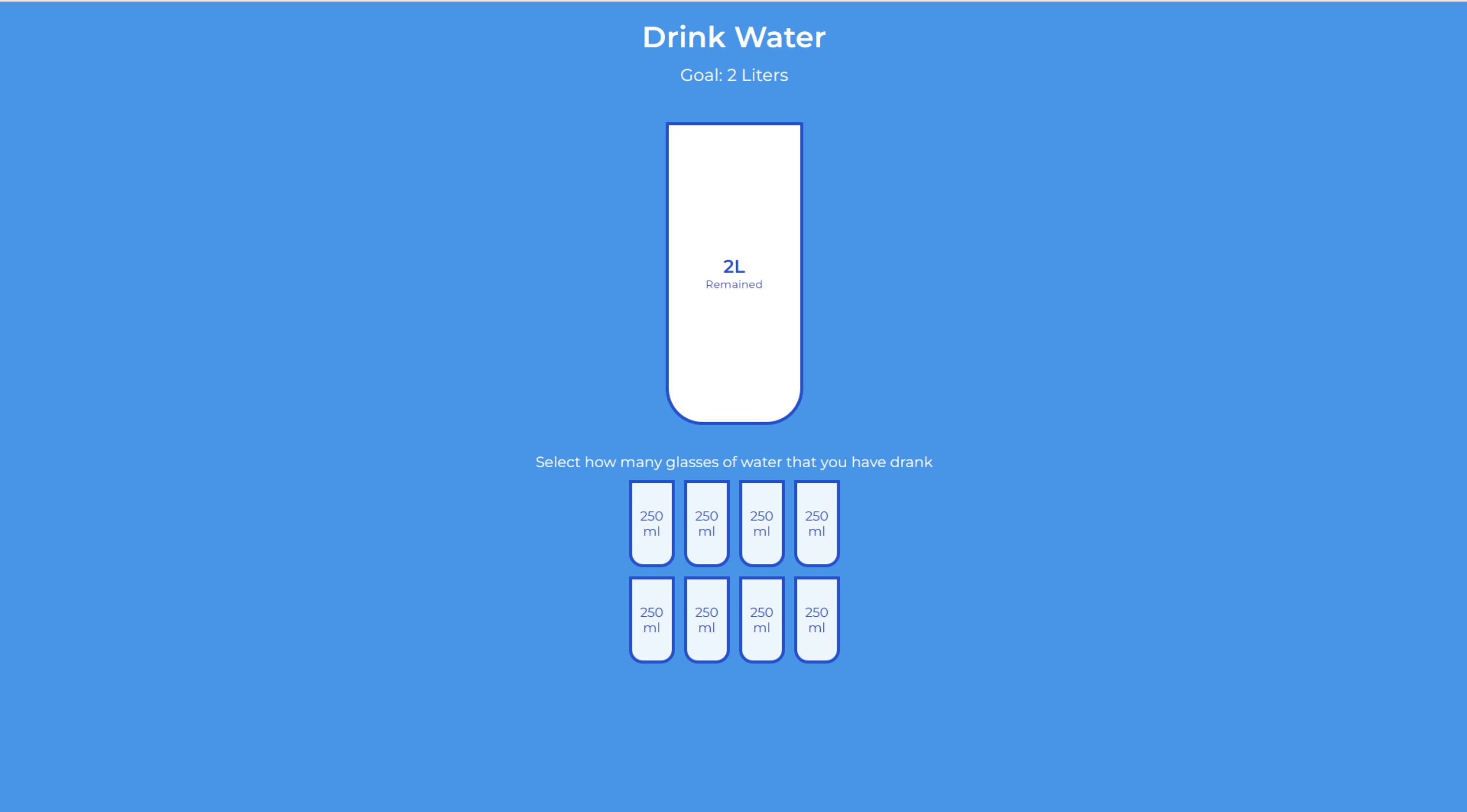}
	\end{minipage}
	\begin{minipage}{0.33\linewidth}
		\centering
		\includegraphics[width=0.9\linewidth, trim=0 0 0 0, clip]{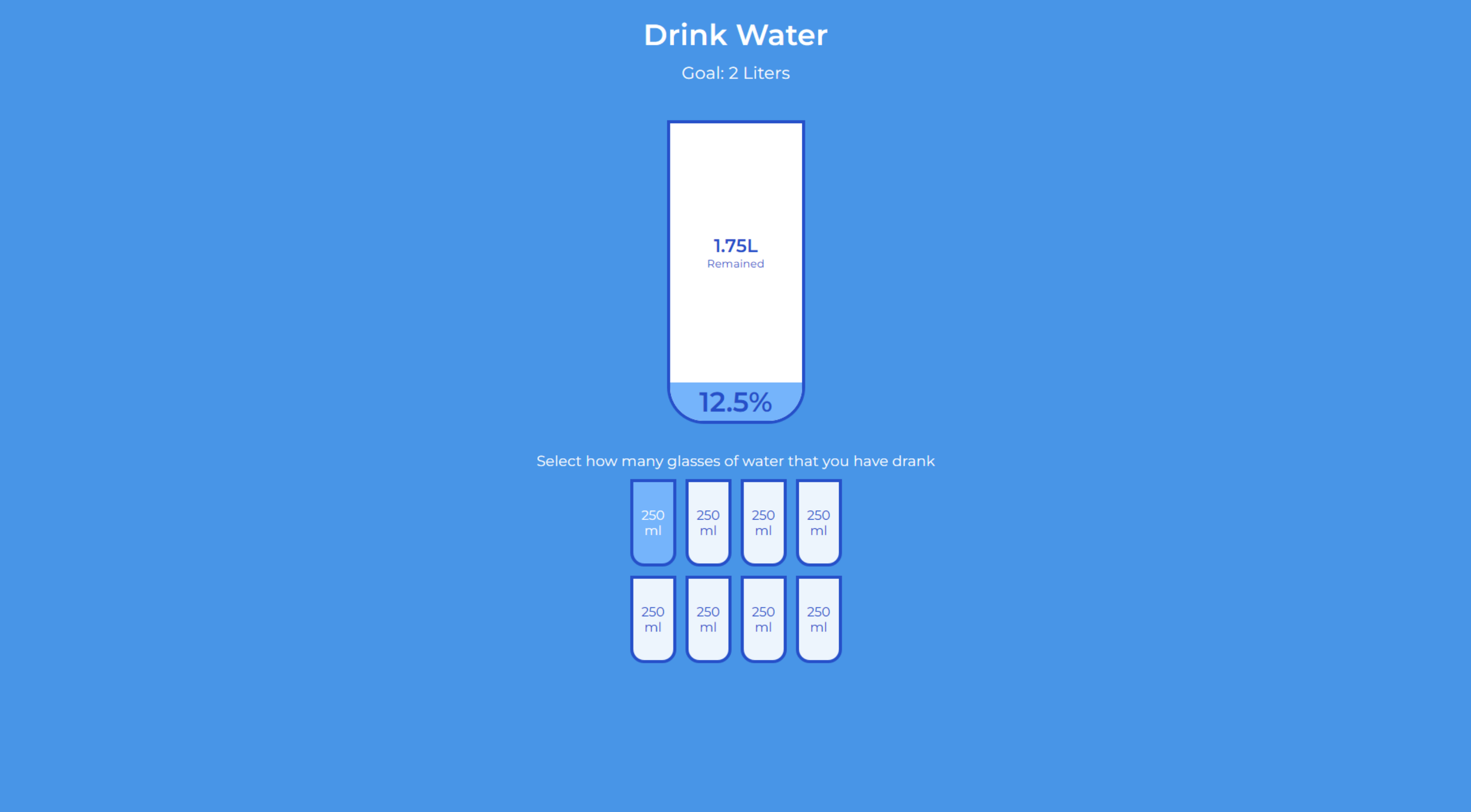}
	\end{minipage}
        \begin{minipage}{0.33\linewidth}
		\centering
		\includegraphics[width=0.9\linewidth, trim=0 0 0 0, clip]{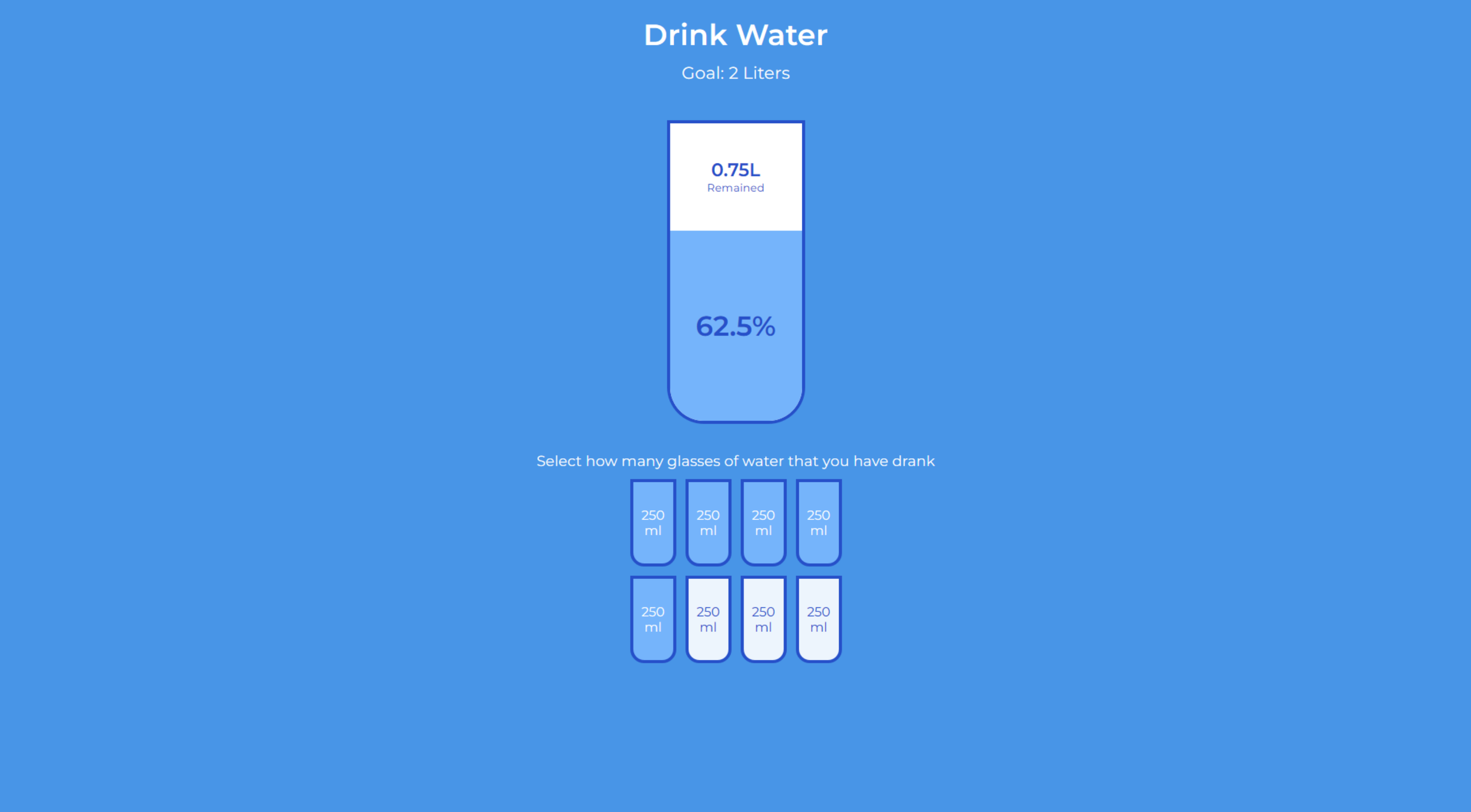}
	\end{minipage}
    \caption{A sample web page is provided where, in this example, clicking the small container button below will inject the corresponding water into the large container.}
    \label{fig:webexample}
\end{figure*}

\begin{figure*}
    \centering
    \includegraphics[width=\linewidth, trim=0 450 0 0, clip]{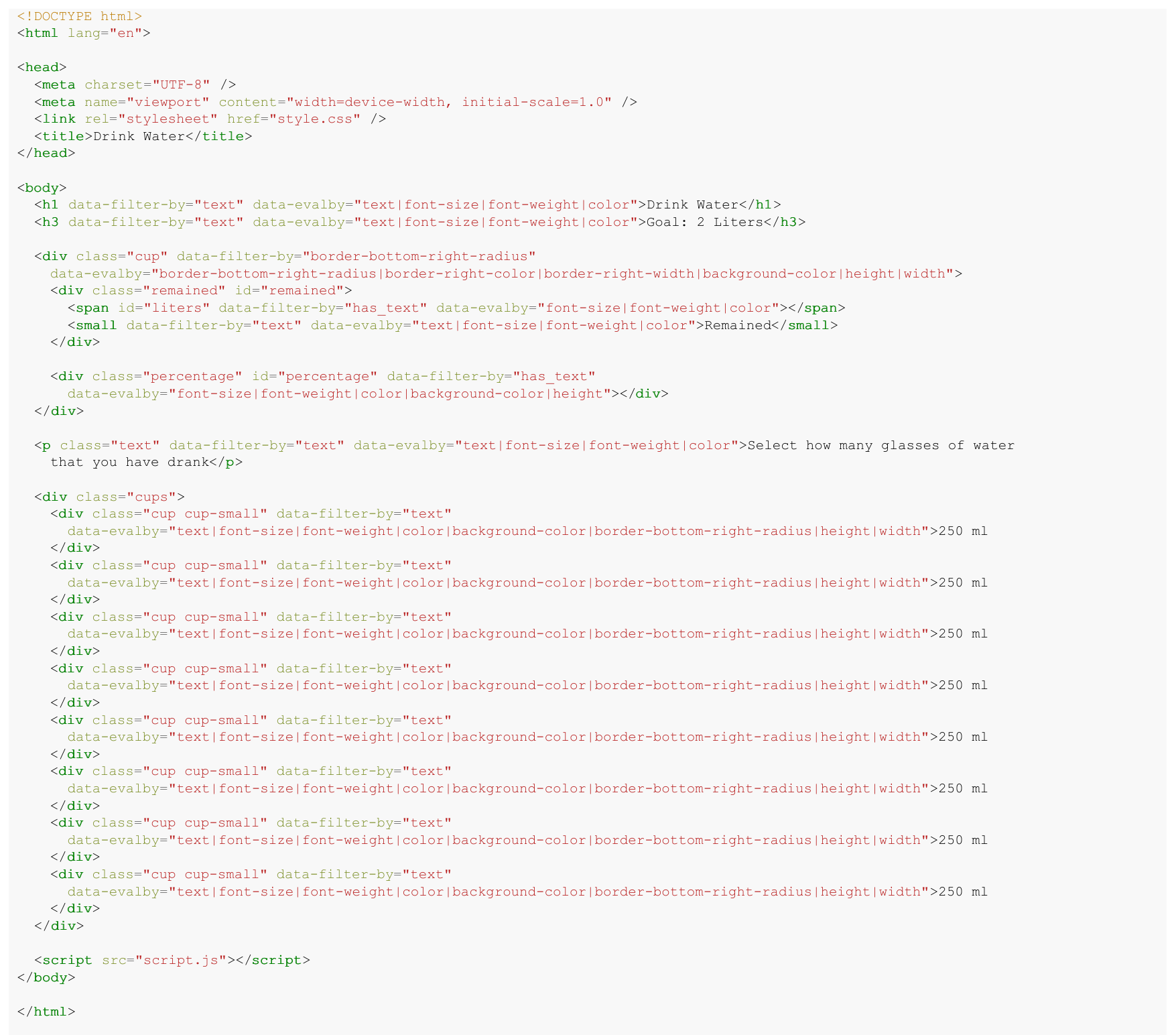}
    \caption{An example of preprocessed HTML file.}
    \label{fig:htmlexample}
\end{figure*}

In this subsection, we will present specific examples of web page processing. First, regarding the preprocessing of the target web page's HTML, as mentioned in Section \ref{sec:intro-webui}, we will annotate two new attributes to the atomic elements of the web page: `data-filter-by' and `data-eval-by'. Figure \ref{fig:webexample} shows a screenshot of a sample web page, and the processed code is shown in Figure \ref{fig:htmlexample}. 

\subsubsection{Interactions}
\label{sec:actcode}

\begin{figure}
    \centering
    \includegraphics[width=\linewidth, trim=0 520 0 0, clip]{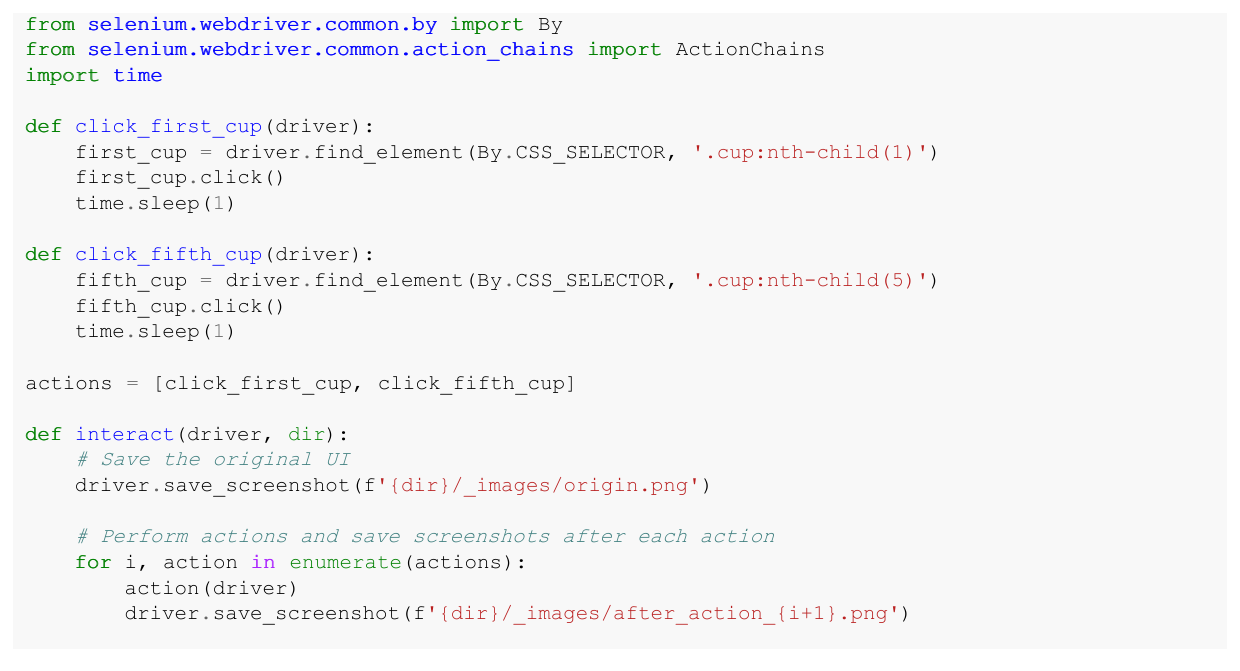}
    \caption{An example python file for interaction.}
    \label{fig:interactionexample}
\end{figure}

In order to simulate webpage interactions, we have implemented automated interaction scripts for each target webpage, as illustrated in Figure \ref{fig:interactionexample}. These scripts consist of predefined interaction code. In the example provided, the code demonstrates interactions with the first and fifth small containers on the webpage depicted in Figure \ref{fig:webexample}. 

It is noteworthy that the webpage interactions are element-based. For instance, identifying the fifth element with the class name ``cup" (see Figure \ref{fig:interactionexample}) necessitates that the model correctly assigns the specified class name, ID, or other attributes. In the task description, we explicitly instruct the model to set the appropriate attribute names to facilitate automated testing.

If the model incorrectly sets these class names, it will result in a failure of the automated interaction. Consequently, this action will be considered a failure when calculating the score, corresponding to an ``interaction error" as defined in the main text.
\begin{figure*}[htbp]
	\centering
	\begin{minipage}{0.33\linewidth}
		\centering
		\includegraphics[width=0.9\linewidth, trim=0 0 0 0, clip]{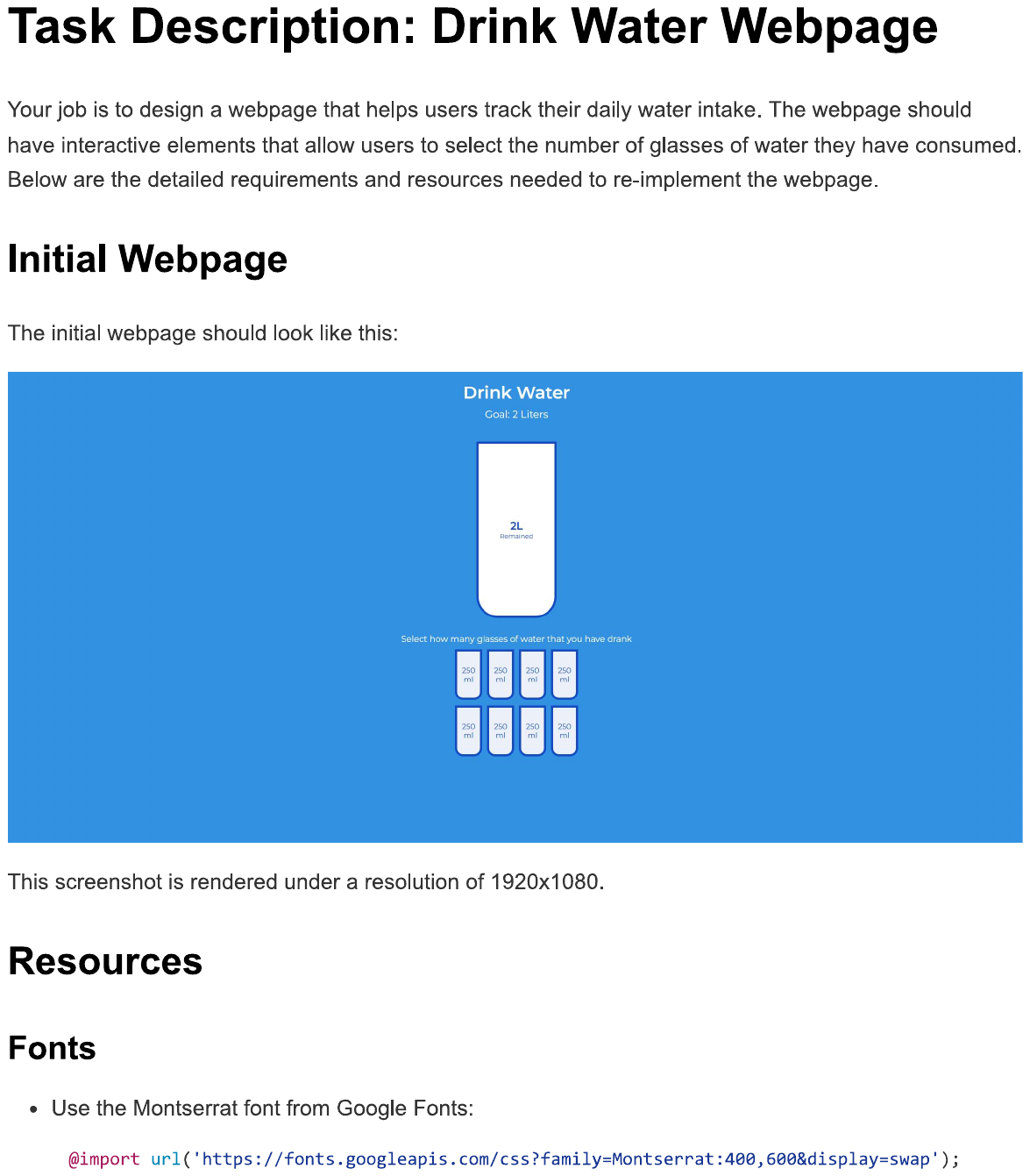}
	\end{minipage}
	\begin{minipage}{0.33\linewidth}
		\centering
		\includegraphics[width=0.9\linewidth, trim=0 0 0 0, clip]{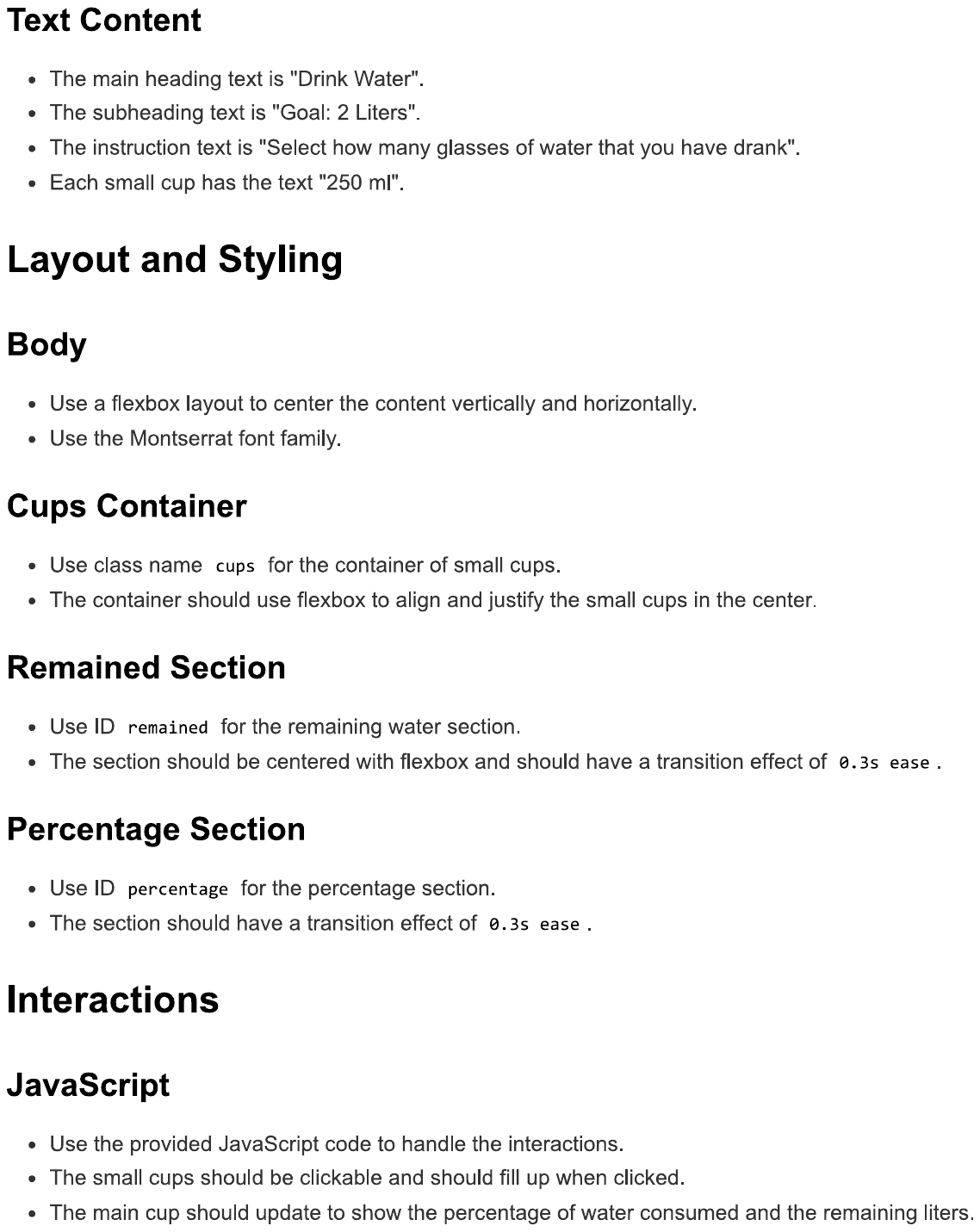}
	\end{minipage}
        \begin{minipage}{0.33\linewidth}
		\centering
		\includegraphics[width=0.9\linewidth, trim=0 0 0 0, clip]{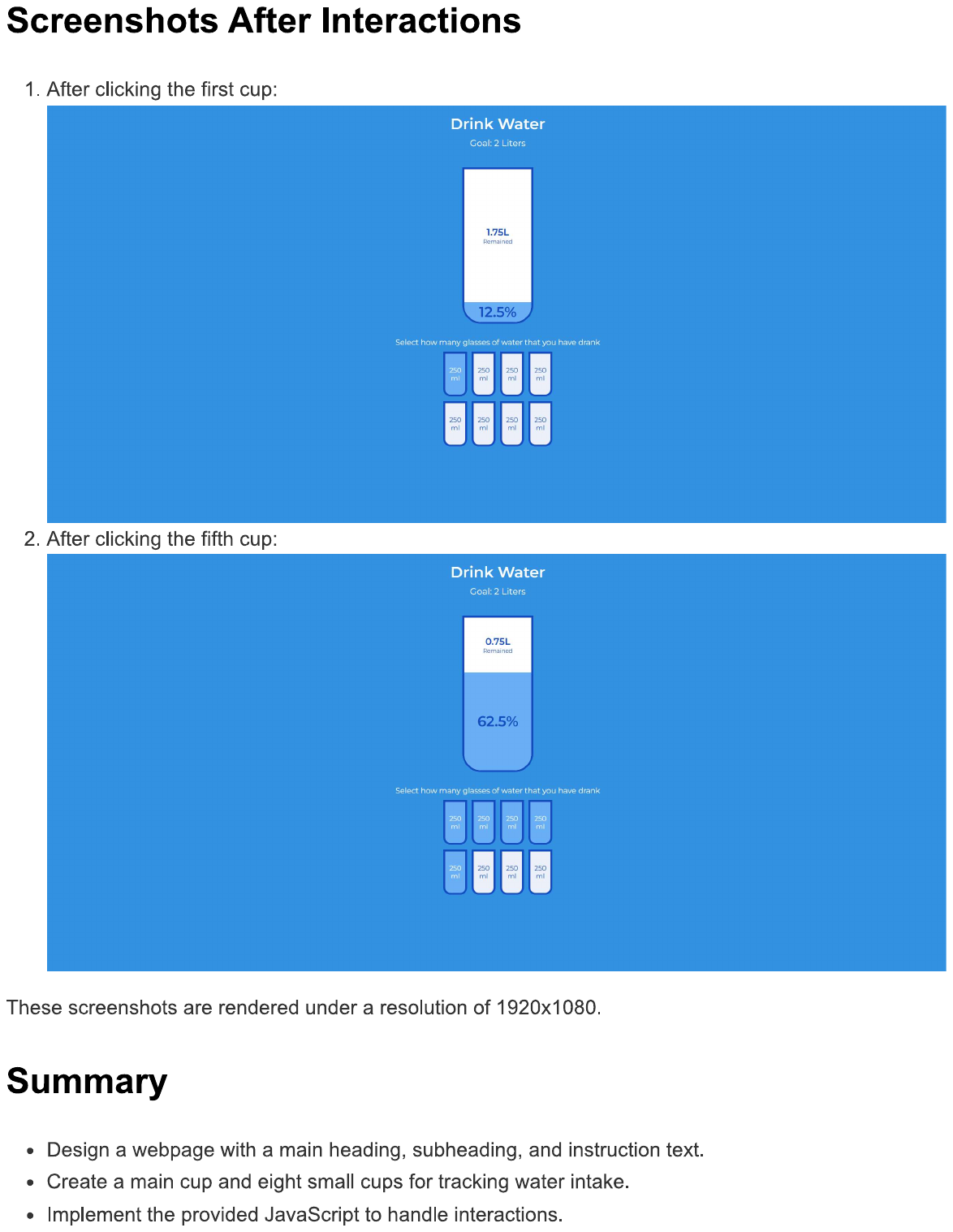}
	\end{minipage}
    \caption{An example of task description document.}
    \label{fig:descriptionexample}
\end{figure*}

\subsubsection{Task descriptions}
\label{sec:webtaskdes}

Figure \ref{fig:descriptionexample} illustrates the task description for reproducing the webpage shown in Figure \ref{fig:webexample}. We visualize this using markdown format. When provided as input to the large model, the task description is an interleaved image-text document. It details the functionality of the target webpage, includes webpage screenshots, various resources, and specifies class name requirements for automated testing.

\subsubsection{Task prompts $p_{cot},  p_{io}$}
\label{sec:webtaskprompt}

In this section, we will present several examples of input-output prompts, as well as the detailed and complete input format used during the model invocation process.

\newpage
\textbf{Global setting.} The prompt format for WebUI-Global setting is shown as follow:

\begin{verbatim}
{SYSTEM PROMPT}

{TASK DESCIPTION}

Generate all necessary codes at a time. 
Your output should contain code blocks 
in markdown format, e.g., 
```html\nxxxx```. generate javascript 
and css code if necessary. They will 
be writen to `index.html`, `script.js`
and `style.css`.

\end{verbatim}

The content of \{SYSTEM PROMPT\} should be replaced with that from Sec. \ref{sec:webuisys}. Conversely, the \{TASK DESCRIPTION\} is similar to the document depicted in Figure \ref{fig:descriptionexample}. Upon generating the model output, we directly identify code snippets, using regular expressions. These code snippets are then stored in the corresponding project folder.

\textbf{Online setting.} Under this setting, we start with the webpage generated by the model under the Global setting. This approach allows us to fairly compare the improvements made by different online prompts to the same initial code, ensuring a fairer comparison. Specifically, when running the Online setting, we already have an initial codebase. The model then modifies this code based on the initial code and the images rendered by the browser. There are 3 online prompt methods mentioned in the main part: ``0s-CoT'', ``HD-CoT'' and ``HD-CoT-NoIMG''. We will show their prompt implementation separately.

\textbf{0s-CoT:}
\begin{verbatim}
{SYSTEM PROMPT}

{TASK DESCIPTION}

Current implementation:
```html
{PREV HTML CODE}
```

```css
{PREV CSS CODE}
```

```javascipt
{PREV JAVASCRIPT CODE}
```

{RANDER FEEDBACK IMAGE}


Please adapt the code according to the 
current rendered screen shot of previous 
implementation, with this output format: 

# Analyze
Your Analyze
# Regenerate
re-generated code

If some of the files (e.g., javascript) 
do not need to be changed, just write
'*javascript do not need to change*' and
do not generate any javascript code, the
same for html and css.
\end{verbatim}
The \{RENDER FEEDBACK IMAGE\} is a PNG image that represents the result of the current code being rendered in the browser. As you can see, the prompt ask the model to analyze, but do not point out how to, thus we call this prompt as ``0-shot''.

\textbf{HD-CoT:}
\begin{verbatim}
{SYSTEM PROMPT}

{TASK DESCIPTION}

Current implementation:
```html
{PREV HTML CODE}
```

```css
{PREV CSS CODE}
```

```javascipt
{PREV JAVASCRIPT CODE}
```

{RANDER FEEDBACK IMAGE}


The above is an image of the previous
implementation webpage screen shot. 
Adapt the previous code with the 
following step by step flow and output
format:

# Analyze
answer: 1. What is the difference 
between the required webpage screen 
shot and the previous implementation 
webpage screen shot.
2. which parts of the previous code 
need to change and why.

# Regenerate
re-generate the files (html, css or 
javascript) that need to be changed.

If some of the files (e.g., javascript)
do not need to be changed, just write 
'*javascript do not need to change*'
and do not generate any javascript 
code, the same for html and css.
\end{verbatim}

\textbf{HD-CoT-NoIMG:}
\begin{verbatim}
{SYSTEM PROMPT}

{TASK DESCIPTION}

Current implementation:
```html
{PREV HTML CODE}
```

```css
{PREV CSS CODE}
```

```javascipt
{PREV JAVASCRIPT CODE}
```
The above is the previous implementation
webpage. 
Self reflect on the previous 
implementation and adapt the previous 
code with the following step by step 
flow and output format:

# Analyze
Analyze and self-reflect: 1. Are all the 
requirements in the description correctly 
fultilled.
2. If not, which parts of the previous 
code need to change and why.

# Regenerate
re-generate the files (html, css or
javascript) that need to be changed.
If some of the files (e.g., javascript)
do not need to be changed, just write 
'*javascript do not need to change*'
and do not generate any javascript 
code, the same for html and css.

\end{verbatim}

\begin{figure*}
	\centering
	\begin{minipage}{0.33\linewidth}
		\centering
		\includegraphics[width=\linewidth, trim=70 0 70 0, clip]{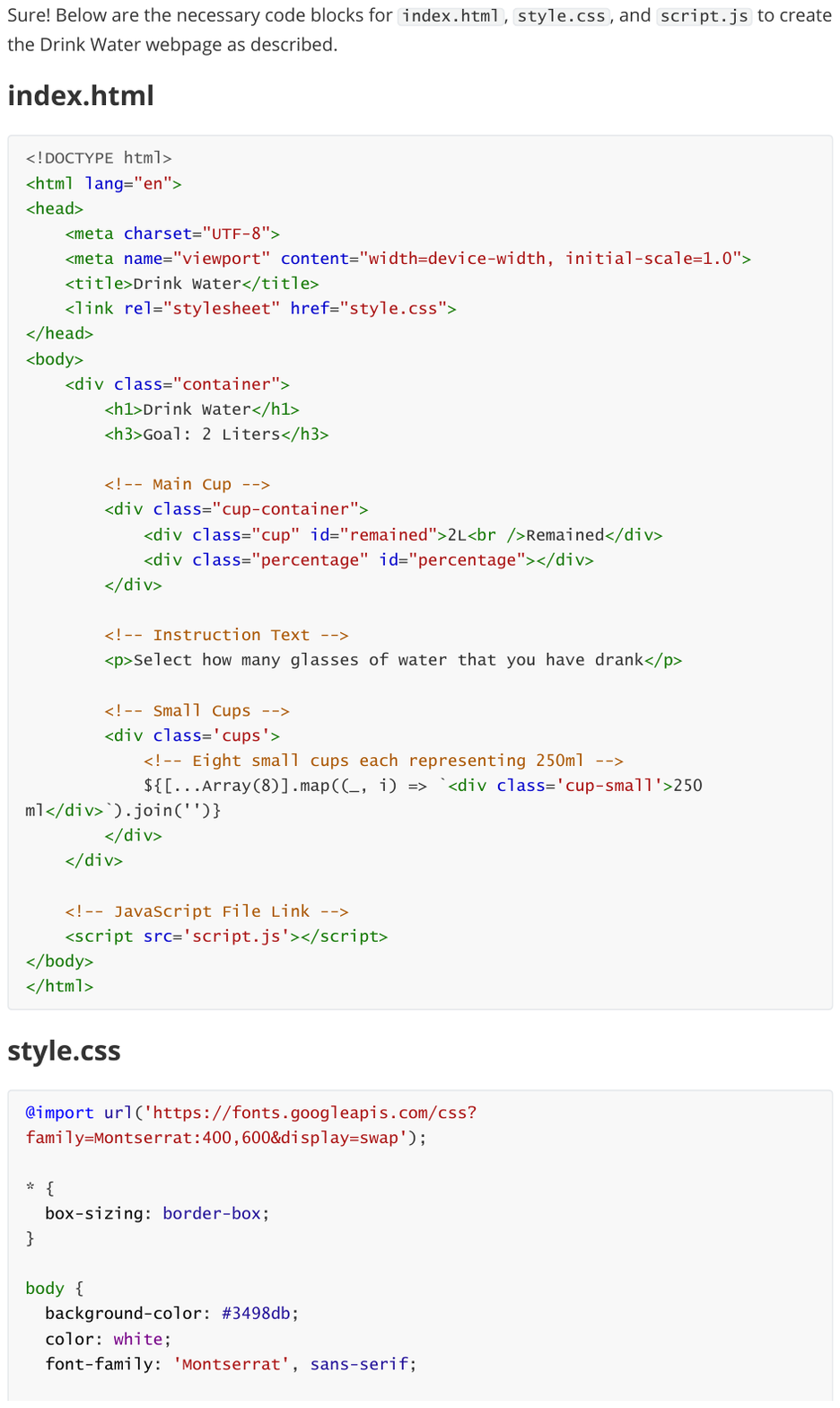}
	\end{minipage}
	\begin{minipage}{0.33\linewidth}
		\centering
		\includegraphics[width=\linewidth, trim=70 0 70 0, clip]{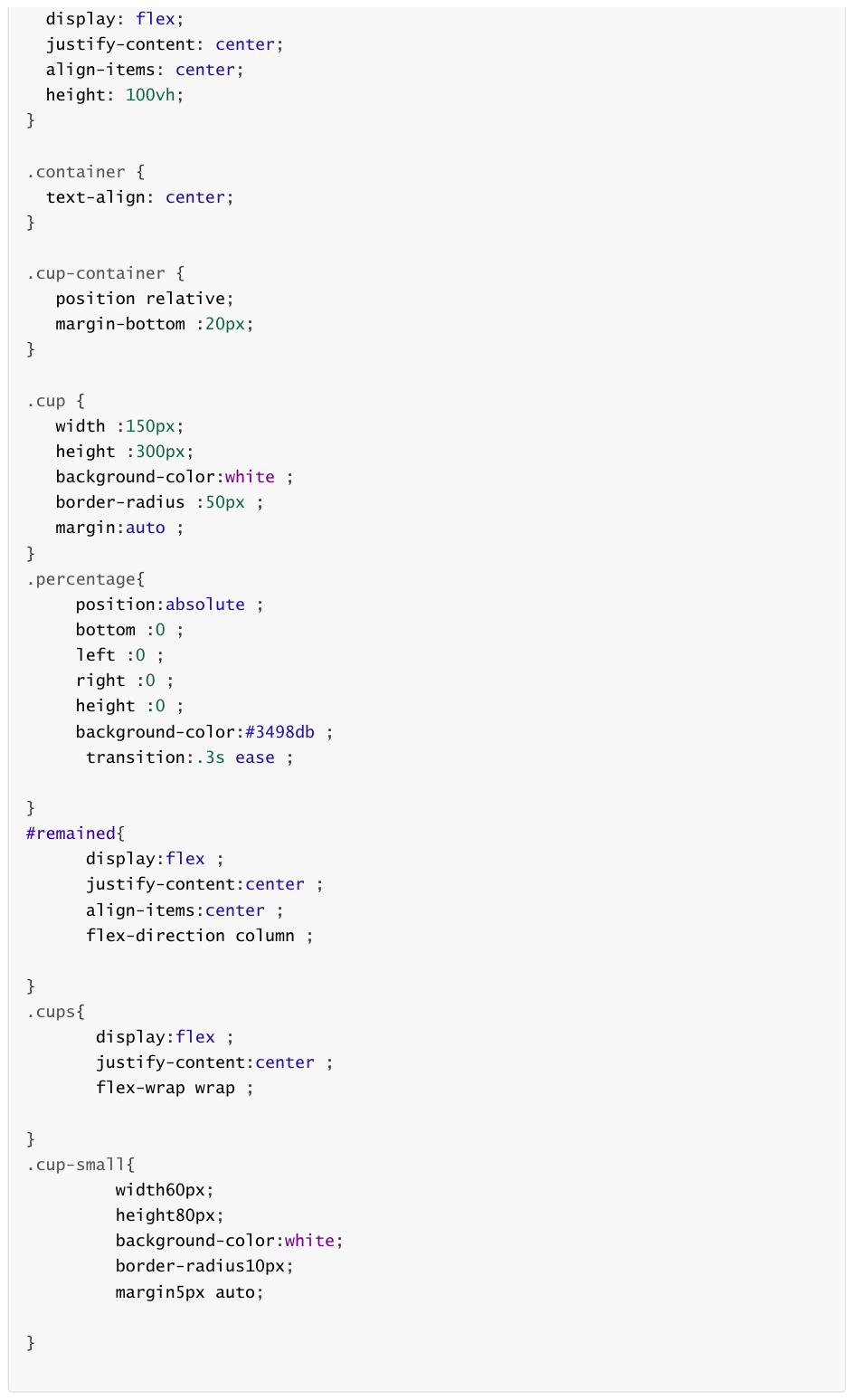}
	\end{minipage}
        \begin{minipage}{0.33\linewidth}
		\centering
		\includegraphics[width=\linewidth, trim=70 0 70 0, clip]{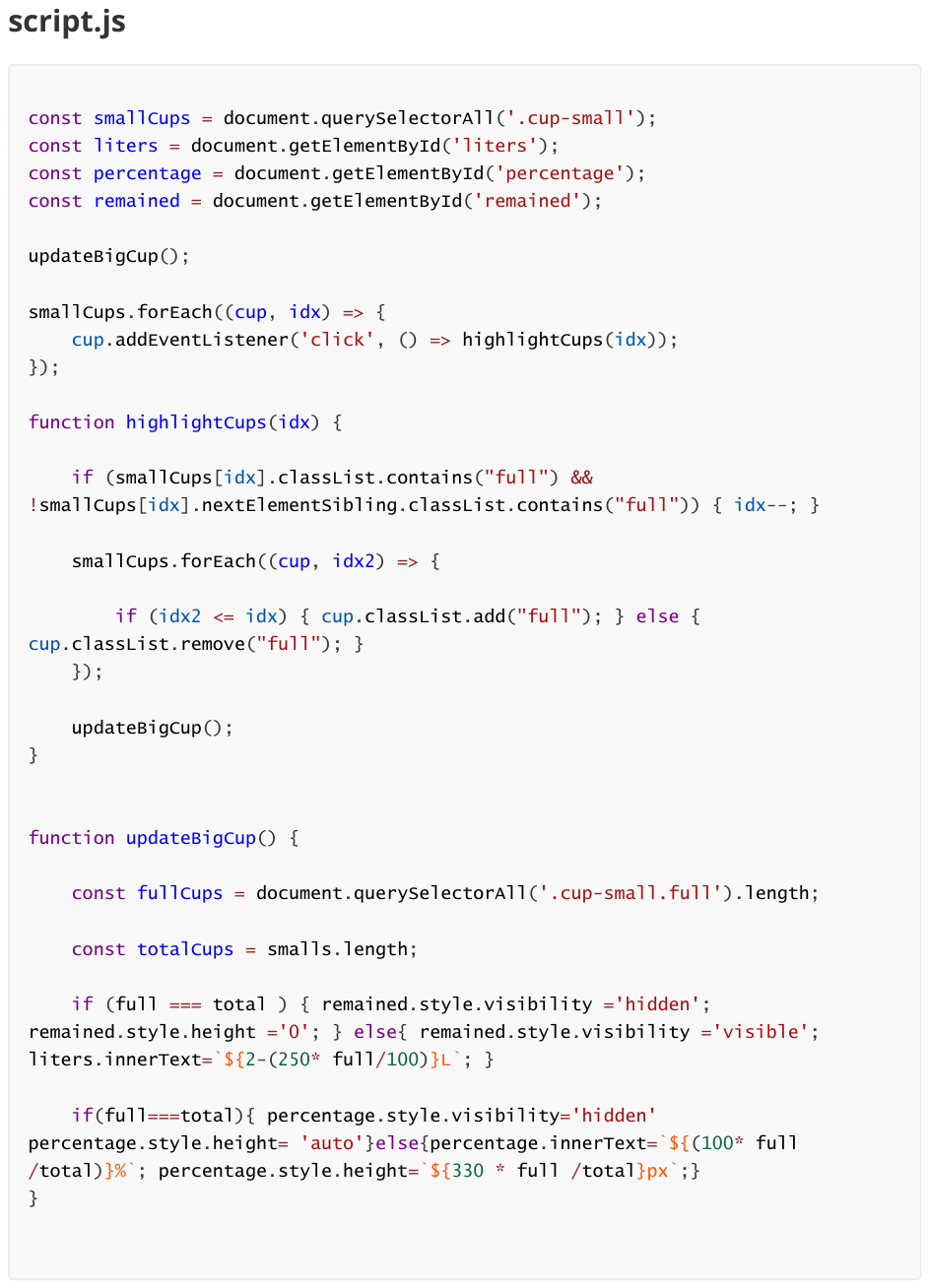}
	\end{minipage}

    \begin{minipage}{\linewidth}
		\centering
		\includegraphics[width=\linewidth, trim=0 0 0 0, clip]{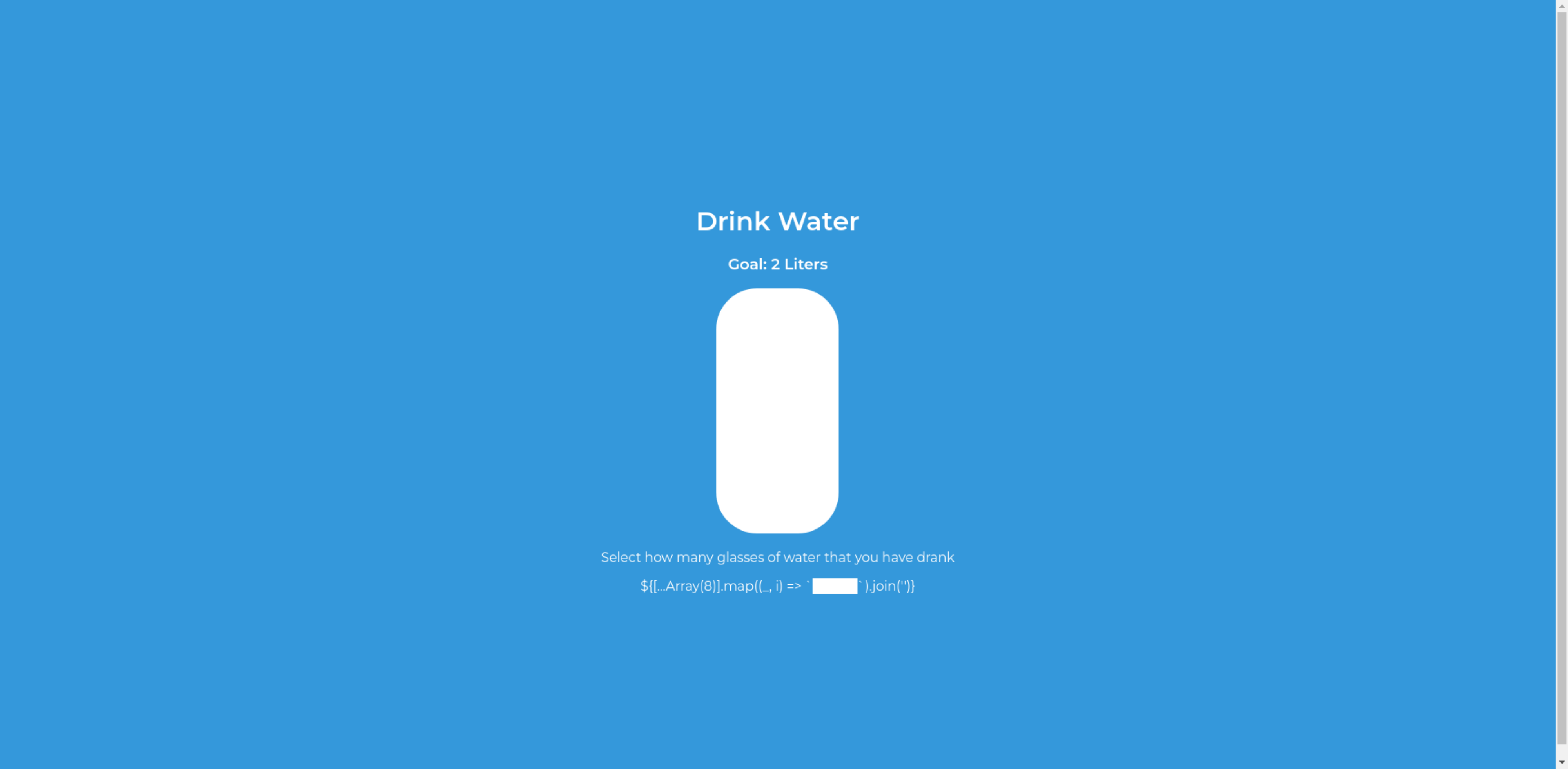}
	\end{minipage}
    \caption{GPT-4o's output text(Upper image) and screen shot of GPT-4o's implementation(Lower image). This implementation has a score of $AES=29.23$}
    \label{fig:web-out-g4o}
\end{figure*}

\subsubsection{Output examples}
\label{sec:webout}
\begin{figure*}[htbp]
	\centering
	\begin{minipage}{0.33\linewidth}
		\centering
		\includegraphics[width=\linewidth, trim=70 0 70 0, clip]{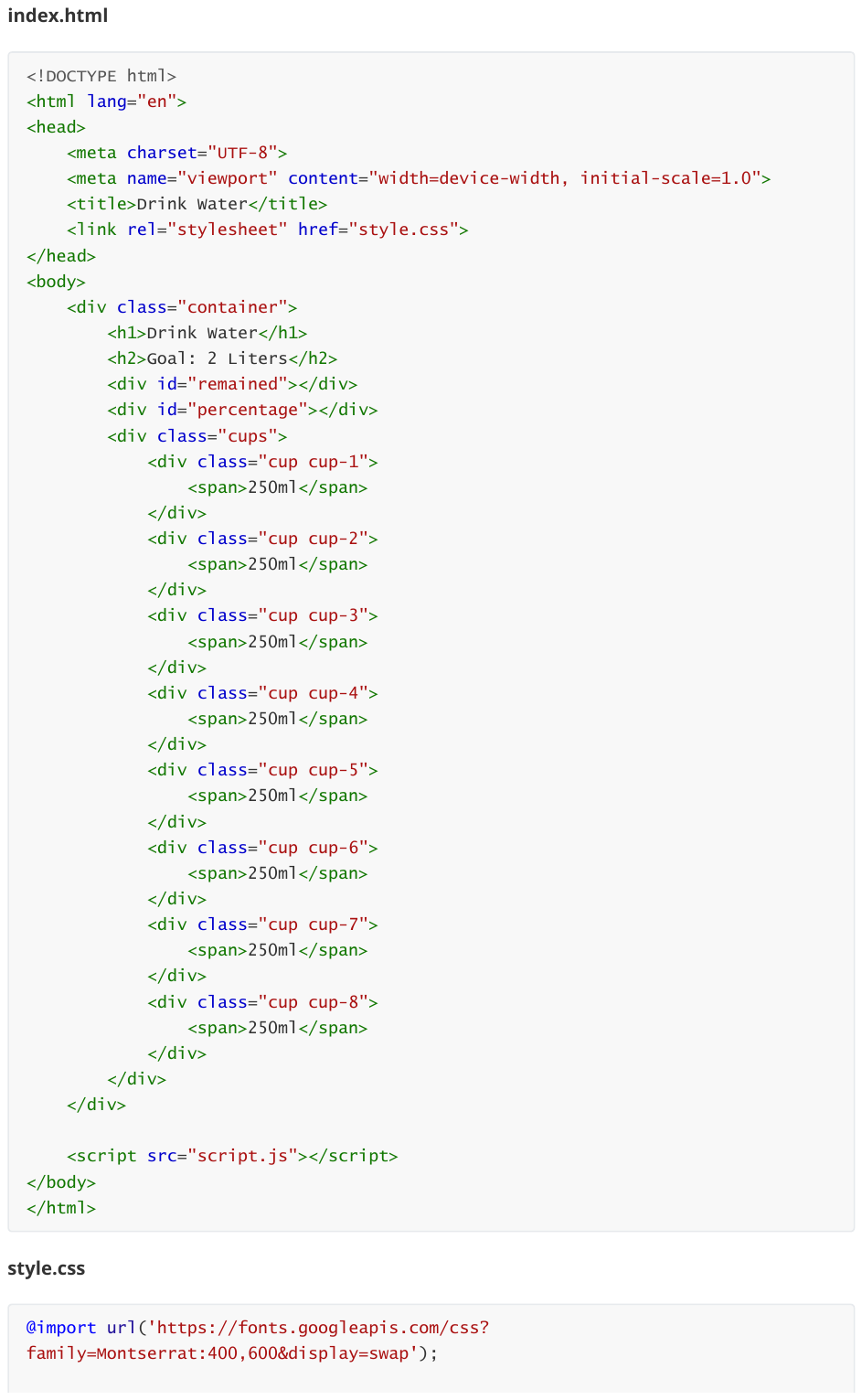}
	\end{minipage}
	\begin{minipage}{0.33\linewidth}
		\centering
		\includegraphics[width=\linewidth, trim=70 0 70 0, clip]{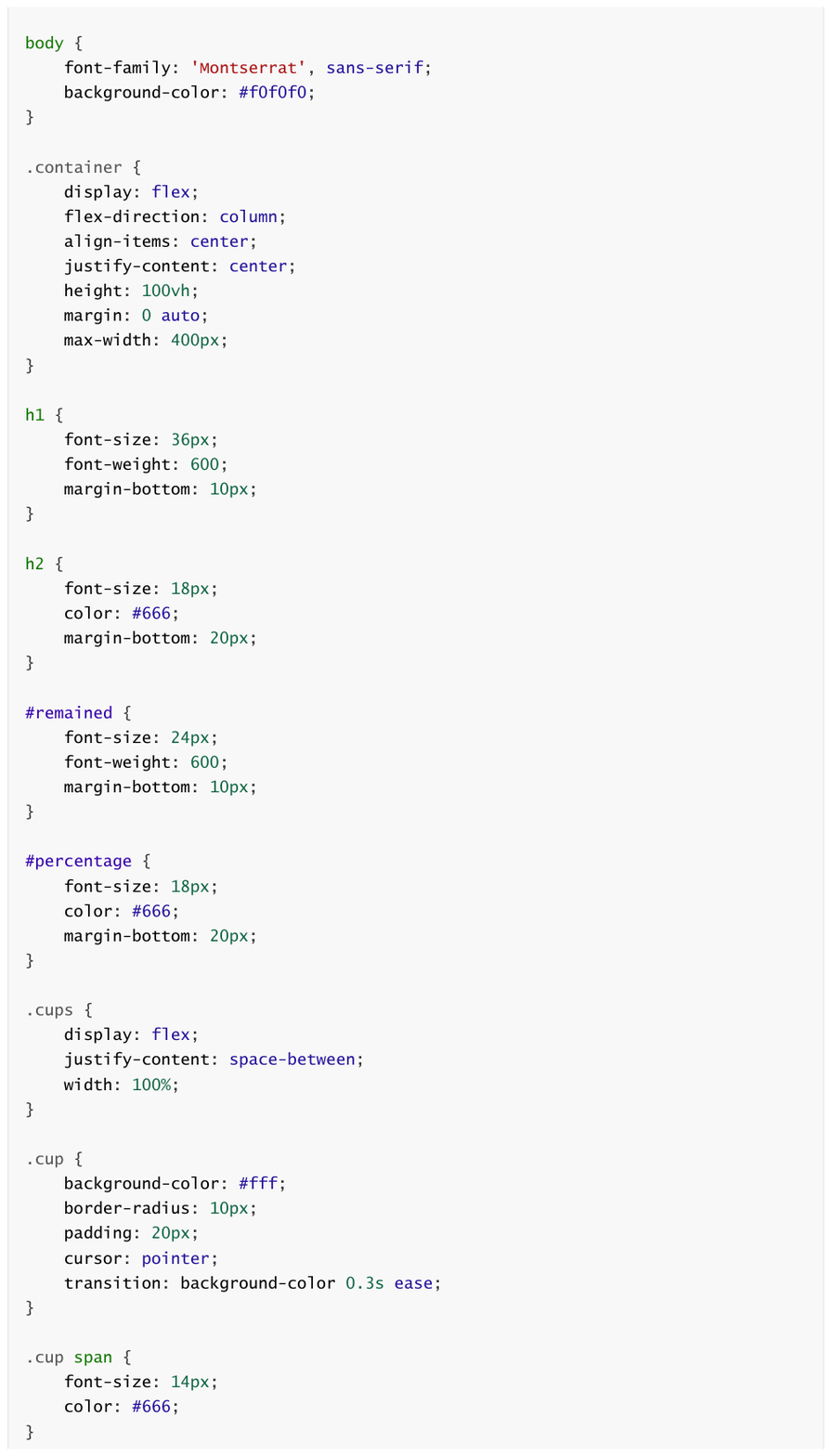}
	\end{minipage}
        \begin{minipage}{0.33\linewidth}
		\centering
		\includegraphics[width=\linewidth, trim=70 0 70 0, clip]{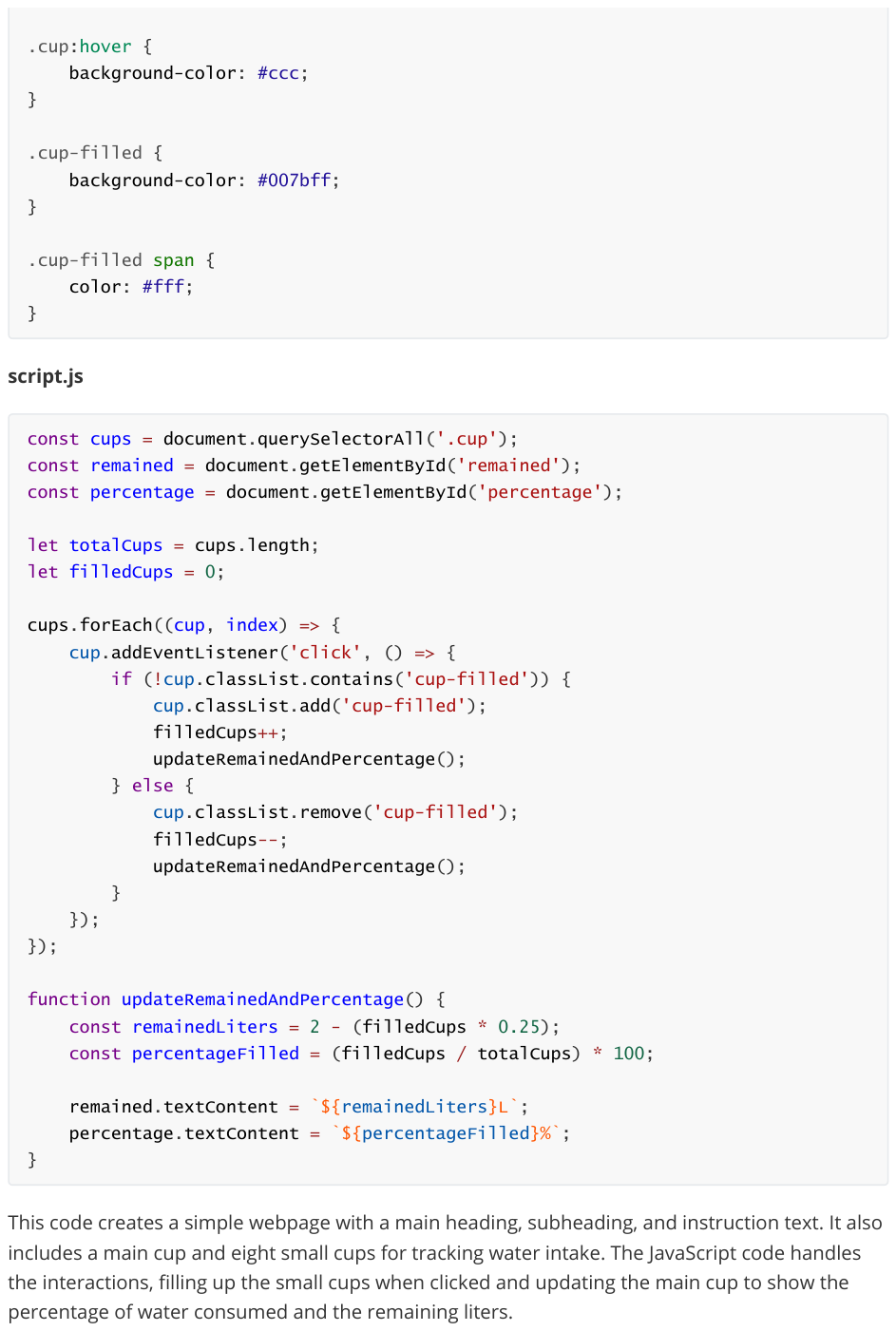}
	\end{minipage}

    \begin{minipage}{\linewidth}
		\centering
		\includegraphics[width=\linewidth, trim=0 0 0 0, clip]{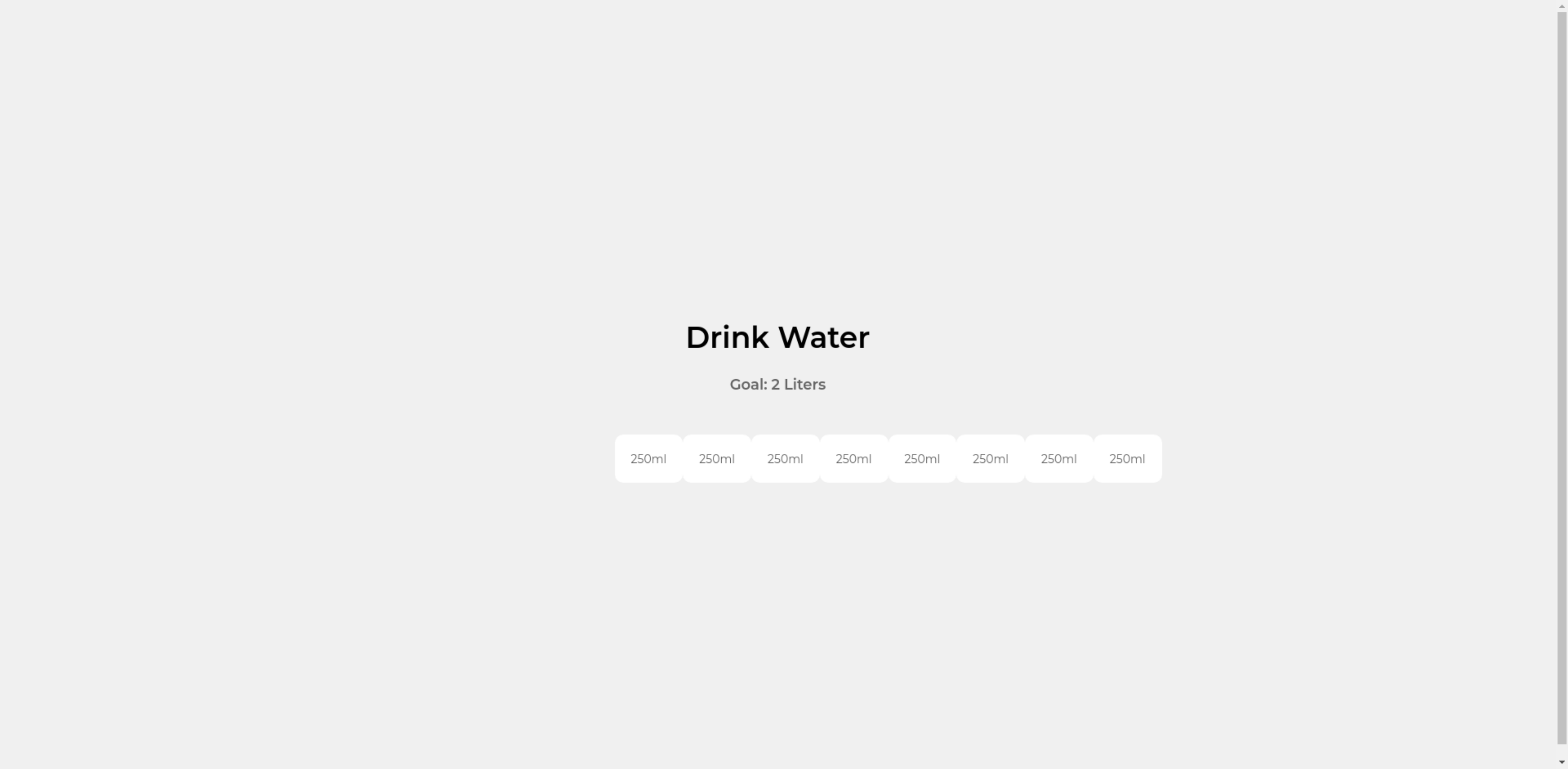}
	\end{minipage}
    \caption{LLaMA-3.2-Vision-Instruct's output text(Upper image) and screen shot of LLaMA-3.2-Vision-Instruct's implementation(Lower image). This implementation has a score of $AES=9.59$}
    \label{fig:web-out-llama}
\end{figure*}

Figures \ref{fig:web-out-g4o} and \ref{fig:web-out-llama} illustrate the outputs of the GPT-4o and LLaMA-3.2-90B models, respectively, along with the corresponding website renderings. Please refer to the previous task description in conjunction with this.

\newpage
\hfill
\subsection{Sokoban}

\subsubsection{Task prompts $p_{cot},  p_{io}$}
\label{sec:sotaskprompt}

\textbf{Global setting.} The prompt format of Sokoban-Global setting is shown as follow:
\begin{verbatim}
{SYSTEM PROMPT}

{INITIAL FRAME}

Please analyze and directly output a 
list of actions to take. 

Generate all actions to fulfill the 
task at a time. Your output should 
follow this format:

### Analyze
your analyze.
### Actions
A long sequence of Left, Right, Up, 
Down, separate by ','
\end{verbatim}
The same as WebUI, the content of \{SYSTEM PROMPT\} should be replaced with that from Sec. \ref{sec:sokobansys}. The \{INITIAL FRAME\} will be replaced by the image of the initial game scene. After the model generate the output, we just parse out the actions sequence and input to the environment.

\textbf{Online setting.} Differ from WebUI, we use a memory larger than 1 in Sokoban and football, and the memory in our codebase is implemented using multi-turns chat. In the following prompt, `-----User:' and `-----Assistant:' are not part of the prompt but an indicator of roles in the multi-turns chat. Now let's suppose we are at the time step $T$, the model needs to generate its response for action at time step $T+1$, and we use the standard settings: $AM=5, OM=1$.
\begin{verbatim}
-----User:
{SYSTEM PROMPT}
image not available.

You should perform 1 action for each 
generation, with this format: 
# analyze
some analyze
# action
action ID

For example: 
# analyze
...
# action
Up

-----Assistant:
{output at T-4}

-----User:
Please continue to act according to the 
current game scene.
image not available.

-----Assistant:
{output at T-3}

-----User:
Please continue to act according to the 
current game scene.
image not available.

-----Assistant:
{output at T-2}

-----User:
Please continue to act according to the 
current game scene.
image not available.

-----Assistant:
{output at T-1}

-----User:
Please continue to act according to the 
current game scene.
image not available.

-----Assistant:
{output at T}

-----User:
Please continue to act according to the 
current game scene.
{FRAME T}
\end{verbatim}

Here, as $OM=1$, the only game scene inputted is the \{FRAME T\} part. 

\subsubsection{Output examples}
\label{sec:soout}

\begin{figure*}
    \centering
    \vspace{-20pt}
    \includegraphics[width=\linewidth, trim=65 160 70 90, clip]{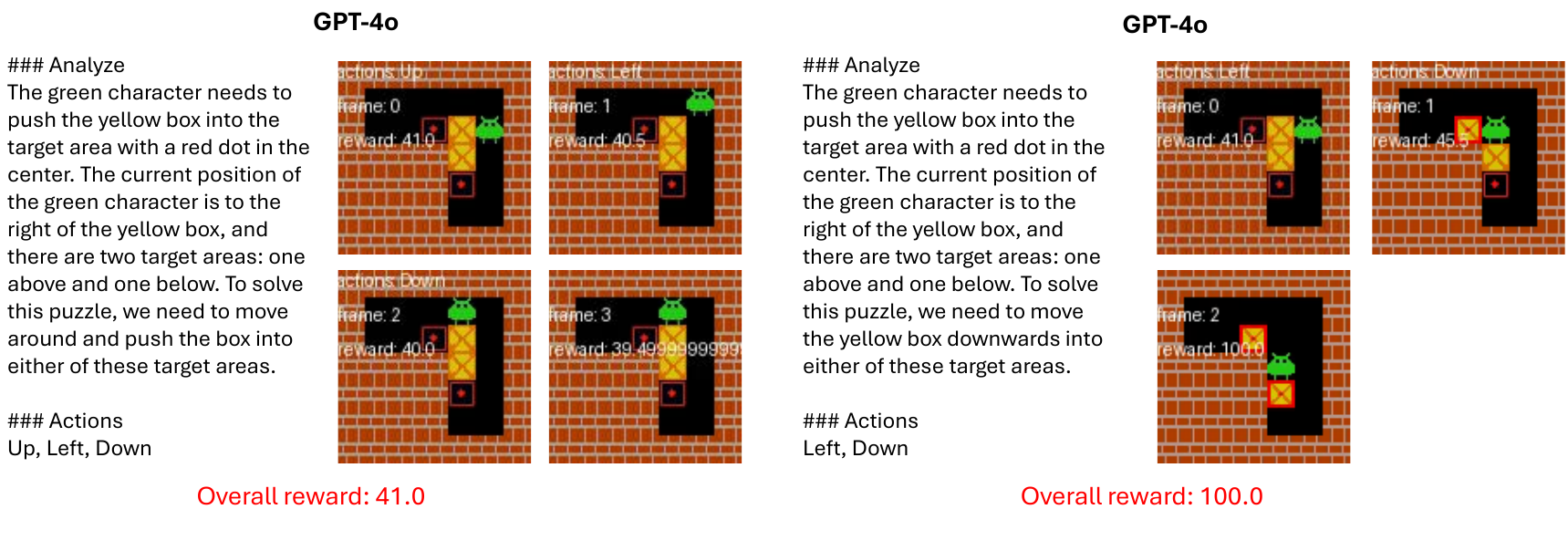}
    \vspace{-30pt}
    \caption{An example of GPT-4o's performances on Sokoban-Global settings, failed case (Left) and success case (Right).}
    \vspace{-10pt}
    \label{fig:sokobang4oexample}
\end{figure*}

\begin{figure*}
    \centering
    \includegraphics[width=\linewidth, trim=35 150 30 100, clip]{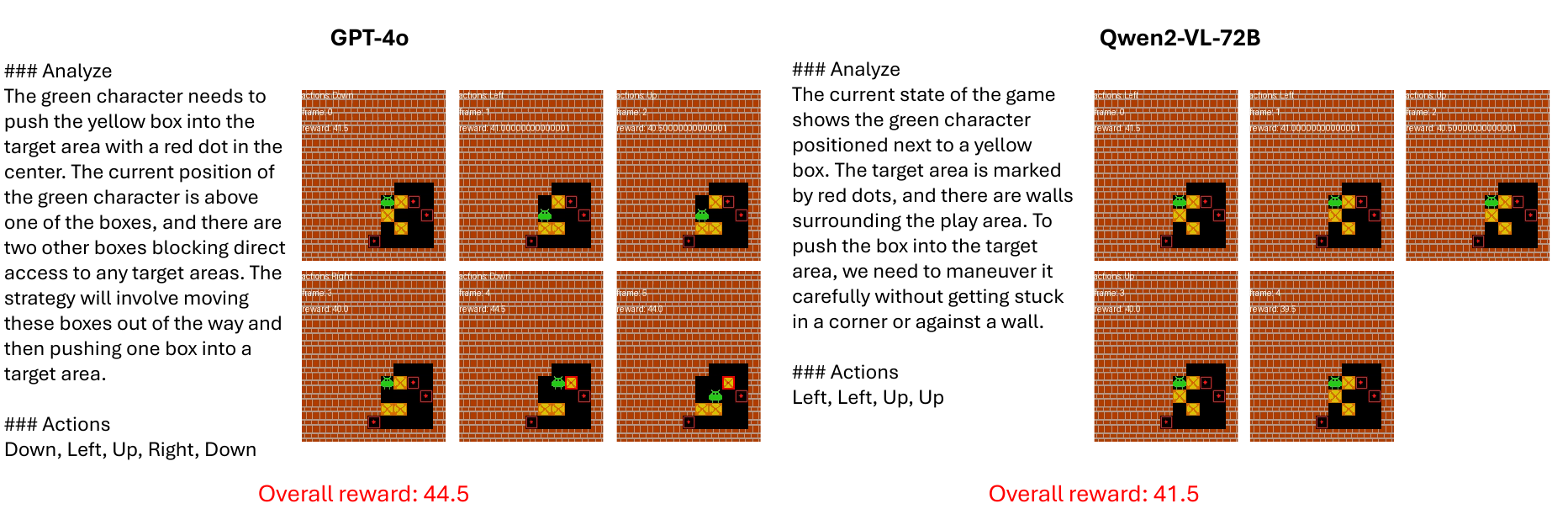}
    \vspace{-20pt}
    \caption{An example of GPT-4o's and Qwen2-VL-72B's performances on Sokoban-Global settings.}
    \vspace{-10pt}
    \label{fig:sokobang4oandqwenexample}
\end{figure*}

\begin{figure*}
    \centering
    \includegraphics[width=\linewidth, trim=50 70 50 70, clip]{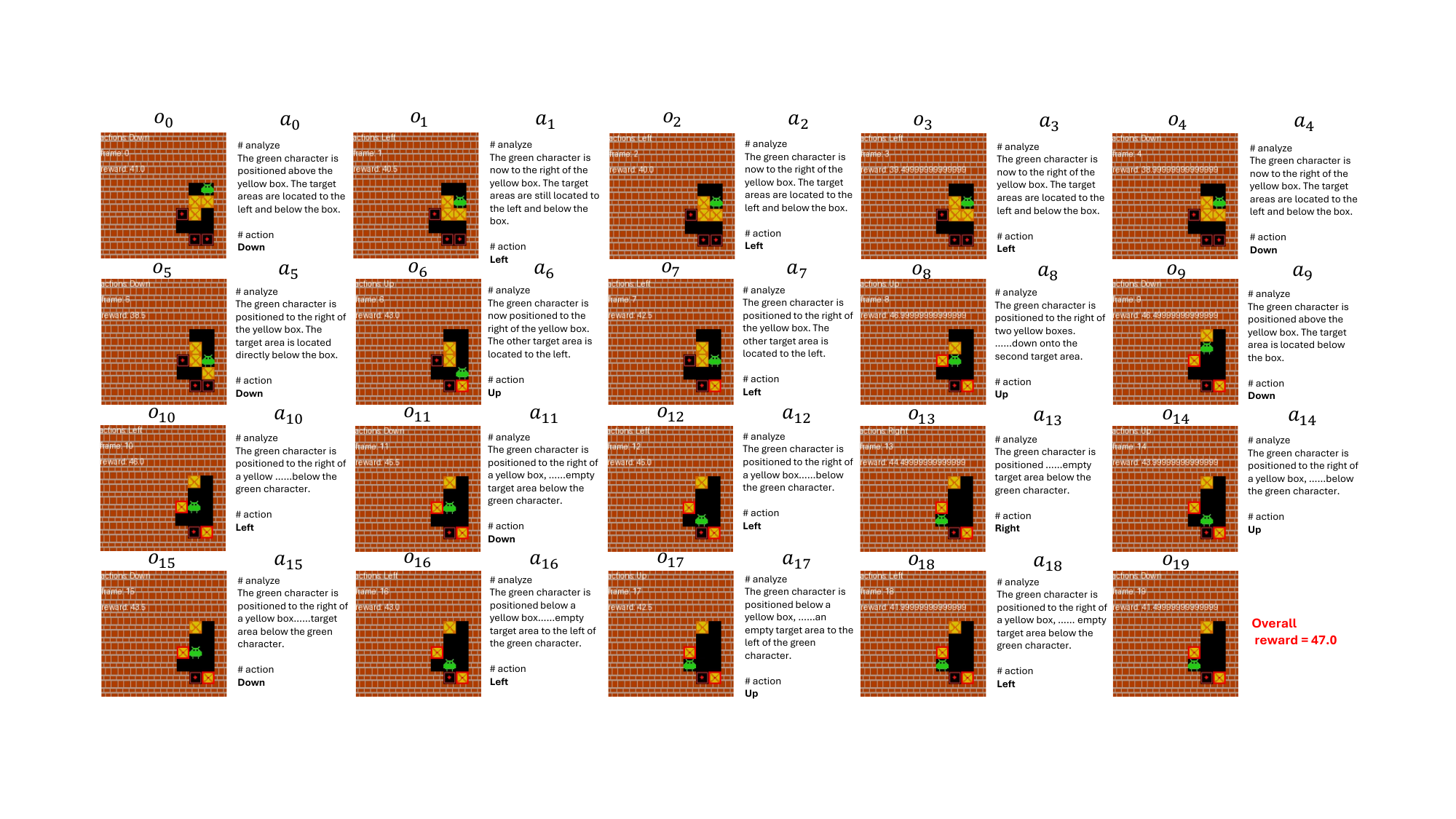}
    \vspace{-20pt}
    \caption{An example of GPT-4o's  performances on Sokoban-Online settings with max-loop=20.}
    \label{fig:sokobang4oonline}
\end{figure*}

Figures \ref{fig:sokobang4oexample} and \ref{fig:sokobang4oandqwenexample} present several examples within the Sokoban-Global environment. Figure \ref{fig:sokobang4oexample} illustrates both successful and unsuccessful cases of the GPT-4o model within the same scenario. Figure \ref{fig:sokobang4oandqwenexample}, on the other hand, compares the performance of different models within the identical scenario. Figure \ref{fig:sokobang4oonline} depicts the entire process under the Online setting.


\begin{figure*}
    \centering
    \includegraphics[width=\linewidth, trim=0 0 0 0, clip]{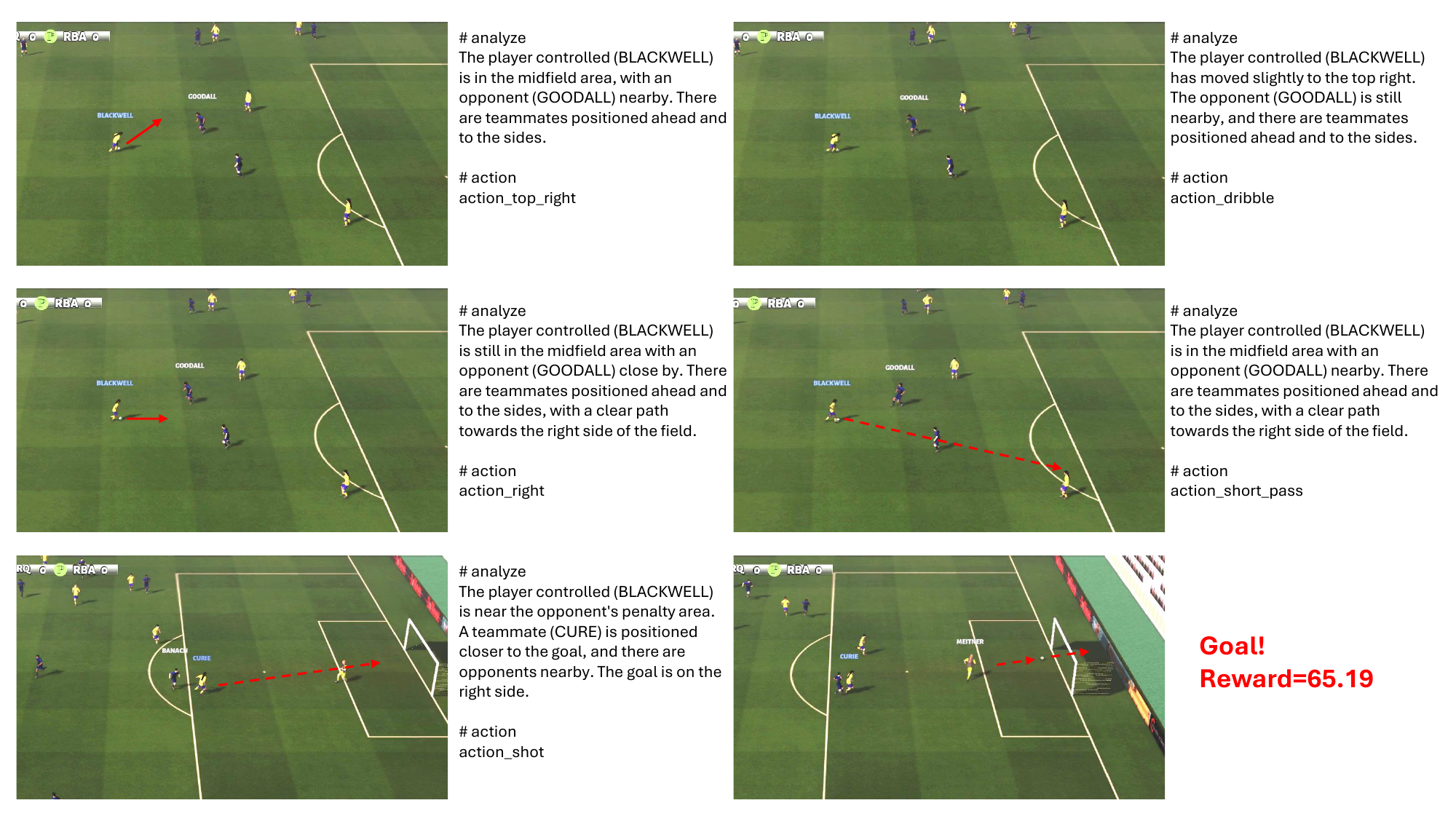}
    \caption{An example of GPT-4o's  performances on Football-Online settings.}
    \label{fig:footballg4oonline}
\end{figure*}

\subsection{Football}

\subsubsection{Task prompts $p_{cot},  p_{io}$}
\label{sec:ftbtaskprompt}
Because football is an multi-agent system, and there are 21 players controlled by the built-in AI bot, which bring a giant amount of randomness, it is meaningless to setup Global setting for this environment. Thus, this environment only supports \textbf{Online setting}, whose prompt is exactly the same as Sokoban-Online, except that the IO prompt part is replaced with:

\begin{verbatim}
You should perform 1 action for each 
generation, with this format: 
# analyze
some analyze
# action
action ID

For example: 
# analyze
...
# action
action_left
\end{verbatim}

\newpage
\subsubsection{Output examples}
\label{sec:ftbout}
Figure \ref{fig:footballg4oonline} shows the output and images of GPT-4o under the Football-Online setting. In the video, the player successfully dribbles past opponents, then passes the ball, and finally scores a goal.

\end{document}